\documentclass[11pt]{article}

\usepackage{titlesec}

\usepackage{amsfonts,amsmath,amssymb,amsthm,color,fullpage}
\definecolor{weborange}{rgb}{.8,.3,.3}
\definecolor{webblue}{rgb}{0,0,.8}
\definecolor{internallinkcolor}{rgb}{0,.5,0}
\definecolor{externallinkcolor}{rgb}{0,0,.5}

\providecommand{\remove}[1]{}

\usepackage{amsthm} 

\usepackage{comment}

\usepackage{enumerate}
\usepackage[labelfont=bf]{caption}

\usepackage[
hyperfootnotes=false,
colorlinks=true,
urlcolor=externallinkcolor,
linkcolor=internallinkcolor,
filecolor=externallinkcolor,
citecolor=internallinkcolor,
breaklinks=true,
pdfstartview=FitH,
pdfpagelayout=OneColumn]{hyperref}

\usepackage[backend=bibtex,style=ieee-alphabetic,natbib=true,backref=true]{biblatex} 
\usepackage{cleveref}

\usepackage{graphicx}

\providecommand{\remove}[1]{}

\ifdefined\IsDraft
    \newcommand{\authnote}[2]{{\bf [{\color{red} #1's Note:} {\color{blue} #2}]}}
\else
    \newcommand{\authnote}[2]{}
\fi

\titleclass{\subsubsubsection}{straight}[\subsection]

\newcounter{subsubsubsection}[subsubsection]
\renewcommand\thesubsubsubsection{\thesubsubsection.\arabic{subsubsubsection}}

\titleformat{\subsubsubsection}
{\normalfont\normalsize\bfseries}{\thesubsubsubsection}{1em}{}
\titlespacing*{\subsubsubsection}
{0pt}{3.25ex plus 1ex minus .2ex}{1.5ex plus .2ex}

\makeatletter
\renewcommand\paragraph{\@startsection{paragraph}{5}{\z@}%
	{3.25ex \@plus1ex \@minus.2ex}%
	{-1em}%
	{\normalfont\normalsize\bfseries}}
\renewcommand\subparagraph{\@startsection{subparagraph}{6}{\parindent}%
	{3.25ex \@plus1ex \@minus .2ex}%
	{-1em}%
	{\normalfont\normalsize\bfseries}}
\def\toclevel@subsubsubsection{4}
\def\toclevel@paragraph{5}
\def\toclevel@paragraph{6}
\def\l@subsubsubsection{\@dottedtocline{4}{7em}{4em}}
\def\l@paragraph{\@dottedtocline{5}{10em}{5em}}
\def\l@subparagraph{\@dottedtocline{6}{14em}{6em}}
\makeatother

\newcommand{\sdotfill}{\textcolor[rgb]{0.8,0.8,0.8}{\dotfill}} 

\newenvironment{algorithm}{\begin{mybox} \vspace{-.1in}\begin{algo}}{ \vspace{-.1in} \end{algo}\end{mybox}}

\newenvironment{mybox}{\begin{center}\begin{tabular}{|p{0.97\linewidth}|c|}   \hline} {  \\ \hline \end{tabular} \end{center}}	

\let\originalleft\left
\let\originalright\right
\renewcommand{\left}{\mathopen{}\mathclose\bgroup\originalleft}
\renewcommand{\right}{\aftergroup\egroup\originalright}

\newcommand{\ceil}[1]{\left\lceil #1 \right\rceil}
\newcommand{\ip}[1]{\iprod{#1}}
\newcommand{\iprod}[1]{\langle #1 \rangle}

\newcommand{\set}[1]{\ens{#1}}

\newcommand{\paren}[1]{\left(#1\right)}

\newcommand{\floor}[1]{\left \lfloor#1 \right \rfloor}

\newcommand{\norm}[1]{\left\lVert#1\right\rVert}

\newcommand{\eqdef}{:=}

\newcommand{\N}{{\mathbb{N}}}

\newcommand{\zo}{\set{0,1}}

\newcommand{\condition}{\;\ifnum\currentgrouptype=16 \middle\fi|\;}

\newcommand{\eps}{\varepsilon}

\newcommand{\la}{\gets}

\newcommand{\argmin}{\operatorname*{argmin}}

\renewcommand{\cref}{\Cref}
\usepackage{aliascnt}

\newtheorem{theorem}{Theorem}[section]

\newaliascnt{lemma}{theorem}
\newtheorem{lemma}[lemma]{Lemma}
\aliascntresetthe{lemma}
\crefname{lemma}{Lemma}{Lemmas}

\newaliascnt{observation}{theorem}
\newtheorem{observation}[observation]{Observation}
\aliascntresetthe{observation}
\crefname{observation}{Observation}{Observation}

\newaliascnt{claim}{theorem}
\newtheorem{claim}[claim]{Claim}
\aliascntresetthe{claim}
\crefname{claim}{Claim}{Claims}

\newaliascnt{corollary}{theorem}
\newtheorem{corollary}[corollary]{Corollary}
\aliascntresetthe{corollary}
\crefname{corollary}{Corollary}{Corollaries}

\newaliascnt{construction}{theorem}

\aliascntresetthe{construction}
\crefname{construction}{Construction}{Constructions}

\newaliascnt{fact}{theorem}
\newtheorem{fact}[fact]{Fact}
\aliascntresetthe{fact}
\crefname{fact}{Fact}{Facts}

\newaliascnt{proposition}{theorem}
\newtheorem{proposition}[proposition]{Proposition}
\aliascntresetthe{proposition}
\crefname{proposition}{Proposition}{Propositions}

\newaliascnt{conjecture}{theorem}

\aliascntresetthe{conjecture}
\crefname{conjecture}{Conjecture}{Conjectures}

\newaliascnt{definition}{theorem}
\newtheorem{definition}[definition]{Definition}
\aliascntresetthe{definition}
\crefname{definition}{Definition}{Definitions}

\newaliascnt{notation}{theorem}

\aliascntresetthe{notation}
\crefname{notation}{Notation}{Notation}

\newaliascnt{assertion}{theorem}

\aliascntresetthe{assertion}
\crefname{assertion}{Assertion}{Assertion}

\newaliascnt{assumption}{theorem}

\aliascntresetthe{assumption}
\crefname{assumption}{Assumption}{Assumption}

\newaliascnt{remark}{theorem}
\newtheorem{remark}[remark]{Remark}
\aliascntresetthe{remark}
\crefname{remark}{Remark}{Remarks}

\newaliascnt{question}{theorem}
\newtheorem{question}[question]{Question}
\aliascntresetthe{question}
\crefname{question}{Question}{Question}

\newaliascnt{example}{theorem}
\ifdefined\excludeexample
\else
   
\fi

\aliascntresetthe{example}
\crefname{exmaple}{Example}{Examples}

\crefname{equation}{Equation}{Equations}

\newaliascnt{proto}{theorem}

\newtheorem{proto}[proto]{Protocol}

\aliascntresetthe{proto}
\crefname{proto}{protocol}{protocols}

\newaliascnt{algo}{theorem}
\newtheorem{algo}[algo]{Algorithm}
\aliascntresetthe{algo}
\crefname{algo}{algorithm}{algorithms}

\newaliascnt{expr}{theorem}
\newtheorem{expr}[expr]{Experiment}
\aliascntresetthe{expr}
\crefname{experiment}{experiment}{experiments}

\def\FullBox{$\Box$}
\def\qed{\ifmmode\qquad\FullBox\else{\unskip\nobreak\hfil
\penalty50\hskip1em\null\nobreak\hfil\FullBox
\parfillskip=0pt\finalhyphendemerits=0\endgraf}\fi}

\def\qedsketch{\ifmmode\Box\else{\unskip\nobreak\hfil
\penalty50\hskip1em\null\nobreak\hfil$\Box$
\parfillskip=0pt\finalhyphendemerits=0\endgraf}\fi}

{\proof[#1]\list{}{\leftmargin=1cm\rightmargin=1cm}}
{\endproof%
  \vspace{10pt}
}

\newcommand{\ex}[1]{\Ex\left[#1\right]}
\newcommand{\Ex}{{\mathrm E}}
\renewcommand{\Pr}{{\mathrm {Pr}}}
\newcommand{\pr}[1]{\Pr\left[#1\right]}
\newcommand{\ppr}[2]{\Pr_{#1}\left[#2\right]}

\newcommand{\Ac}{\mathsf{A}}

\newcommand{\Bc}{\mathsf{B}}

\newcommand{\cC}{{\mathcal{C}}}

\newcommand{\tcP}{{\tilde{\cP}}}

\newcommand{\ens}[1]{\{#1\}}
\newcommand{\size}[1]{\left|#1\right|}

\def\cA{{\cal A}}
\def\cB{{\cal B}}
\def\cC{{\cal C}}
\def\cD{{\cal D}}

\def\cG{{\cal G}}
\def\cH{{\cal H}}
\def\cI{{\cal I}}
\def\cJ{{\cal J}}

\def\cN{{\cal N}}

\def\cP{{\cal P}}
\def\cQ{{\cal Q}}
\def\cR{{\cal R}}
\def\cS{{\cal S}}
\def\cT{{\cal T}}
\def\cU{{\cal U}}

\def\cX{{\cal X}}

\def\bbI{{\mathbb I}}

\def\bbN{{\mathbb N}}

\def\bbR{{\mathbb R}}

\newcommand{\Tableofcontents}{
\thispagestyle{empty}
\pagenumbering{gobble}
\clearpage
\tableofcontents
\thispagestyle{empty}
\clearpage
\pagenumbering{arabic}
}

\DeclareMathOperator{\Bin}{Bin}

\newcommand{\pt}[1]{{\bf #1}}
\newcommand{\px} {\pt{x}}
\newcommand{\py} {\pt{y}}
\newcommand{\pz} {\pt{z}}
\newcommand{\pa} {\pt{a}}

\newcommand{\pc} {\pt{c}}

\newcommand{\pX} {\pt{X}}

\newcommand{\pZ} {\pt{Z}}
\newcommand{\pW} {\pt{W}}
\newcommand{\pG} {\pt{G}}

\newcommand{\pmu}{\mathbf{\mu}}

\newcommand{\Cov}{{\rm Cov}}

\newcommand{\hpa} {\hat{\pa}}

\newcommand{\hpc} {\hat{\pc}}

\newcommand{\tpa} {\tilde{\pa}}

\newcommand{\tpc} {\tilde{\pc}}

\newcommand{\OPT}{{\rm OPT}}

\newcommand{\COST}{{\rm COST}}

\newcommand{\VC}{{\rm VC}}

\newcommand{\error}{{\rm error}}

\newcommand{\Lap}{{\rm Lap}}

\newcommand{\cball}{\cC_{\text{ball}}}
\newcommand{\cballs}{\cC_{k\text{-balls}}}

\newcommand{\Alg}{\cA}

\newcommand{\AlgTestPartition}{\mathsf{PrivateTestPartition}}

\newcommand{\AlgTestCloseTuples}{\mathsf{PrivateTestCloseTuples}}

\newcommand{\AlgPrivatekAverages}{\mathsf{PrivatekAverages}}
\newcommand{\AlgPrivateNoisykCenters}{\mathsf{PrivatekNoisyCenters}}
\newcommand{\AlgPrivatekMeans}{\mathsf{PrivatekMeans}}
\newcommand{\AlgGenerateCenters}{\mathsf{GenCenters}}
\newcommand{\AlgEstimateAverages}{\mathsf{EstimateAverages}}

\newcommand{\tAlgGenerateCenters}{\widetilde{\mathsf{Gen}}{\mathsf{Centers}}}
\newcommand{\tAlgPrivatekMeans}{\widetilde{\mathsf{Priv}}{\mathsf{atekMeans}}}

\newcommand{\AlgCollectEmpiricalMeans}{\mathsf{GenEmpiricalMeans}}
\newcommand{\AlgPrivatekGaussians}{\mathsf{PrivatekGMM}}

\newcommand{\poly}{{\rm poly}}
\newcommand{\polylog}{{\rm polylog}}
\newcommand{\Status}{{ Status}}

\newcommand{\Avg}{{\rm Avg}}

\newcommand{\dTV}{{\rm d}_{\rm TV}}

\newcommand{\grid}{g}

\newcommand{\lleft}{{\rm left}}
\newcommand{\rright}{{\rm right}}

\newcommand{\Partition}{{\rm Partition}}

\newcommand{\Points}{{\rm Points}}
\newcommand{\pass}{{\rm pass}}

\makeatletter
\let\xx@thm\@thm
\AtBeginDocument{\let\@thm\xx@thm}
\makeatother

\newcommand{\Enote}[1]{\authnote{Eliad}{#1}}
\newcommand{\Hnote}[1]{\authnote{Haim}{#1}}
\newcommand{\YMnote}[1]{\authnote{YM:}{#1}}
\newcommand{\ECnote}[1]{\authnote{EC:}{#1}}

\title{Differentially-Private Clustering of Easy Instances}
\author{Edith Cohen\thanks{Google Research and Blavatnik School of Computer Science, Tel Aviv University.
E-mails:\{\texttt{edith@alumni.stanford.edu}, \texttt{haimk@tau.ac.il}, \texttt{mansour.yishay@gmail.com},    \texttt{eliadtsfadia@gmail.com}\}}
\and Haim Kaplan$^\ast$
\and Yishay Mansour$^\ast$
\and Uri Stemmer\thanks{Google Research and Ben-Gurion University. E-mail: \{\texttt{u@uri.co.il}\}.}
\and Eliad Tsfadia$^\ast$}

\begin{document}

\maketitle
\begin{abstract}

Clustering is a fundamental problem in data analysis. In differentially private clustering, the goal is to identify $k$ cluster centers without disclosing information on individual data points. Despite significant research progress, the problem had so far resisted practical solutions.  In this work we aim at providing simple implementable differentially private clustering algorithms that provide utility when the data is "easy," e.g., when there exists a significant separation between the clusters.

We propose a framework that allows us to apply non-private clustering algorithms to the easy instances and privately combine the results.
 We are able to get improved sample complexity bounds in some cases of Gaussian mixtures and $k$-means. We complement our theoretical analysis with an empirical evaluation on synthetic data.

\remove{
\YMnote{
Continuing with Edith abstract:

Clustering is a fundamental problem in data analysis. In differentially private clustering, the goal is to identify $K$ cluster centers without disclosing information on individual data points. Despite significant research progress, the problem had so far resisted practical solutions.  In this work we aim at providing simple implementable differentrially private clustering algorithms when the the data is "easy," e.g., when there exists a significant separation between the clusters.

For the easy instances we consider, we have a simple implementation based on utilizing non-private clustering algorithms, and combining them privately. We are able to get improved sample complexity bounds in some cases of Gaussian mixtures and k-means. We complement our theoretical algorithms with experiments of simulated data. 


}

\Enote{OLD:
   Consider the following task, which we refer to as the $k$-tuples-problem: Let $\cT$ be a database of unordered $k$-tuples over $\bbR^d$, and assume that there exists $k$ very far balls over $\bbR^d$, such that each ball contains exactly one point of each $k$-tuple in $\cT$. The goal is to privately compute a new $k$-tuple that is (close to be) in these $k$ far balls.
    
    We present two simple (very similar) differentially-private algorithms for solving the $k$-tuples-problem. The first one assumes a smaller separation, but the second one requires less tuples, making it more practical when the separation is large.
    
    We then show that clustering problems of easy instances can be reduced to the $k$-tuples-problem. As a first example, we present a differentially-private algorithm for $k$-means clustering of well-separated instances, which simplifies the state-of-the-art result of [Shechner et al. AISTATS 2020]. As a second example, we present a differentially-private algorithm for learning the parameters of mixtures of well-separated Gaussians, which simplifies and improves the state-of-the-art result of [Kamath et al. NeurIPS 2019].
    
    Finally, we present experimental results for learning Mixtures of very separated Gaussians, making our work be the first that present a practical differentially-private algorithm for this task.
    
    \ECnote{
    \emph{Perhaps start with something like this before jumping into $k$-tuples.}
    
    Clustering is a fundamental problem in data analysis. In differentially private clustering, the goal is to identify $K$ cluster centers without disclosing information on individual data points. Despite significant research progress, the problem had so far resisted practical solutions.  In this work we introduce an approach towards this goal when the data is "easy."
    
    We introduce the problem of separating $k$-tuples.... and design differentially private algorithms....
    \em {I would not get into the details in the abstract on the different algorithms to $k$-tuples...}
    
    }

}
}

\end{abstract}

\Tableofcontents

\section{Introduction}
Differential privacy \cite{DworkMNS06} is a mathematical definition of privacy, that aims to enable statistical analyses of databases while providing strong guarantees that individual-level information does not leak. Privacy is achieved in differentially private algorithms through randomization and the introduction of ``noise'' to obscure the effect of each individual, and thus differentially private algorithms can be less accurate than their non-private analogues. In most cases, this loss in accuracy is studied theoretically, using asymptotic tools. As a result, there is currently a significant gap between what is known to be possible {\em theoretically} and what can be done {\em in practice} with differential privacy. In this work we take an important step towards bridging this gap 
in the context of {\em clustering related tasks}.

The construction of differentially private clustering algorithms has attracted a lot of attention over the last decade, and many different algorithms have been suggested.\footnote{\cite{BDMN05,NRS07,FFKN09,
McSherry09,GuptaLMRT10,Mohan2012,Wang2015,NockCBN16,
Su2016,NSV16,DannyPrivatekMeans,Balcan17a,
NS18_1Cluster,HuangL18,KaplanSt18,Stemmer20,ShechnerSS20,Ghazi0M20,Nguyen20}} However, to the best of our knowledge, none of these algorithms have been implemented:  They are not particularly simple and suffer from large hidden constants that translate to a significant loss in utility, compared to non-private implementations.

\begin{question}
How hard is it to cluster privately with a practical implementation? 
\end{question}

We take an important step in this direction using the following approach. Instead of directly tackling ``standard'' clustering tasks, such as $k$-means clustering, we begin by identifying a very simple clustering problem that still seems to capture many of the challenges of practical implementations (we remark that this problem is completely trivial without privacy requirements). We then design effective (private) algorithms for this simple problem. Finally, we reduce ``standard'' clustering tasks to this simple problem, thereby obtaining private algorithms for other tasks.

In more detail, we introduce the following problem, called the {\em $k$-tuple clustering} problem. 

\begin{definition}[informal, revised in Definition~\ref{def:ktupleclustering}]
An instance of the {\em $k$-tuple clustering} problem is a collection of $k$-tuples. 
Assuming that the input tuples can be partitioned into $k$ ``obvious clusters'', each consisting of one point of each tuple, then the goal is to report $k$ ``cluster-centers'' that correctly partition the input tuples into clusters.
If this assumption on the input structure does not hold, then the outcome is not restricted.
\end{definition}

\begin{remark}\;
\begin{enumerate}
    \item By ``obvious clusters'' we mean clusters which are far away from each other.\vspace{-7pt}
    \item The input tuples are {\em unordered}. This means, e.g.,\ that the ``correct'' clustering might place the first point of one tuple with the fifth point of another tuple.\vspace{-7pt}
    \item Of course, we want to solve this problem while guaranteeing differential privacy. Intuitively, this means that the outcome of our algorithm should not be significantly effected when arbitrarily modifying one of the input tuples.
\end{enumerate}

\end{remark}

Observe that without the privacy requirement this task is trivial: We can just take one arbitrary input tuple $(x_1,...,x_k)$ and report it. With the privacy requirement, this task turns out to be non-trivial. It’s not that this problem cannot be solved with differential privacy. It can. It’s not even that the problem requires large amounts of data asymptotically. It does not. However, it turns out that
designing an implementation with a practical privacy-utility tradeoff, that is effective  on  
 finite datasets (of reasonable size), is quite challenging. 


\subsection{Our algorithms for the ${\boldsymbol k}$-tuple problem}

We present two (differentially private) algorithms for the $k$-tuple clustering problem, which we call $\AlgPrivatekAverages$ and $\AlgPrivateNoisykCenters$.
Both algorithms first privately test if indeed the input is partitioned into $k$ obvious clusters and quit otherwise.
They differ by the way they compute the centers in case this test passes.
Algorithm $\AlgPrivatekAverages$  privately averages each identified cluster. Algorithm $\AlgPrivateNoisykCenters$, on the other hand, does not operate by averaging clusters. Instead, it  selects one of the input $k$-tuples, and then adds a (relatively small) Gaussian noise to every point in this tuple. We prove that this is private if indeed there are $k$ obvious clusters in the input.
We evaluate these two algorithms empirically, and show that, while algorithm $\AlgPrivatekAverages$ is ``better in theory'', algorithm $\AlgPrivateNoisykCenters$ is much more practical for some interesting regimes of parameters.

We now give a simplified overview of the ideas behind our algorithms. For concreteness, we focus here on $\AlgPrivatekAverages$. Recall that in the $k$-tuple clustering problem, we are only required to produce a good output assuming the data is ``nice'' in the sense that the input  tuples can be clustered into $k$ ``far clusters'' such that every cluster contains exactly one point from every tuple. However, with differential privacy we are ``forced'' to produce good outputs even when this niceness assumption does not hold. This happens because if the input data is ``almost nice'' (in the sense that modifying a small number of tuples makes it nice) then differential privacy states that the outcome of the computation should be close to what it is when the input data is nice. 

So, the definition of differential privacy forces us to cope with ``almost nice'' datasets. Therefore, the niceness test that we start with has to be a bit clever and ``soft'' and
succeed with some probability also for 
data which is ``almost nice''.
Then, in order to achieve good performances, we have  to utilize the assumption that the data is  ``almost nice'' when we compute the private centers.
To compute these centers,  Algorithm $\AlgPrivatekAverages$ determines ({\em non-privately}) a clustering of the input tuples, and then averages (with noise) each of the clusters. The conceptual challenge here is to show that even though the clustering of the data is done non-privately, it is stable enough such that the outcome of this algorithm still preserves privacy.

\subsection{Applications}\label{sec:introApplications}

The significance of algorithms $\AlgPrivatekAverages$ and $\AlgPrivateNoisykCenters$ is that 
many clustering related tasks can be privately solved by a reduction to the $k$-tuple clustering problem. In this work we explore two important use-cases: (1) Privately approximating the $k$-means under stability assumption, and (2) Privately learning the parameters of a mixture of well-separated Gaussians.

\smallskip
\noindent
{\bf ${\boldsymbol k}$-Means Clustering}

In $k$-means clustering, we are given a database $\cP$ of $n$ input points in $\bbR^d$, and the goal is to identify a set $C$ of $k$ {\em centers} in $\bbR^d$ that minimizes the sum of squared distances from each input point to its nearest center. This problem is NP-hard to solve exactly, and even NP-hard to approximate to within a multiplicative factor smaller than $1.0013$ \citep{lee2017improved}. The current (non-private) state-of-the-art algorithm achieves a multiplicative error of $6.357$ \citep{ahmadian2019better}.

One avenue that has been very fruitful in obtaining more accurate algorithms (non-privately) is to look beyond worst-case analysis \cite{OstrovskyRSS12,awasthi2010stability,ABS12,balcan2009approximate,bilu2012stable,kumar2010clustering}. In more details, instead of constructing algorithms which are guaranteed to produce an approximate clustering for any instance, works in this vain  give stronger accuracy guarantees by focusing only on instances that adhere to certain ``nice'' properties (sometimes called stability assumptions or separation conditions). The above mentioned works showed that such ``nice'' inputs can be clustered much better than what
is possible in the worst-case (i.e., without assumptions on the data).

Given the success of non-private stability-based clustering, it is not surprising that such stability assumptions were also utilized in the privacy literature, specifically by \citet{NRS07,Wang2015,HuangL18,ShechnerSS20}. While several interesting concepts arise from these four works, none of their algorithms have been implemented, their algorithms are relatively complex, and their practicability on finite datasets is not clear.

We show that the problem of stability-based clustering (with privacy) can be reduced to the $k$-tuple clustering problem. Instantiating this reduction with our algorithms for the $k$-tuple clustering problem, we obtain a simple and practical algorithm for clustering ``nice'' $k$-means instances privately. 

\smallskip
\noindent
{\bf Learning Mixtures of Gaussians.}
Consider the task of {\em privately} learning the parameters of an unknown mixtures of Gaussians given i.i.d.\ samples from it.  By now, there are various private algorithms that learn the parameters of a {\em single} Gaussian \cite{KV18,KLSU19,CWZ19,BS19,KSU20,BDKU20}. Recently, \cite{KSSU19} presented a private algorithm for learning mixtures of well-separated (and bounded) Gaussians. 
We remark, however, that besides the result of \cite{BDKU20}, which is a practical algorithm for learning a single Gaussian, all the other results are primarily theoretical. 

By a reduction to the $k$-tuples clustering problem, we present a simple algorithm that privately learns the parameters of a separated (and bounded) {\em mixture} of $k$ Gaussians. 
From a practical perspective, compared with  the construction of the main algorithm of \cite{KSSU19}, our algorithm is simple and implementable. From a theoretical perspective, our algorithm offers reduced sample complexity, weaker separation assumption, and modularity. See \cref{sec:Gauss:comparison} for the full comparison.


\subsection{Other Related Work}
The work of \citet{NRS07} presented the sample-and-aggregate method to convert a non-private algorithm into a private algorithm, and applied it to easy clustering problems. However, their results are far from being tight, and they did not explore certain considerations (e.g., how to minimize the impact of a large domain in learning mixture of Gaussians). 

Another work by \citet{BKSW19} provides a general method to convert from a cover of a class of distributions to a private learning algorithm for the same class. The work gets a near-optimal sample complexity, but the algorithms have exponential running time in both $k$ and $d$ and their learning guarantees are incomparable to ours (they perform proper learning, while we provide clustering and parameter estimation).

In the work of \cite{KSSU19}, they presented an alternative algorithm for learning mixtures of Gaussians, which optimizes the sample-and-aggregate approach of \cite{NRS07}, and is somewhat similar to our approach. That is, their algorithm executes a non-private algorithm several times, each time for obtaining a new ``$k$-tuple'' of means estimations, and then aggregates the findings by privately determine a new $k$-tuple of means estimation. 
But their approach has two drawbacks. First, in order to privately do that, their algorithm ignores the special $k$-tuples structure, and apply a more wasteful and complicated ``minimal enclosing ball'' algorithm from \cite{NS17_clustering,NSV16}. Second, in contrast to them, for creating a $k$-tuple, our algorithm only applies a non-private algorithm for \emph{separating} the samples in the mixture (i.e., for determine which samples belong to the same Gaussian), and not for estimating their parameters. This yields that we need less samples per invocation of the non-private algorithm for creating a single $k$-tuple, which results with an improved sample complexity (each $k$-tuple in our case is just the averages of each set of samples, which might not necessarily be very close to the true means, but is close enough for our setting where the Gaussians are well-separated). Finally, given a private separation of the sample, we just apply some private algorithm for estimating the parameters of each (single) Gaussian (e.g., \cite{KV18,KLSU19,CWZ19,BS19,KSU20,BDKU20}).
For more details about our construction, see \cref{sec:mixture-of-gaus}.

Furthermore, there are many differentially-private algorithms that are related to learning mixture of Gaussians (notably PCA) \cite{BlumDwMcNi05,KT13,CSS13,DTTZ14}, and differentially-private algorithms for clustering \cite{NRS07,GuptaLMRT10,NSV16,NS17_clustering,Balcan17a,KaplanSt18,HuangL18,Ghazi0M20}. We remark that for the learning Gaussians mixtures problem, applying these algorithms naively would introduce a polynomial dependence on the range of the data, which we seek to avoid.

\remove{
\Enote{Old intro:}

\section{Introduction}

\ECnote{Add general text about what is privacy and its importance ... }

put text

\ECnote{Clustering-type problems, define the DP requirement (central model) in that context}

In $k$-means clustering, we are given a database $\cP$ of $n$ points in $\bbR^d$, and the goal is to identify a set $C$ of $k$ centers in $\bbR^d$ that approximately minimizes $\COST_{\cP}(C) = \sum_{\px \in \cP} \min_{\pc \in C}\norm{\px - \pc}^2$. We denote the lowest possible cost as $\OPT_{k}(\cP)$. Since the task of minimizing the $k$-means is NP-hard, the literature has focused on approximation algorithms, with the current (non-private) state-of-the-art achieving multiplicative error of $6.357$ \cite{ahmadian2019better}. 
\ECnote{Mention practical solutions to the problem. Cite, EM, $k$-means++, bi-criteria.  Also cite earlier results on multiplicative factors theoretical.  Make a distinction between theory (very large running times, asymptotically) and practice.}

\ECnote{Condense the below to precisely define neighboring and DP} 
\begin{definition}[Neighboring databases]\label{def:neighboring}
	Let $\cD = \set{x_1,\ldots,x_n}$ and $\cD' = \set{x_1',\ldots,x_n'}$ be two databases over a domain $\cX$. We say that $\cD$ and $\cD'$ are \textbf{neighboring} if there is exactly one index $i \in [n]$ with $x_i \neq x_i'$.
\end{definition}

\begin{definition}[$(\eps,\delta)$-indistinguishable]\label{def:indis}
	Two random variable $X,X'$ over a domain $\cX$ are called $(\eps,\delta)$-indistinguishable, if for any event $T \subseteq \cX$, it holds that
	$\pr{X \in T} \leq e^{\eps} \cdot \pr{Y \in T} + \delta$. If $\delta = 0$, we say that $X$ and $X'$ are $\eps$-indistinguishable.
\end{definition}

\begin{definition}[$(\eps,\delta)$-differential privacy \cite{DworkMNS06}]\label{def:DP}
	An algorithm $\Alg$ is called $(\eps,\delta)$-differentially private, if for any two neighboring databases $\cD,\cD'$ it holds that $\Alg(\cD)$ and $\Alg(\cD')$ are $(\eps,\delta)$-indistinguishable. If $\delta = 0$ (i.e., pure privacy), we say that $\Alg$ is called $\eps$-differentially private.
\end{definition}

\ECnote{Add text, borrowing from below, on results for DP clustering in the central model, difficulties, gap from practice. Why these problems so far defied practical solutions. Also on mixtures of Gaussians.}

Unlike in the non-private literature, it is known that every private algorithm for approximating the $k$-means cost must have an additive error (even computationally unbounded algorithms), which scales with the diameter of the input space. Hence, it is standard to assume an upper bound $\Lambda$ on the $\ell_2$ norm of all the input points $\cP$. Typically (though not always), one aims to minimize the multiplicative error while keeping the additive error at most polylogarithmic in the size of the database. The current state-of-the-art construction \Enote{ .....}
\ECnote{cite line of prior work on DP clustering (without assumptions on input). Culminating in Ravi/Nguyen. stressing not practical but illuminating}

\ECnote{Meta question:  Can we get closer to practice when the data is "Easy."?}

\ECnote{Discuss easy or stable instances, what is known without privacy. In particular discuss $k$-means clustering and separating mixtures.}

The non-private literature is ripe with results geared for efficient solutions on easier instances.  Including dodging the NP hardness of unrestricted $k$-means clustering.  

{\bf edit below}
As a common concrete example of such $\cP$ and $\cA$ that satisfy the above stability property is when $\cA$ is
a good (non-private) approximation algorithm for $k$-means, and $\cP$ is well-separated for $k$-means, a notion that was first introduced in \citet{OstrovskyRSS12}. Formally, $\cP$ is called $\phi$-separated if $\OPT_{k}(\cP) \leq \phi^2 \OPT_{k-1}(\cP)$. It turns out that if $\cP$ is $\phi$-separated for $k$-means, and $\cA$ is a $\omega$-approximation algorithm for $k$-means, where $\phi^2(1 + \omega)$ is sufficiently small, then $\cP$ and $\cA$ satisfy our required stability property, yielding that our algorithm $\AlgPrivatekMeans$ succeed well over $\cP$ using $\cA$.

Following the work of \citet{OstrovskyRSS12}, several other works have related other notions of input-stability to clustering \cite{awasthi2010stability,awasthi2012center,balcan2009approximate,bilu2012stable,kumar2010clustering}. 

\ECnote{Discuss easy or stable instances, what is known with privacy.  } 

{\bf edit below}
In addition, several works gave differentially-private algorithms for approximating the minimal Wasserstein distance (\cite{Wasserstein1969}) of  well-separated instances \cite{NRS07,wang2015differentially,huang2018optimal}, and in a recent work, \citet{ShechnerSS20} presented a differentially-private algorithm for approximating the $k$-means of such well-separated instances. The main result of \cite{ShechnerSS20} is a very simple transformation, called Private-Stable-$k$-Means, from any private $\omega$-approximation algorithm to a private $(1 + O(\phi^2))$-one, that works over $\phi$-separated database $\cP$ with sufficiently small $\phi^2(1 + \omega)$. In addition, they presented Algorithm SampleAggregate-$k$-means that transform any (non-private) $k$-means approximation algorithm $\cA$ into a private one, with very similar guarantees to our construction. Their algorithm also randomly partitions $\cP$ into $T$ subsets (for large enough $T \ll n$), and executes $\cA$ on each subset to obtain $k$ centers that are close to the optimal centers. But then, in order to privately estimating the averages of each cluster of the resulting partitioned multiset of $k$-tuples, their algorithm ignores the $k$-tuples structure and uses the (complicated) \cite{NS17_clustering}'s minimal enclosing ball  algorithm for determine the clusters, while our algorithm uses Algorithm $\AlgPrivatekAverages$ which is much more simpler and practical.

\ECnote{Perhaps here discuss mixture of Gaussians, what is known, what would we consider stable.} 

\ECnote{Our approach and contributions at the high level}

We define a basic problem of \emph{$k$-tuple clustering}.
The input to $k$-tuple clustering is a set $\cT$ of unordered $k$-tuples of vectors in $\bbR^d$.   The desired output is a single $k$-tuple that (informally) perfectly separates a maximum number of tuples from $\cT$.
We design differentially private algorithms for $k$-tuple clustering that (informally) provide good utility on datasets for which a good solution exist. 
The algorithms we design for $k$-tuple clustering are relatively simple and friendlier to implement compared with alternative solutions based on existing differentially private tools and at the same time also provide tighter asymptotic bounds. 

We propose a framework to tackle private solutions for $k$-clustering type problems, where the input is a dataset of points in $\bbR^d$ and the output a set of $k$ points (a $k$-tuple).
Our framework uses 
private $k$-tuple clustering as a tool and also assumes a non-private algorithm for the original problem that inputs a set of points and generates a $k$-tuple (e.g., $k$ cluster centers).
The framework applies the non-private algorithm multiple times, to subsets (random samples) of the data.  These runs yield a set $\cT$ of $k$-tuples.  Our private $k$-tuple clustering algorithm is then applies to $\cT$ to generate a private $k$-tuple.
The utility we obtain is conditioned on the instance being "stable" with respect to the original problem: Informally, that the $k$-tuples generated by the multiple applications of the non-private algorithm cluster well.

We apply our framework to two well-studied $k$-clustering type problems:
Our first application is to $k$-means clustering on well-separated instances.  We obtain a differentially-private algorithm that is considerably simpler and more efficient than the state-of-the-art result of \citet{ShechnerSS20}.
\ECnote{Put here more direct comparison to Shechner... (what we have that is not there)}

The second application is to privately learn the parameters of mixtures of well-separated Gaussians.
Our solution is both simpler and provides an asymptotic improvement over the state-of-the-art result of [Kamath et al. NeurIPS 2019].
    
Finally, we present empirical results for learning mixtures of very separated Gaussians, making our work the first that present a practical differentially-private algorithm for this task.
 
\section{The tuple clustering problem}    

\ECnote{Precise definition, discussion on non-private and alternative private approaches and what is obtained,  pseudocode,  statement of properties} 

A $k$-tuple 
$X = \set{\px_1,\ldots,\px_k}$
is an unordered set of $k$ vectors $\px_i \in \bbR^d$.
For a $k$-tuple $X$ and parameter $\Delta >0$
we define radii $(r_i^{X})_{i=1}^k$ so that
$r_i^X = \min_{j \neq i} \norm{\px_i - \px_j}$.

\begin{definition}[$\Delta$-partition of a tuple by a tuple]
We say that a $k$-tuple $X$ $\Delta$-partitions another $k$-tuple $Y$
if each entry $\px_i$ of $X$ can be uniquely matched to entry $\py_{i_j}$ of $Y$ so that
\[
\norm{\px_i - \py_{i_j}} \leq \frac1{\Delta} \cdot r_i^X(\Delta) .\]
\end{definition}

\Hnote{ Why $r_i^X(\Delta)$ ? and not $r_i^X$}

\begin{definition}[$\Delta$-far balls]
For a $k$-tuple $X$ and $\Delta$ we can consider a set of $k$ balls
$\cB = \set{B_1,\ldots,B_k}$, where $B_i(\px_i,\frac{1}{\Delta} r^X_i)$ has center $\px_i$ and radius, $\frac{1}{\Delta} r^X_i$.
We refer to such balls as $\Delta$-far balls.
\end{definition}
Equivalently, A tuple $Y$ is $\Delta$-partitioned by $X$ if each ball in $\cB$ contains exactly one entry from $Y$.
 
\Hnote{Don't we want to parameterize the balls by X ? $\cB^X$ ?}

\begin{definition}[$k$-tuple clustering]
The input to the problem is database $\cT$ of $k$-tuples of size $n$ and $\Delta>1$. The output is a tuple  that $\Delta$-partitions the maximum number of other tuples.  
\end{definition}

We define "easy" instances to be those with a perfect solution, that is, 
we are promised that there is a tuple that $\Delta$-partition all other tuples.

Without the requirement of privacy, the "easy" problem is quite easy to solve exactly.  
One can just pick an arbitrary $k$-tuple $\set{\px_1,\ldots,\px_k} \in \cT$, construct the cluster $\cP_1,\ldots,\cP_k$ where $\cP_i = \set{\px \in \Points(\cT) \colon i = \argmin_{j \in [k]} \norm{\px - \px_j}}$, and then compute the average $\pa_i$ of each $\cP_i$. \Hnote{why not just spit an arbitrary tuple ?}
However, this algorithm is not differentially private. To be differentially private, any outcome should be obtained with approximately the same probability when we change a single $k$-tuple (possibly moving to a neighboring database with no perfect solution \Hnote{We also have to deal with databases which are not easy ?}). In this algorithm, changing a single $k$-tuple may completely change the resulting partition (e.g., if the arbitrary tuple that we choose at the first step is exactly the different tuple), and hence, change the resulting averages.

We therefore must consider approximate solutions and probabilistic outputs. 
Even for $k=1$ (which is a task of finding a point close to all other points), we need to assume some upper bound $\Lambda$ on the $\ell_2$ norm of the points in each tuple in $\cT$.  Second, we have to assume $r_{\min} > 0$

\ECnote{We need to explain $r_{\min}$ this better.  Perhaps we can factor in $r_{\min}$ to the approximation. }

We say that an algorithm is an $(\alpha,\beta)$-approximation to a perfect $k$-tuple clustering instance $\cT$ if with probability $1-\beta$, it outputs $\set{pa_1,\ldots,\pa_k}$ that $\alpha\Delta$ partitions all tuples.

\ECnote{I did some editing.  Is the above ok? }

\ECnote{discuss approaches with existing components and why they are not satisfactory. Not simple, large constants.  }

A direct approach to a private solution is through the Minimal Enclosing Ball algorithm of \citet{NS17_clustering}. This algorithm is given $n$ points and a (large enough) parameter $t \leq n$, and privately outputs a ball of radius $O(r_{opt})$ that contains (almost) $t$ points, where $r_{opt}$ is the radius of the smallest ball that contains $t$ points.  To apply this algorithm to our problem, we ignore the $k$-tuple structure and execute \cite{NS17_clustering}'s algorithm $k$ times on the database $\Points(\cT)$. Each time it would privately determine a ball that contains (almost) all the points of one of the $\cP_i$'s, and remove them from $\Points(\cT)$ before the next call to the algorithm. In order to privately compute an estimate $\tpa_i$ of each average $\pa_i$ such that $\norm{\pa_i - \tpa_i}$ is proportional to the (minimal) radius of a ball that contains $\cP_i$, one can execute the private average algorithm of \citet{NSV16}. It can be shown that with this process, we can achieve $(\eps,\delta)$-differential privacy for $(\alpha,\beta)$-estimating the $k$ averages of $\cP$, where $\alpha = O\paren{\frac{\sqrt{d}}{\eps n} \log \paren{\frac{nd}{\beta}} \log\paren{\frac{1}{\delta}}}$.
However, the downside of this approach is that \cite{NS17_clustering}'s algorithm is not simple, and does not exploits the special $k$-tuples structure of the database.\Enote{maybe also explain why it is not simple. E.g., that it uses LSH, RecConcave, and more ...}\ECnote{Uri mentioned polynomial super-linear dependence on $kd$ also.  We need to make this paragraph concise and crisp.  We also use averaging (in one of the approaches), we just have a different method of creating the subsets.}

\ECnote{
Present our solution taking things from section 9 and hiding all the unnecessary.  Yes to Pseudocode. Precise statement of properties. just say enough so that we can precisely state the results and explain the algorithm.  No proofs but intutions. yes to pictures.  }

\section{Application to $k$-means clustering}

\section{Application to separating mixtures of Gaussians}

\section{Empirical results}

\section{Conclusion}

\subsection{tuple clustering}\label{sec:intro:kAver}
\ECnote{
Previous: Estimating the Averages of Partitioned $k$-Tuples
}

Suppose that we are given a database $\cT$ of size $n$, each element in $\cT$ is an (unordered) $k$-tuple over $\bbR^d$, and we are promised that $\cT$ can be partitioned by $k$ far balls in $\bbR^d$. Namely, there exist $k$ balls $B_1,\ldots,B_k$ over $\bbR^d$, each $B_i$ is centered at some point $\pc_i \in \bbR^d$ and has radius $r_i \geq 0$, and the following holds: (1) For every $i \neq j: \text{} \norm{\pc_i - \pc_j} > 16 \cdot \max\set{r_i,r_j}$ (i.e., the balls are very far from each other),\footnote{Throughout this work, we denote by $\norm{\cdot}$ the standard $\ell_2$ norm.} and (2) For every $i \in [k]$ and a $k$-tuple $X = \set{\px_1,\ldots,\px_k} \in \cT:\text{ }\size{X \cap B_i} = 1$ (i.e., each tuple in $\cT$ contains exactly one point in each ball).\Enote{add picture?}

Given this promise, let $\Points(\cT)$ be the multiset (of size $kn$) of all the points in all the $k$-tuples in $\cT$, and consider the task of estimating the average $\pa_i$ of each cluster $\cP_i = \Points(\cT) \cap B_i$.  Specifically, we say that an algorithm $(\alpha,\beta)$-estimates the averages of partitioned points, if given such a database $\cT$ as input, then with probability $1-\beta$, it outputs $\set{\tilde{\pa}_1,\ldots,\tilde{\pa}_k}$ such that, for each $i \in [k]$, there exists an average (call it $\pa_i$) with $\norm{\tpa_i - \pa_i} \leq \alpha \cdot r_i$.

Without privacy, it is quite easy to $(0,0)$-estimate the $k$ averages. 
One can just pick an arbitrary $k$-tuple $\set{\px_1,\ldots,\px_k} \in \cT$, construct the cluster $\cP_1,\ldots,\cP_k$ where $\cP_i = \set{\px \in \Points(\cT) \colon i = \argmin_{j \in [k]} \norm{\px - \px_j}}$, and then compute the average $\pa_i$ of each $\cP_i$. 
However, this algorithm is not differentially private. To be differentially private, any outcome should be obtained with approximately the same probability when we change a single $k$-tuple (possibly moving to a neighboring database that is not partitioned by far balls). In this algorithm, changing a single $k$-tuple may completely change the resulting partition (e.g., if the arbitrary tuple that we choose at the first step is exactly the different tuple), and hence, change the resulting averages.

In order to privately estimate the $k$ averages, we first need to make some assumptions on the database $\cT$, even for $k=1$ (which is just the task of estimating the average of all points). First, we need to assume some upper bound $\Lambda$ on the $\ell_2$ norm of the points in each tuple in $\cT$. Second, since we are interested in an average estimation that is proportional to the (minimal) radius of a ball that contains them, we also need to assume some lower bound $r_{\min} > 0$ on the radii of the (minimal) $k$ balls that partition $\cT$, or alternatively, assume that the points in $\Points(\cT)$ belong to a finite grid.

\Enote{Ref?}\Hnote{From ``Second, ...'':  I do not follow this, is it still related to $k=1$ ?
if all the points are at exactly the same place then we can report this place, i think ?
For $k>1$ maybe you can say that there is a reduction from finding an interior point, but we need to be careful}

One way to privately estimate the $k$ averages is to use the minimal enclosing ball algorithm of \citet{NS17_clustering}. This algorithm is given $n$ points and a (large enough) parameter $t \leq n$, and privately outputs a ball of radius $O(r_{opt})$ that contains (almost) $t$ points, where $r_{opt}$ is the radius of the smallest ball that contains $t$ points. In our case, one can ignore the $k$-tuple structure and execute \cite{NS17_clustering}'s algorithm $k$ times on the database $\Points(\cT)$. Each time it would privately determine a ball that contains (almost) all the points of one of the $\cP_i$'s, and remove them from $\Points(\cT)$ before the next call to the algorithm. In order to privately compute an estimation $\tpa_i$ of each average $\pa_i$ such that $\norm{\pa_i - \tpa_i}$ is proportional to the (minimal) radius of a ball that contains $\cP_i$, one can execute the private average algorithm of \citet{NSV16}. It can be shown that with this process, we can achieve $(\eps,\delta)$-differential privacy for $(\alpha,\beta)$-estimating the $k$ averages of $\cP$, where $\alpha = O\paren{\frac{\sqrt{d}}{\eps n} \log \paren{\frac{nd}{\beta}} \log\paren{\frac{1}{\delta}}}$.
However, the downside of this approach is that \cite{NS17_clustering}'s algorithm is not simple, and does not exploits the special $k$-tuples structure of the database.\Enote{maybe also explain why it is not simple. E.g., that it uses LSH, RecConcave, and more ...}

In this work we present a much more simpler and practical $(\eps,\delta)$-differentially private algorithm that  $(\alpha,\beta)$-estimates the averages of partitioned $k$-tuples, with essentially the same parameter $\alpha$. In order to understand the main idea of our algorithm,
suppose we are given an $n$-size database $\cT$ of $k$-tuples, and a set of far balls $\cB = \set{B_1,\ldots,B_k}$ around the centers $\set{\pc_1,\ldots,\pc_k}$ (respectively), that ``almost'' partitions $\cT$. By ``almost'' we mean that, there exists some $\ell \ll n$ such that, by removing $\ell$ tuples from $\cT$, we obtain a database that is partitioned by $\set{B_1,\ldots,B_k}$. Throughout this work, we call it $\ell$-partitioning. 

Now consider the following process $\AlgEstimateAverages(\cT,\cB)$: (1)  Partition $\Points(\cT)$ into $k$ multisets $\cP_1,\ldots,\cP_k$, where each $\cP_i$ consists of the points in $\cP$ that $\pc_i$ is closest center to them, and (2) Compute a noisy average $\tpa_i$ of each $\cP_i$ using \cite{NSV16}'s algorithm with privacy parameters $\eps,\delta \in (0,1]$, and output the set $\set{\tpa_1,\ldots,\tpa_k}$.

Our main observation is that, for any neighboring databases $\cT,\cT'$, and any sets of far balls $\cB$ and $\cB'$ that $\ell$-partition $\cP$ and $\cP'$ (respectively),   the distributions of the outputs $\AlgEstimateAverages(\cP,\cB)$ and $\AlgEstimateAverages(\cP',\cB')$ are $(O( k \ell \eps), O(k \ell \delta))$-indistinguishable.\Enote{define ind} The intuition is that, since $\cT$ and $\cT'$ are neighboring, the partition $\set{\cT_1,\ldots,\cT_k}$ in  $\AlgEstimateAverages(\cP,\cB)$ and $\set{\cT_1',\ldots,\cT_k'}$ in $\AlgEstimateAverages(\cP',\cB')$ must be almost the same (in fact, we show that there are at most $2\ell + 1$ tuples that the two partitions do not agree on). 

The first attempt for translating the above observation into a differentially private algorithm that estimates the $k$ averages of $\cT$, is to fix some small $\ell$, compute (non-privately) a set of far balls $\cB = \set{B_1,\ldots,B_k}$ that $\ell$-partitioned $\cT$ (fail if such balls do not exist), and then output $\AlgEstimateAverages(\cP,\cB)$. The main problem is that differential privacy is a worst-case guarantee. If we execute the above process with two neighboring databases $\cT,\cT'$ such that $\cT$ is $\ell$-partitioned by far balls and $\cT'$ is not $\ell$-partitioned by any far balls, then the execution over $\cT'$ will fail, meaning that the output over $\cT$ is very distinguishable from the output over $\cT'$.

In order the overcome the above obstacle, we present a simple and efficient algorithm $\AlgTestPartition$ that privately tests whether the input database $\cT$ is well partitioned or not. 
For pedagogical reasons, assume that $\AlgTestPartition$ gets as input two different databases of $k$-tuples $\cP$ and $\cQ$ with the promise that $\cP \cup \cQ$ is partitioned by far balls (later, given a multiset $\cT$, we apply $\AlgTestPartition$ with $\cP = \cQ = \cT$ and slightly pay a group privacy of size $2$).

Roughly, given $\cP$ and $\cQ$ as input, $\AlgTestPartition$ chooses a random $J=\tilde{O}\paren{1/\eps}$-size sub-multiset $\cR \subset \cP$. For each tuple $X = \set{\px_1,\ldots,\px_k} \in \cR$, it defines a set of far balls $\cB_X$ around the centers $\px_1,\ldots,\px_k$, privately checks (by adding Laplace noise of magnitude $\tilde{O}(J/\eps)$) if the value of $\ell_X = \min\set{\ell \colon \cQ\text{ is }\ell\text{-partitioned by }\cB_{\px}}$ does not exceed some (small) threshold (denote this decision bit by $\pass_X$), and then applies a private counting algorithm on all the $J$ decisions bits do decide whether the test succeed or failed. On success, it outputs one of the balls $\cB_X$ that had $\pass_X = 1$ (if $\pass_X = 0$ for all $X \in \cR$, it outputs empty balls). For proving utility, we note that when the test succeed, then with high probability there exists at least one $X \in \cR$ with $\pass_X = 1$, yielding that with high probability, any choice of such $X$ results with balls $B_X$ that $\ell$-partitions $\cQ$ with some small $\ell$.   
For proving that the test is private, consider two executions of  $\AlgTestPartition$ over neighboring databases $(\cP,\cQ)$ and $(\cP',\cQ')$. If $\cP \neq \cP'$ (and $\cQ = \cQ'$), then the privacy holds since at most one decision bit can be affected. If $\cQ \neq \cQ'$ (and $\cP = \cP'$), then each estimation of $\ell_X$ is $\eps/J$-differentially private, and by basic composition, all of them together are $\eps$-differentially private. Hence, the privacy of this case now holds by simple post-processing argument (we remark that, since $\cR$ is just a small subset of $\cP$, we actually reduce its value using sub-sampling argument, which is significant for practice).

In summary, our algorithm $\AlgPrivatekAverages$ on input $\cT$, executes $\AlgTestPartition$ with $\cP = \cQ = \cT$ which outputs $(\Status,\cB)$. If $\Status = ``Success"$, it executes $\AlgEstimateAverages(\cT,\cB)$ and outputs the resulting $k$ average estimations $\set{\tpa_1,\ldots,\tpa_k}$.

\subsection{Applications}

The importance of Algorithm $\AlgPrivatekAverages$ is that 
many classical clustering-based tasks can be privately solved by a reduction to the problem of estimating the averages of partitioned tuples. In this work, we chose to present two important use-cases: (1) Privately approximating the $k$-means under stability assumption, and (2) Privately learning the parameters of a mixture of well-separated Gaussians.

\paragraph{$k$-Means Clustering}

In $k$-means clustering, we are given a database $\cP$ of $n$ points in $\bbR^d$, and the goal is to identify a set $C$ of $k$ centers in $\bbR^d$ that approximately minimizes $\COST_{\cP}(C) = \sum_{\px \in \cP} \min_{\pc \in C}\norm{\px - \pc}^2$. We denote the lowest possible cost as $\OPT_{k}(\cP)$. Since the task of minimizing the $k$-means is NP-hard, the literature has focused on approximation algorithms, with the current (non-private) state-of-the-art achieving multiplicative error of $6.357$ \cite{ahmadian2019better}. 

Unlike in the non-private literature, it is known that every private algorithm for approximating the $k$-means cost must have an additive error (even computationally unbounded algorithms), which scales with the diameter of the input space. Hence, it is standard to assume an upper bound $\Lambda$ on the $\ell_2$ norm of all the input points $\cP$. Typically (though not always), one aims to minimize the multiplicative error while keeping the additive error at most polylogarithmic in the size of the database. The current state-of-the-art construction \Enote{ .....}


Now, suppose that we are given a database $\cP$ and a (non-private) $k$-means approximation algorithm $\cA$, such that the following stability property holds: When executing $\cA$ on a (large enough) random subset of $\cP$, then with high probability, the output $\pc_1,\ldots,\pc_k$ is very close to the optimal $k$-means $\pc_1^*,\ldots,\pc_k^*$ of $\cP$. Then using this property, we present our private algorithm $\AlgPrivatekMeans$ that easily approximate the $k$-means of $\cP$. Roughly, Algorithm $\AlgPrivatekMeans$ randomly partitions $\cP$ into $T$ subsets (for large enough $T \ll n$), and execute $\cA$ on each subset to obtain $k$ centers that are close to the optimal centers. Then, it collects all the $T$ $k$-tuples of centers and executes Algorithm $\AlgPrivatekAverages$ over them. The privacy follows since, for any given partition, two neighboring databases differ by only one of the subsets, and therefore, the resulting $T$ databases of $k$-tuples differ by at most $1$ tuple. For utility, note that by the stability property, w.h.p. all the $T$ $k$-tuples are partitioned by small $k$ balls around the optimal centers $\pc_1^*,\ldots,\pc_k^*$. Therefore, when executing $\AlgPrivatekAverages$, we expect to get $k$ centers $\tilde{\pc}_1,\ldots,\tilde{\pc}_k$ that are close to $\pc_1^*,\ldots,\pc_k^*$. 
We then show that by performing a single (noisy) Lloyd step that preserve privacy, the resulting $k$ centers well approximate the $k$-means cost. 

As a common concrete example of such $\cP$ and $\cA$ that satisfy the above stability property is when $\cA$ is
a good (non-private) approximation algorithm for $k$-means, and $\cP$ is well-separated for $k$-means, a notion that was first introduced in \citet{OstrovskyRSS12}. Formally, $\cP$ is called $\phi$-separated if $\OPT_{k}(\cP) \leq \phi^2 \OPT_{k-1}(\cP)$. It turns out that if $\cP$ is $\phi$-separated for $k$-means, and $\cA$ is a $\omega$-approximation algorithm for $k$-means, where $\phi^2(1 + \omega)$ is sufficiently small, then $\cP$ and $\cA$ satisfy our required stability property, yielding that our algorithm $\AlgPrivatekMeans$ succeed well over $\cP$ using $\cA$.

Following the work of \citet{OstrovskyRSS12}, several other works have related other notions of input-stability to clustering \cite{awasthi2010stability,awasthi2012center,balcan2009approximate,bilu2012stable,kumar2010clustering}. In addition, several works gave differentially-private algorithms for approximating the minimal Wasserstein distance (\cite{Wasserstein1969}) of  well-separated instances \cite{NRS07,wang2015differentially,huang2018optimal}, and in a recent work, \citet{ShechnerSS20} presented a differentially-private algorithm for approximating the $k$-means of such well-separated instances. The main result of \cite{ShechnerSS20} is a very simple transformation, called Private-Stable-$k$-Means, from any private $\omega$-approximation algorithm to a private $(1 + O(\phi^2))$-one, that works over $\phi$-separated database $\cP$ with sufficiently small $\phi^2(1 + \omega)$. In addition, they presented Algorithm SampleAggregate-$k$-means that transform any (non-private) $k$-means approximation algorithm $\cA$ into a private one, with very similar guarantees to our construction. Their algorithm also randomly partitions $\cP$ into $T$ subsets (for large enough $T \ll n$), and executes $\cA$ on each subset to obtain $k$ centers that are close to the optimal centers. But then, in order to privately estimating the averages of each cluster of the resulting partitioned multiset of $k$-tuples, their algorithm ignores the $k$-tuples structure and uses the (complicated) \cite{NS17_clustering}'s minimal enclosing ball  algorithm for determine the clusters, while our algorithm uses Algorithm $\AlgPrivatekAverages$ which is much more simpler and practical.

\paragraph{Learning Mixtures of Gaussians}
In the learning mixtures of Gaussians problem,  
we are given samples for a mixture $\cD$ of $k$ Gaussians over $\bbR^d$. The mixture is specified by $k$ components $G_1,\ldots,G_k$, where each $G_i$ is selected with probability $\omega_i \in [0,1]$ and is distributed as a Gaussian with mean $\mu_i \in \bbR^d$ and a covariance matrix $\Sigma_i \in \bbR^{d \times d}$.  The goal is to recover the parameters $\set{(\omega_i, \mu_i, \Sigma_i)}_{i=1}^k$ from the samples. That is, we would like to output a mixture $\hat{\cD} = \set{(\hat{\omega}_i, \hat{\mu}_i, \hat{\Sigma}_i)}_{i=1}^k$ such that $\cD$ and $\hat{\cD}$ are close in total-variation distance (which we denote by $\dTV(\cdot)$). We say that an algorithm $(\alpha,\beta)$-learns a mixture $\cD$ with sample complexity $n$, if given $n$ sample from $\cD$, with probability $1-\beta$ it outputs $\hat{\cD}$ such that $\dTV(\cD,\hat{\cD}) \leq \alpha$.

\Enote{cite non-private works here? I guess that there are plenty }

It turns out that in order to privately learn a mixture, even for $d = k=1$ (i.e., learn a single univariable Gaussian), we must assume a priori bounds $R, \sigma_{\min},\sigma_{\max}$ such that $\forall i:\text{}\norm{\mu_i} \leq R\text{ and } \sigma_{\min} \leq \norm{\Sigma_i} \leq \sigma_{\max}$ \cite{BS16,KV18} (where by $\norm{\cdot}$ we denote the standard $\ell_2$ norm of a matrix).

There are various private algorithms that learns the parameters of a single Gaussian \cite{KV18,KLSU19,CWZ19,BS19,KSU20,BDKU20}. Recently, \cite{KSSU19} presented a private algorithm for learning mixtures of well-separated (and bounded) Gaussians. Specifically, under the separation assumption $\forall i \neq j: \text{ } \norm{\mu_i - \mu_j} \geq \Omega\paren{\sqrt{k \log(nk/\beta)} + \sqrt{\frac1{w_i} + \frac1{w_j}}}\cdot \max \set{\norm{\Sigma_i}^{1/2}, \norm{\Sigma_j}^{1/2}}$,
they presented an $(\eps,\delta)$-differentially private algorithm that $(\alpha,\beta)$-learns such a separated and $(R,\sigma_{\min},\sigma_{\max})$-bounded mixture of $k$ Gaussians with sample complexity $n = \paren{\frac{d^2}{\alpha^2 w_{\min}} + \frac{d^2}{\eps \alpha w_{\min}} + \frac{\poly(k) d^{3/2}}{w_{\min} \eps}} \cdot \polylog\paren{\frac{d k R \sigma_{\max}}{\alpha \beta \eps \delta \sigma_{\min}}}$. We remark, however, that besides the result of \cite{BDKU20}, which is a practical algorithm for learning a single Gaussian, all the other results are primarily theoretical. \Enote{maybe here is the place to explain why the construction of  \cite{KSSU19}  is far from being practical}

In this work, we present a simple and practical algorithm $\AlgPrivatekGaussians$ that privately $(\alpha,\beta)$-learns the parameters of an $(R,\sigma_{\min},\sigma_{\max})$-bounded mixture of $k$ Gaussians under a weaker separation assumption from the one made by  \cite{KSSU19}. Roughly, Algorithm $\AlgPrivatekGaussians$ can use any non-private algorithm $\cA$ that learns the means of a Gaussian mixture. Given such $\cA$, it executes it $T$ times (for large enough $T$), each time with enough new samples for guaranteeing w.h.p. that the means estimations $\tilde{\mu}_1,\ldots,\tilde{\mu}_k$ of $\cA$ will be close enough to the actual means $\mu_1,\ldots,\mu_k$ (specifically, we need that $\forall i:\text{ }\norm{\mu_i - \tilde{\mu}_i} < \frac1{16} \cdot \min_{j \neq i} \norm{\mu_i - \mu_j}$), and then it executes our algorithm $\AlgPrivatekAverages$ on the resulting $T$-size database of all the $k$-tuples means estimations. Since w.h.p. all the $T$ $k$-tuples are partitioned by small balls around the actuals means $\mu_1,\ldots,\mu_k$,  then for large enough $T$ (but much smaller than $n$), we are guaranteed to obtain $k$ avegares estimations $\tilde{\pa}_1,\ldots,\tilde{\pa}_k$ that are very close to the actual means.

In the next step, Algorithm $\AlgPrivatekGaussians$ partitions a fresh set of $n$ samples 
according to $\tilde{\pa}_1,\ldots,\tilde{\pa}_k$, where each sample $\px$ belongs to the $i$'th set if $\pa_i$ is the closest point to it among $\tilde{\pa}_1,\ldots,\tilde{\pa}_k$. Using a standard projection argument, we show that if we assume a separation of the form $\forall i \neq j: \text{ } \norm{\mu_i - \mu_j} >  2\sqrt{2\log(\beta/n)} \cdot  \max \set{\norm{\Sigma_i}^{1/2}, \norm{\Sigma_j}^{1/2}}$ (which is independent of the dimension $d$), then with probability $1-\beta$, the above partition perfectly classifiy the points correctly (i.e., two points belong to the same set iff they both were sampled from the same Gaussian). Finally, we apply a private algorithm $\cA'$ for learning the parameters of each single Gaussian (e.g., \cite{KLSU19} or \cite{BDKU20}) in each of the $k$ sets in the partition.

Overall, our algorithm $\AlgPrivatekGaussians$ wraps (using algorithm $\AlgPrivatekAverages$) any given non-private algorithm $\cA$ that learns a mixture of Gaussians, along with any given private algorithm $\cA'$ that learns a single Gaussian, in order to learn the parameters of a mixture of separated (and bounded) Gaussians under a weaker separation assumption than the one used by \cite{KSSU19}. The construction uses only black-box accesses to $\cA$ and $\cA'$, and also can be parallelized very easily. Furthermore, we show that using an additional step of sub-sampling, our algorithm improves the sample complexity of \cite{KSSU19}. See \cref{sec:Gauss:comparison} for the full comparison.

\Enote{add discussion why JL is not usefull when we apply it directly on the learning mixtures of Gaussians problem}

\paragraph{Other Problems}

\Enote{TBD}

%
%
%

}
\section{Preliminaries}

\subsection{Notation}

In this work, a $k$-tuple 
$X = \set{\px_1,\ldots,\px_k}$
is an \emph{unordered} set of $k$ vectors $\px_i \in \bbR^d$.
For $\px \in  \bbR^d$, we denote by $\norm{\px}$ the $\ell_2$ norm of $\px$. For $\pc \in \bbR^d$ and $r > 0$, we denote $B(\pc,r) \eqdef \set{\px \in \bbR^d \colon \norm{\px - \pc} \leq r}$.
For a multiset $\cP \in (\bbR^d)^*$ we denote by $\Avg(\cP) \eqdef \frac1{\size{\cP}}\cdot \sum_{\px \in \cP} \px$ the average of all points in $\cP$. Throughout this work, a database $\cD$ is a multiset. For two multisets $\cD = \set{x_1,\ldots,x_n}$ and $\cD' = \set{x_1',\ldots,x_m'}$, we let $\cD \cup \cD'$ be the multiset $\set{x_1,\ldots,x_n,x_1',\ldots,x_m'}$. For a multiset $\cD = \set{x_1,\ldots,x_n}$ and a set $S$, we let $\cD \cap S$ be the multiset $\set{x_i}_{i \in \cI}$ where $\cI = \set{i \in [n] \colon x_i \in S}$.
All logarithms considered here are natural logarithms (i.e., in base $e$).

\subsection{Indistinguishability and  Differential Privacy}

\begin{definition}[Neighboring databases]\label{def:neighboring}
	Let $\cD = \set{x_1,\ldots,x_n}$ and $\cD' = \set{x_1',\ldots,x_n'}$ be two databases over a domain $\cX$. We say that $\cD$ and $\cD'$ are \textbf{neighboring} if there is exactly one index $i \in [n]$ with $x_i \neq x_i'$.
\end{definition}

\begin{definition}[$(\eps,\delta)$-indistinguishable]\label{def:indis}
	Two random variable $X,X'$ over a domain $\cX$ are called $(\eps,\delta)$-indistinguishable, iff for any event $T \subseteq \cX$, it holds that
	$\pr{X \in T} \leq e^{\eps} \cdot \pr{X' \in T} + \delta$. If $\delta = 0$, we say that $X$ and $X'$ are $\eps$-indistinguishable.
\end{definition}

\begin{definition}[$(\eps,\delta)$-differential privacy \cite{DworkMNS06}]\label{def:DP}
	An algorithm $\Alg$ is called $(\eps,\delta)$-differentially private, if for any two neighboring databases $\cD,\cD'$ it holds that $\Alg(\cD)$ and $\Alg(\cD')$ are $(\eps,\delta)$-indistinguishable. If $\delta = 0$ (i.e., pure privacy), we say that $\Alg$ is $\eps$-differentially private.
\end{definition}

\begin{lemma}[\cite{BS16}]\label{lem:indis}
	Two random variable $X,X'$ over a domain $\cX$ are $(\eps,\delta)$-indistinguishable, iff there exist events $E, E' \subseteq \cX$ with $\pr{X \in E},\pr{X' \in E'} \geq 1-\delta$ such that $X|_{E}$ and $X'|_{E'}$ are $\eps$-indistinguishable. 
\end{lemma}

\subsubsection{Basic Facts}

The following fact is a corollary of \cref{lem:indis}.

\begin{fact}\label{fact:indis-cor}
	Let  $X,X'$ be two random variables over a domain $\cX$, and let $E, E' \subseteq \cX$ be two events. If $X|_E$ and $X'|_{E'}$ are $(\eps,\delta_1)$-indistinguishable and $\pr{X \in E},\pr{X' \in E'} \geq 1 - \delta_2$, then $X$ and $X'$ are $(\eps,\delta_1 + \delta_2)$-indistinguishable.
\end{fact}
\begin{proof}
	Since $X|_E$ and $X'|_{E'}$ are $(\eps,\delta_1)$-indistinguishable, we deduce by \cref{lem:indis} that there exists events $F \subseteq E$ and $F' \subseteq E'$ with $\pr{X \in F \mid E}, \pr{X' \in F' \mid E'} \geq 1-\delta_1$ such that $X|_F$ and $X'|_{F'}$ are $(\eps,0)$-indistinguishable. In addition, note that
	\begin{align*}
		\pr{X \in F} = \pr{X \in E}\cdot \pr{X \in F \mid E} \geq (1-\delta_2)(1-\delta_1) \geq 1 - (\delta_1 + \delta_2).
	\end{align*}
	Similarly, it holds that $	\pr{X' \in F'} \geq 1 - (\delta_1 + \delta_2)$. Therefore, by applying the opposite direction of \cref{lem:indis} on the events $F$ and $F'$, we deduce that $X$ and $X'$ are $(\eps, \delta_1 + \delta_2)$-indistinguishable. 
\end{proof}

In addition, we use the following facts.

\begin{fact}\label{fact:conditioning}
    Let  $X,X'$ be two $\eps$-indistinguishable random variables over a domain $\cX$, and let $E, E' \subseteq \cX$ be two events with $\pr{X \in E},\pr{X' \in E'} \geq 1-\delta$. Then $X|_{E}$ and $X'|_{E'}$ are $(\eps - \ln(1-\delta), \: \frac{e^{\eps} \delta }{1-\delta})$-indistinguishable. 
\end{fact}
\begin{proof}
    Fix a subset $T \subseteq \cX$ and compute
    \begin{align*}
        \pr{X \in T \mid E}
        \leq \frac{\pr{X \in T }}{\pr{E}}
        \leq \frac{e^{\eps} \cdot \pr{X' \in T}}{1-\delta}
        &\leq \frac{e^{\eps}}{1-\delta} \cdot \paren{\pr{X' \in T \mid E'} + \pr{X' \notin E'}}\\
        &\leq e^{\eps -\ln(1-\delta)} \cdot \pr{X' \in T \mid E'} + \frac{e^{\eps} \delta}{1-\delta}
    \end{align*}
    where the last inequality holds since $\pr{X' \notin E'} \leq \delta$ by assumption.
\end{proof}

\begin{fact}\label{prop:similar-E}
	Let $X,X'$ be two random variables over a domain $\cX$. Assume there exist events $E,E' \subseteq  \cX$ such that the following holds:
	\begin{itemize}
		\item $\pr{X \in E} \in  e^{\pm \eps}\cdot \pr{X' \in E'}$, and
		
		\item $X|_{E}$ and $X'|_{E'}$ are $(\eps^*,\delta)$-indistinguishable, and
		
		\item $X|_{\neg E}$ and $X'|_{\neg E'}$ are $(\eps^*,\delta)$-indistinguishable.
	\end{itemize}
	Then $X,X'$ are $(\eps + \eps^*,\delta e^{\eps})$-indistinguishable.
\end{fact}
\begin{proof}
	Fix an event $T \subseteq \cX$ and compute 
	\begin{align*}
	\lefteqn{\pr{X \in T}
		= \pr{X \in T \mid E}\cdot \pr{X \in E} + \pr{X \in T \mid \neg E}\cdot \pr{X \notin E}}\\
	&\leq  \paren{e^{\eps^*}\cdot \pr{X' \in T \mid E'} + \delta}\cdot e^{\eps}\cdot \pr{X' \in E'} + \paren{e^{\eps^*}\cdot \pr{X' \in T \mid \neg E'} + \delta}\cdot e^{\eps}\cdot \pr{X' \notin E'}\\
	&= e^{\eps + \eps^*} \cdot \pr{X' \in T} + \delta e^{\eps}.
	\end{align*}
\end{proof}

\subsubsection{Group Privacy and Post-Processing}

\begin{fact}[Group Privacy]\label{fact:group-priv}
	If $\Alg$ is $(\eps,\delta)$-differentially private, then for all pairs of databases $\cS$ and $\cS'$ that differ by $k$ points it holds that $\Alg(\cS)$ and $\Alg(\cS')$ are $(k\eps, k e^{k\eps} \delta)$-indistinguishable.
\end{fact}

\begin{fact}[Post-processing]\label{fact:post-processing}
	If $\Alg$ is $(\eps,\delta)$-differentially private, then for every (randomized) function $F$ it holds that $F \circ \Alg$ is $(\eps,\delta)$-differentially private.
\end{fact}

\subsubsection{Composition}


\begin{theorem}[Basic composition, adaptive case \cite{DRV10}]\label{thm:composition1}
	If $\Alg_1$ and $\Alg_2$ satisfy $(\eps_1,\delta_1)$ and $(\eps_2,\delta_2)$ differential privacy (respectively), then any algorithm that adaptively uses $\Alg_1$ and $\Alg_2$ (and does not access the database otherwise) ensures $(\eps_1+\eps_2,\delta_1+\delta_2)$-differential privacy.
\end{theorem}


\begin{theorem}[Advanced composition~\cite{DRV10}]\label{thm:composition2}
	Let $0<\eps_0,\delta'\leq1$, and let $\delta_0\in[0,1]$. An algorithm that adaptively uses $k$ algorithms that preserve $(\eps_0,\delta_0)$-differential privacy (and does not access the database otherwise) ensures $(\eps,\delta)$-differential privacy, where $\eps=\sqrt{2k\ln(1/\delta')}\cdot\eps_0+2k\eps_0^2$ and $\delta = k\delta_0+\delta'$.
\end{theorem}

\subsubsection{The Laplace Mechanism}

\begin{definition}[Laplace distribution]
	For $\sigma \geq 0$,  let $\Lap(\sigma)$ be the Laplace distribution over $\bbR$ with probability density function $p(z) = \frac1{2 \sigma} \exp\paren{-\frac{\size{z}}{\sigma}}$.
\end{definition}

\begin{fact}\label{fact:laplace-concent}
	Let $\eps > 0$. If $X \sim \Lap(1/\eps)$ then for all $t > 0: \quad \pr{\size{X} > t/\eps} \leq e^{-t}$.
\end{fact}

\begin{definition}[Sensitivity]
	We say that a function $f \colon \cU^n  \rightarrow \bbR$ has sensitivity $\lambda$ if for all neigboring databases $\cS, \cS'$ it holds that $\size{f(\cS) - f(\cS')} \leq \lambda$.
\end{definition}

\begin{theorem}[The Laplace Mechanism \cite{DworkMNS06}]\label{fact:laplace}
	Let $\eps > 0$, and assume $f \colon \cU^n  \rightarrow \bbR$ has sensitivity $\lambda$. Then the mechanism that on input $\cS \in \cU^n$ outputs $f(\cS) + \Lap(\lambda/\eps)$ is $\eps$-differentially private.
\end{theorem}

\subsubsection{The Gaussian Mechanism}

\begin{definition}[Gaussian distribution]
	For $\mu \in \bbR$ and $\sigma \geq 0$,  let $\cN(\mu,\sigma^2)$ be the Gaussian distribution over $\bbR$ with probability density function $p(z) = \frac1{\sqrt{2\pi}} \exp\paren{-\frac{(z-\mu)^2}{2 \sigma^2}}$.
\end{definition}

\begin{fact}\label{fact:one-gaus-concent}
	Let $\pX = (X_1,\ldots,X_d)$, where the $X_i$'s are i.i.d. random variables, distributed according to $ \cN(0,\sigma^2)$. 
	Then for all
	$\beta > 0: \quad \pr{\norm{\pX} \leq \paren{\sqrt{d} + \sqrt{2 \log(1/\beta)}} \cdot \sigma} \geq 1-\beta$.
\end{fact}

\begin{definition}[$\ell_2$-sensitivity]
	We say that a function $f \colon \cU^n  \rightarrow \bbR^d$ has $\ell_2$-sensitivity $\lambda$ if for all neigboring databases $\cS, \cS'$ it holds that $\norm{f(\cS) - f(\cS')} \leq \lambda$.
\end{definition}

\begin{theorem}[The Gaussian Mechanism \cite{DKMMN06}]\label{fact:Gaus}
	Let $\eps,\delta \in (0,1)$, and assume $f \colon \cU^n  \rightarrow \bbR^d$ has $\ell_2$-sensitivity $\lambda$. Let $\sigma \geq \frac{\lambda}{\eps}\sqrt{2 \log(1.25/\delta)}$. Then the mechanism that on input $\cS \in \cU^n$ outputs $f(\cS) + \paren{\cN(0,\sigma^2)}^d$ is $(\eps,\delta)$-differentially private.
\end{theorem}

\begin{observation}\label{obs:Gaus-aver}
	For the case that $\cS \in (\bbR^d)^n$ and $f(\cS) = \Avg(\cS)$, if we are promised that each coordinate of the points is bounded by a segment of length $\Lambda$, then the sensitivity is bounded by $\lambda = \Lambda/n$, and therefore, by taking $\sigma = O(\frac{\Lambda}{\eps n} \sqrt{\log(1/\delta)})$ we get by \cref{fact:one-gaus-concent} that with probability $1-\beta$, the resulting point $\pz$ of the mechanism satisfies 
	$\norm{\pz - \Avg(\cS)} \leq \frac{\Lambda \sqrt{\log(1.25/\delta)}}{\eps n} \paren{\sqrt{d}  + \sqrt{2\log(1/\beta)}}$.
\end{observation}

\begin{remark}\label{remark:Gaus-add-del}
	\cref{fact:Gaus} guarantees differential-privacy whenever two neighboring databases have equal size. However, it can be easily extended to a more general case in which the privacy guarantee also holds in cases of addition and deletion of a point, with essentially the same noise magnitude (e.g., see Appendix A in \cite{NSV16}).
\end{remark}

The following proposition states the following: Assume that $\pX  \sim \mu + (\cN(0,\sigma^2))^d$ for some $\mu \in \bbR^d$,
and let $\py \in \bbR^d$ such that $\norm{\py - \mu}$ is ``large enough'' (i.e., larger than $\Omega\paren{\sigma \sqrt{\log(1/\beta)}}$). Then with probability $1-\beta$ (over $\pX$) it holds that $\norm{\pX - \mu} < \norm{\pX - \py}$. Note that such an argument is trivial when $\norm{\py - \mu}$ is at least $\Omega(\sigma \sqrt{d \log(1/\beta)})$, but here we are aiming for a distance that is independent of $d$. The proof of the proposition, which appears at \cref{missing-proof:thm:kGauss-utility} as a special case of \cref{prop:separation}, is based on a standard projection argument.


\begin{proposition}\label{prop:separation-spherical-case}
	Let $\pX  \sim \mu + (\cN(0,\sigma^2))^d$ and let $\py \in \bbR^d$ with $\norm{\py - \mu} > 2\sqrt{2\log\paren{\frac1{\beta}}} \cdot \sigma$. Then with probability $1-\beta$ (over the choice of $\pX$), it holds that $\norm{\pX - \mu} < \norm{\pX-\py}$.
\end{proposition}

\subsubsection{Estimating the Average of Points}\label{sec:prelim:est-aver}

As mentioned in \cref{obs:Gaus-aver}, the Gaussian mechanism (\cref{fact:Gaus}) allows for privately estimating the average of points in $B(\pt{0},\Lambda) \subseteq \bbR^d$ within $\ell_2$ error of $\approx \frac{\Lambda \sqrt{d}}{\eps n}$. In some cases, we could relax the dependency on $\Lambda$. For example, using the following proposition.

\def\propEstAvgInRd{
	Let $\eps \in (0,1)$, $d, \Lambda > 0$ and let $r_{\min} \in [0,\Lambda]$. There exists an efficient $(\eps,\delta)$-differentially private algorithm that takes an $n$-size database $\cS$ of points inside the ball $B(\pt{0},\Lambda)$ in $\bbR^d$ and satisfy the following utility guarantee: Assume that $n \geq \frac{32\sqrt{2d \log(2/\delta)}}{\eps} \log\paren{\frac{4d\Lambda}{r_{\min} \beta}} + 4$, and
	let $r >0$ be the minimal radius of a $d$-dimensional ball that contains all points in $\cS$. Then with probability $1-\beta$, the algorithm outputs $\hpa \in \bbR^d$ such that 
	\begin{align*}
		\norm{\hpa - \Avg(\cS)} \leq O\paren{\max\set{r,r_{\min}}\cdot  \frac{d \sqrt{\log(1/\delta)}}{\eps n} \paren{\sqrt{\log(d/\delta) \log(d/\beta)} + \log \paren{\frac{\Lambda d}{r_{\min} \beta}}}}.
	\end{align*}
	The algorithm runs in time $\tilde{O}(d n)$ (ignoring logarithmic factors).
}

\begin{proposition}[Estimating the Average of Bounded Points in $\bbR^d$]\label{prop:approx-aver-Rd}
	\propEstAvgInRd
\end{proposition}

\cref{prop:approx-aver-Rd} can be seen as a simplified variant of \cite{NSV16}'s private average algorithm. The main difference is that \cite{NSV16} first uses the Johnson Lindenstrauss (JL) transform \cite{JL84} to randomly embed the input points in $\bbR^{d'}$ for $d' \approx \log n$, and then estimates the average of the points in each axis of $\bbR^{d'}$. As a result, they manage to save a factor of $\sqrt{d}$ upon \cref{prop:approx-aver-Rd} (at the cost of paying a factor of $\log n$ instead). However, for simplifying the construction and the implementation, we chose to omit the JL transform step, and we directly estimate the average along each axis of $\bbR^{d}$. For completeness, we present the full details of \cref{prop:approx-aver-Rd} in \cref{sec:approx-aver}.
\Enote{maybe mention other methods like CoinPress, KV, etc, that are developed for Gaussians, but can be used here as well.}

\subsubsection{Sub-Sampling}

\begin{lemma}[\cite{BKN10,KLNRS11}]\label{lem:subsampling}
	Let $\cA$ be an $(\eps^*,\delta^*)$-differentially private algorithm operating on databases of size $m$. Fix $\eps \leq 1$, and denote $n = \frac{m}{\eps}(3 + \exp(\eps^*))$. Construct an algorithm $\cB$ that on an input database $\cD = (z_i)_{i=1}^n$, uniformly at random selects a subset $\cI \subseteq [n]$ of size $m$, and executes $\cA$ on the multiset $\cD_{\cI} = (z_i)_{i \in \cI}$. Then $\cB$ is $(\eps,\delta)$-differentially private, where $\delta = \frac{n}{4m}\cdot \delta^*$.
\end{lemma}


The following lemma states that switching between sampling
with replacement and without replacement has only a small effect on privacy.

\begin{lemma}[\cite{BNSV15}]\label{lem:DP-with-replacement}
	Fix $\eps \leq 1$ and let $\cA$ be an $(\eps,\delta)$-differentially private algorithm operating on
	databases of size $m$. For $n \geq 2m$, construct an algorithm $\cA'$ that on input a database $\cD$ of size $n$, subsamples (with replacement) $m$ rows from $\cD$, and runs $\cA$ on the result. Then $\cA'$ is $(\eps',\delta')$-differentially
	private for $\eps' = 6 \eps m/n$ and $\delta' = \exp\paren{6 \eps m/n} \cdot \frac{4m}{n}\cdot \delta$.
\end{lemma}

\remove{
\subsection{Preliminaries from Learning Theory}
A concept $c:\cX\rightarrow \{0,1\}$ is a predicate that labels {\em examples} taken from the domain $\cX$ by either 0 or 1.  A \emph{concept class} $\cC$ over $\cX$ is a set of concepts (predicates) mapping $\cX$ to $\{0,1\}$. 

\begin{definition}[VC dimension]
	Let $\cC$ be a concept class over $\cX$ and let $\cS = \set{x_1,\ldots,x_m} \subseteq \cX$. We say that $\cS$ is shattered by $\cC$ if $\size{\set{(c(x_1),\ldots,c(x_m)) \colon c \in \cC}} = 2^m$. The VC dimension of $\cC$, which is denoted by $\VC(\cC)$, is the maximal size of a set $\cS$ that is shattered by $\cC$.
\end{definition}

\begin{theorem}[VC bounds \cite{VC}]\label{thm:vc-bounds}
	Let $\cH = \set{h \colon \cX \rightarrow \zo}$ be a concept class, let $c \colon \cX \rightarrow \zo$ be a function,
	let $\mu$ be a distribution over $\cX$ and let $\cS$ be a set of $m$ i.i.d. samples from $\mu$.  
	Then with probability $1-\delta$ (over the choice of $\cS$), for any $h \in \cH$ with $\error_{\cS}(c,h) = \ppr{x \la \cS}{h(x) \neq c(x)} = 0$ (realizable case) it holds that
	\begin{align*}
		\error_{\mu}(c,h) \leq \frac{8 \VC(\cH) \log \frac{2m}{\VC(\cH)} + 4\log \frac{4}{\delta}}{m}
	\end{align*}
	where $\error_{\mu}(c,h) = \ppr{x \sim \mu}{h(x) \neq c(x)}$.
	In general (agnostic case), with probability $1-\delta$, for every $h \in \cH$ it holds that
	\begin{align*}
		\error_{\mu}(c,h) \leq \error_{\cS}(c,h) + \sqrt{\frac{\VC(\cH)\paren{\log \frac{2m}{\VC(\cH)}+1}+ \log \frac{4}{\delta}}{m}},
	\end{align*}
\end{theorem}

In the following, fix a parameter $d \in \bbN$ and let $\cball$ be the class of all $d$-dimensional balls in $\bbR^d$. Namely, $\cball = \set{h_{\pt{c},r} \colon \bbR^d \rightarrow \zo}_{\pc \in \bbR^d, r \geq 0}$, where $h_{\pt{c},r}(\px) = 1 \iff \px \in B(\pc,r)$.

\begin{fact}[\cite{Dud79}]\label{fact:vc-of-ball}
	The VC-dimension of $\cball$ is $d+1$.
\end{fact}

\begin{fact}[\cite{BlumerEhHaWa89}]\label{fact:k-fold}
	For any concept class $\cC$, the VC-dimension of $\cC^{k \cup}$, the $k$-fold union of $\cC$, is at most $3 \VC(\cC) k \log (3k)$, where $\cC^{k \cup} \eqdef \set{c_1 \cup \ldots \cup c_k \colon c_1,\ldots,c_k \in \cC}$.
\end{fact}

\cref{fact:vc-of-ball,fact:k-fold} imply the following corollary.

\begin{corollary}\label{cor:vc-of-k-ball}
	The VC-dimension of $\cballs$, the class of all $k$-fold union of balls in $\bbR^d$, is at most $3 (d+1) k \log (3k)$.
\end{corollary}
}

\remove{
\subsection{The Johnson Lindenstrauss transform}\label{sec:JL}

\begin{definition}[The JL random projection from $\bbR^d$ to $\bbR^m$]\label{def:JL}
	Let $f \colon \bbR^d \rightarrow \bbR^m$ be the projection that works as follows: Pick $m$ vectors $\pz_1,\ldots,\pz_m$ from a standard $d$-dimensional Gaussian distribution with density $p(\px) = \frac{1}{(2 \pi)^{d/2}} \exp(-\norm{\px}^2/2)$. For any vector $\px \in \bbR^d$, define $f(\px) = (\ip{\px,\pz_1}, \ldots, \ip{\px,\pz_m})$ (where $\ip{\px,\py} = \sum_{i=1}^d x_i y_i$).
\end{definition}

\begin{theorem}[\cite{JL84}]\label{thm:JL}
	Let $f \colon \bbR^d \rightarrow \bbR^m$ be the random projection from \cref{def:JL}, and let $n \in \bbN$, $\eps > 0$. Assuming that $m = \Omega\paren{\frac{1}{\eps^2} \cdot \log\paren{n/\beta}}$, then with probability $1-\beta$ it holds that
	\begin{align*}
		\forall \px,\py \in \cP: \quad \norm{f(\px)-f(\py)} \in (1 \pm \eps) \sqrt{m} \norm{\px - \py}
	\end{align*}
\end{theorem}
}

\subsection{Concentration Bounds}


\begin{fact}[Hoeffding's inequality]\label{fact:Hoeffding}
	Let $X_1,\ldots,X_n$ be independent random variables,  each $X_i$ is strictly bounded by the interval $[a_i,b_i]$, and let $\bar{X} = \frac1n\sum_{i=1}^n X_i$. Then for every $t \geq 0$:
	\begin{align*}
	\pr{\size{\bar{X} - \ex{\bar{X}}} \geq t} \leq 2\exp\paren{-\frac{2n^2 t^2}{\sum_{i=1}^n (b_i-a_i)^2}}
	\end{align*}
\end{fact}

\begin{fact}[{\cite[Theorem 5.3]{AminCO}}]\label{fact:binom_concentration}
	Let  $X \sim \Bin(n,p)$, then for all $t \geq 0$:
	\begin{enumerate}
		\item $\pr{X \geq \ex{X} + t} \leq \exp\paren{-\frac{t^2}{2\left(np + t/3\right)}}$.
		\item $\pr{X \leq \ex{X} - t} \leq \exp\paren{-\frac{t^2}{2np}}$.
	\end{enumerate}
\end{fact}


\section{$k$-Tuples Clustering}

We first introduce a new property of a collection of (unordered) $k$-tuples \Hnote{k-sets (shorter and more accurate) \Enote{Do we really want to change ``tuples'' to ``sets''?}} $\set{\px_1,\ldots,\px_k} \in (\bbR^d)^k$, which we call \emph{partitioned by $\Delta$-far balls}.

\begin{definition}[$\Delta$-far balls]\label{def:far-balls}
	A set of $k$ balls $\cB = \set{B_i = B(\pc_i,r_i)}_{i=1}^k$ over $\bbR^d$ is called \textbf{$\Delta$-far balls}, if for every $i \in [k]$ it holds that $\norm{\pc_i - \pc_j} \geq \Delta \cdot \max\set{r_i,r_j}$ (i.e., the balls are relatively far from each other). 
\end{definition}

\begin{definition}[partitioned by $\Delta$-far balls]\label{def:sep-balls}	
    \Hnote{add picture}
	A $k$-tuple $X \in (\bbR^d)^k$  is partitioned by a given set of $k$ $\Delta$-far balls $\cB = \set{B_1,\ldots,B_k}$, if for every $i \in [k]$ it holds that $\size{X \cap B_i} = 1$. A multiset of $k$-tuples $\cT \in ((\bbR^d)^k)^*$ is \textbf{partitioned by} $\cB$, if each $X \in \cT$ is partitioned by $\cB$. We say that  $\cT$ is \textbf{partitioned by $\Delta$-far balls} if such a set $\cB$ of $k$ $\Delta$-far balls exists. 
\end{definition}

In some cases we want to use a notion of a multiset $\cT$ \emph{almost} partitioned by $\Delta$-far balls. This is defined below using the additional parameter $\ell$.

\begin{definition}[$\ell$-nearly partitioned by $\Delta$-far balls]
	A multiset $\cT \in ((\bbR^d)^k)^*$ is \textbf{$\ell$-nearly partitioned by} a given set of $\Delta$-far balls $\cB = \set{B_1,\ldots,B_k}$,  if there are at most $\ell$ tuples in $\cT$ that are not partitioned by $\cB$. We say that $\cT$ is \textbf{$\ell$-nearly partitioned by $\Delta$-far balls} if such a set of $\Delta$-far balls $\cB = \set{B_1,\ldots,B_k}$ exists.
\end{definition}

For a database of $n$ $k$-tuples $\cT \in ((\bbR^d)^k)^*$, we let $\Points(\cT)$ be the collection of all the points in all the $k$-tuples in $\cT$.

\begin{definition}[The points in a collection of $k$-tuples]
    For $\cT = \set{\set{\px_{1,j}}_{j=1}^k,\ldots,\set{\px_{n,j}}_{j=1}^k} \in ((\bbR^d)^k)^n$, we define $\Points(\cT) = \set{\px_{i,j}}_{i\in [n], j \in [k]} \in (\bbR^d)^{kn}$.
\end{definition}

The following proposition states that if $\cT$ is partitioned by $\Delta$-far balls for $\Delta > 3$, then each choice of $\Delta$-far balls that partitions $\cT$ induces the same partition.

\begin{proposition}\label{prop:unique-partition}
    Let $\cT \in ((\bbR^d)^k)^*$ be a multiset that is partitioned by a set of $\Delta$-far balls $\cB = \set{B_1,\ldots,B_k}$ for $\Delta > 3$. Then for every $k$-tuple $X = \set{\px_1,\ldots,\px_k} \in \cT$ and for every $i \in [k]$, there exists a ball in $\cB$ (call it $B_i$), such that $\Points(\cT) \cap B_i = \set{\px \in \Points(\cT) \colon i = \argmin_{j \in [k]} \norm{\px - \px_j}}$.
\end{proposition}
\begin{proof}
    Let $X = \set{\px_1,\ldots,\px_k} \in \cT$, and for every $i \in [k]$ let $B_i = B(\pc_i,r_i) \in \cB$ be the ball that contains $\px_i$. We prove the proposition by showing that for every $i$ and every $\px \in \Points(\cT) \cap B_i$, it holds that $i = \argmin_{j \in [k]} \norm{\px - \px_j}$. 
    
    In the following, fix $\px \in \Points(\cT) \cap B_i$. On the one hand, since $\px \in B_i$, it holds that $\norm{\px - \px_i} \leq r_i$.  On the other hand, for any $j \neq i$ it holds that
    \begin{align*}
        \norm{\px - \px_j} \geq \norm{\pc_i - \pc_j} - \norm{\pc_i - \px} -  \norm{\pc_j - \px_j} > 3\max\set{r_i,r_j} - r_i - r_j \geq r_i,
    \end{align*}
    where the strict inequality holds since $B_i,B_j$ are $\Delta$-far balls for $\Delta > 3$. Namely, we deduce that $\norm{\px - \px_i} < \norm{\px - \px_j}$, as required. 
\end{proof}

\remove{
\ECnote{Perhaps there is even a simpler proof with $k$-tuples?  Isn't it the case that in any partition to far balls, each point from one tuple is in the same ball with the closest point to it from each other tuple?  Also, for this purpose it might be enough to use "4" instead of "6" in the far balls def (not sure what we need later).} 
\begin{proof}
    Fix a $k$-tuple $X = \set{\px_1,\ldots,\px_k} \in \cT$, and let $\cP_i = \set{\px \in \Points(\cT) \colon i = \argmin_{j}\norm{\px-\px_j}}$. 

	The proof trivially holds for $k=1$. Therefore, in the following we assume that $k \geq 2$. 
	Assume towards a contradiction that there exist $\px,\py \in \Points(\cT)$ such that $\px, \py \in B_i = B(\pc_i,r_i)$ (i.e., belong to the same ball in $\cB$) but $\px \in B_s' = B(\pc_s',r_s')$ and $\py \in B_{t}' = B(\pc_t',r_t')$ for $s \neq t$ (i.e., belong to different balls in $\cB'$). We next fix $\pz \in \paren{B_{s}' \cup B_{t}'} \cap \Points(\cT)$ and prove that $\pz \in B_i$ (see \cref{fig1}).
		
	\begin{figure}[htbp]
		\centerline{\includegraphics[scale=.2]{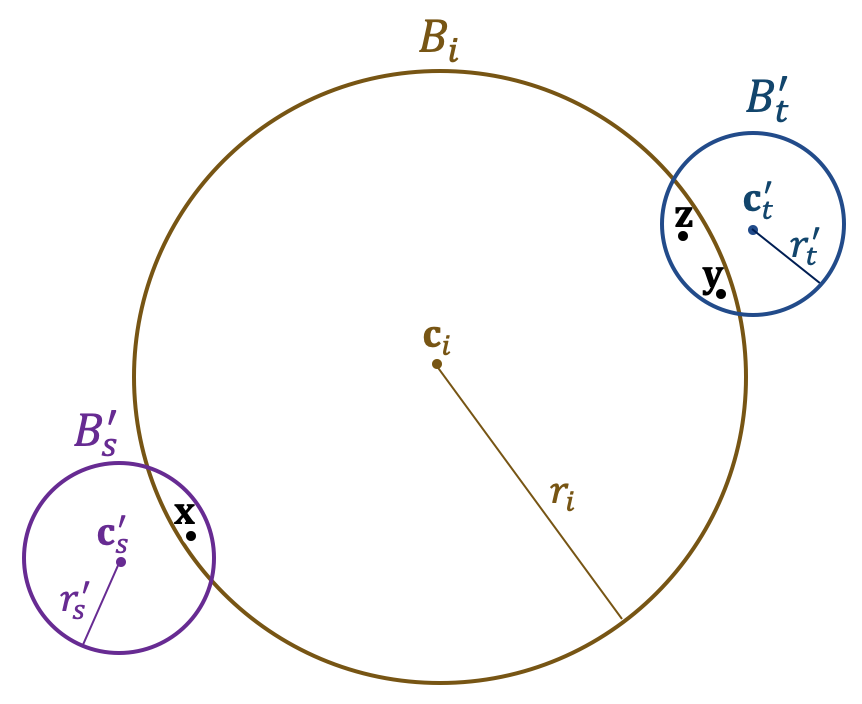}}
		\caption{An example of $\px,\py,\pz \in \cP$ such that $\px, \py \in B_i$,  $\px \in B_s'$,  $\py \in B_{t}'$ and $\pz \in B_{s}' \cup B_{t}'$.}
		\label{fig1}
	\end{figure}
	\noindent Since both $\cB$ and $\cB'$ are sets of far balls, the assumptions yield that 
	\begin{align*}
		4 \max\set{r_s',r_t'} < \norm{\pc_s' - \pc_t'} - \norm{\px - \pc_s'} -  \norm{\py - \pc_t'} < \norm{\px - \py} \leq 2 r_i.
	\end{align*}
	Therefore,  it holds that 
	\begin{align}\label{eq:pz-px-py}
		\min\set{\norm{\pz - \px}, \norm{\pz - \py}}  \leq 2\max\set{r_s',r_t'} < r_i,
	\end{align}
	which yields that
	\begin{align}\label{eq:pz-pc_i}
		\norm{\pz - \pc_i} 
		&\leq \min\set{\norm{\pz - \px} + \norm{\px - \pc_i}, \norm{\pz - \py} + \norm{\py - \pc_i}}\nonumber\\
		&\leq \min\set{\norm{\pz - \px}, \norm{\pz - \py}}  + r_i\nonumber\\
		&< 2r_i,
	\end{align}
	where the second inequality holds since $\px,\py \in B_i$ and the last one by \cref{eq:pz-px-py}. Note that \cref{eq:pz-pc_i} immediately yields that $\pz \in B_i \cap \Points(\cT)$ since $\cT$ is partitioned by $\cB$, and therefore, $\pz$ cannot belong to a different ball in $\cB$ since it is too close to $\pc_i$. 
	
	In summary,  we proved that
	\begin{align}\label{eq:contain-two-balls}
		\Points(\cT) \cap \paren{B_{s}' \cup B_{t}'} \subseteq  \Points(\cT) \cap B_i.
	\end{align}
	Now fix some $k$-tuple $\set{\px_1,\ldots,\px_k} \in \cT$. Since $\cB'$ partitions $\cT$, there exists a permutation $\pi$ of $[k]$ such that $\px_{\pi(s)} \in B_{s}'$ and $\px_{\pi(t)} \in B_{t}'$. Therefore, we deduce by \cref{eq:contain-two-balls} that both $\px_{\pi(s)}$ and $\px_{\pi(t)}$ belong to the same ball $B_i$ in $\cB$, in contradiction to the assumption that $\cB$ partitions $\cT$.
	
\end{proof}
}

We now formally define the partition of a database $\cT \in ((\bbR^d)^k)^*$ which is partitioned by $\Delta$-far balls for $\Delta > 3$.

\begin{definition}[$\Partition(\cT)$]\label{def:clusters-rel}
	Given a multiset $\cT \in ((\bbR^d)^k)^*$ which is partitioned by $\Delta$-far balls for $\Delta > 3$, 
	we define the partition of $\cT$, which we denote by $\Partition(\cT) = \set{\cP_1,\ldots,\cP_k}$,  by fixing an (arbitrary) $k$-tuple $X = \set{\px_1,\ldots,\px_k} \in \cT$ and setting $\cP_i = \set{\px \in \Points(\cT) \colon i = \argmin_{j \in [k]} \norm{\px - \px_j}}$.
\end{definition}

By \cref{prop:unique-partition}, this partition is well defined (i.e., is independent of the choice of the $k$-tuple $X$).

We now define the $k$-tuple clustering problem.

\begin{definition}[$k$-tuple clustering]\label{def:ktupleclustering}
The input to the problem is a database $\cT \in ((\bbR^d)^k)^n$ and a parameter $\Delta > 3$. The goal is to output a $k$-tuple $Y = \set{\py_1,\ldots,\py_k} \in (\bbR^d)^k$ such that the following holds: If $\cT$ is partitioned by $\Delta$-far balls, then for every $i \in [k]$, there exists a cluster in $\Partition(\cT)$ (call it $\cP_i$) such that $\cP_i = \set{\px \in \Points(\cT) \colon i = \argmin_{j \in [k]} \norm{\px - \py_j}}$. 
\end{definition}
Namely, in the $k$-tuple clustering problem, the goal is to output a $k$-tuple $Y$ that partitions $\cT$ correctly.
We remark that without privacy, the problem is completely trivial, since any $k$-tuple $X \in \cT$ is a good solution by definition. 
We also remark that for applications, we are also interested in the quality of the solution. Namely, how small is the distance between $\py_i$ and $\cP_i$, compared to the other clusters in $\Partition(\cT)$. This is captured in the following definition.

\begin{definition}[good and good-averages solutions]\label{def:gamma-good}
    \Hnote{Better to separate the "goodness" and the requirement that all radii are at least some minimum value}\Enote{I'm not sure it's better}
	Let $\cT \in ((\bbR^d)^k)^n$ and $\alpha, r_{\min} \geq 0$. We say that a $k$-tuple $Y = \set{\py_1,\ldots,\py_k} \in (\bbR^d)^k$ is an \emph{$(\alpha,r_{\min})$-good solution} for clustering $\cT$, if there exists a set of $\Delta$-far balls (for $\Delta > 3$) $\cB = \set{B_i = B(\pc_i,r_i)}_{i=1}^k$ that partitions $\cT$ such that for every $i \in [k]$ it holds that
	\begin{align*}
		\norm{\py_i - \pc_i} \leq \alpha \cdot \max\set{r_i,r_{\min}}
	\end{align*}
	If such $\cB$ exists with $r_i \geq r_{\min}$ for every $i \in [k]$, we say that $Y$ is an \emph{$\alpha$-good-average} solution.\Hnote{How does this relate to average ?
    it is also still related to $r_{\min}$}
	In a special case where such set of balls exists for $\pc_i = \Avg(\cP_i)$ where $\set{\cP_1,\ldots,\cP_k} = \Partition(\cP)$, we say that $Y$ is an \emph{$(\alpha,r_{\min})$-good-averages} solution.
\end{definition}
We remark that the additional parameter $r_{\min}$ is usually needed when considering differentially private algorithms, even for simpler problems like estimating the average of points (e.g., see \cref{prop:approx-aver-Rd}).
In addition, note that the quality of the solution is measured by how small is $\alpha$. 
The following claim states that if $\cT$ is partitioned by $\Delta$-far balls for $\Delta > 3$, then any $\alpha$-good solution according to \cref{def:gamma-good} for $\alpha < \Delta/2 - 1$ is also a $k$-tuples clustering solution according to \cref{def:ktupleclustering}.
\begin{claim}\label{claim:good-sol-is-valid}
	If $\cT \in ((\bbR^d)^k)^n$ is partitioned by $\Delta$-far balls for $\Delta > 3$, and $Y \in (\bbR^d)^k$ is an $\alpha$-good solution for clustering $\cT$ for $\alpha < \Delta/2 - 1$, then  for every $i \in [k]$, there exists a cluster in $\Partition(\cT)$ (call it $\cP_i$) such that $\cP_i = \set{\px \in \Points(\cT) \colon i = \argmin_{j \in [k]} \norm{\px - \py_j}}$.
\end{claim}
\begin{proof}
	Since $Y$ is an $\alpha$-good solution, there exists a set of $\Delta$-far balls $\cB = \set{B_i = B(\pc_i,r_i)}_{i=1}^k$ such that for every $i \in [k]$ it holds that
	\begin{align*}
	\norm{\py_i - \pc_i} \leq \alpha \cdot r_i.
	\end{align*}
	Now denote by $\cP_i$ the cluster $\Points(\cT) \cap B_i \in \Partition(\cT)$. 
	It remains to prove that for every $\px \in \cP_i$ and $j \neq i$ it holds that $\norm{\px - \py_i} < \norm{\px - \py_j}$. In the following, fix such $\px$ and $j$.
	
	On the one hand, it holds that
	\begin{align*}
	\norm{\px - \py_i} \leq \norm{\px - \pc_i} + \norm{\py_i - \pc_i}
	\leq (1 + \alpha) r_i.
	\end{align*}
	On the other hand, we have
	\begin{align*}
	\norm{\px - \py_j} \geq \norm{\pc_i - \pc_j} - \norm{\px-\pc_i} - \norm{\py_j - \pc_j}
	\geq \Delta \cdot \max\set{r_i,r_j} - r_i - \alpha \cdot r_j
	\geq (\Delta - 1 - \alpha) r_i.
	\end{align*}
	Hence, we conclude that $\norm{\px - \py_i} < \norm{\px - \py_j}$ whenever $\alpha < \Delta/2 - 1$, as required.
\end{proof}

For applications, we focus on a specific type of algorithms for the $k$-tuple clustering problems, that outputs a good-averages solution.  

\begin{definition}[averages-estimator for $k$-tuple clustering]\label{def:averages-estimator}
	Let $\Ac$ be an algorithm that gets as input a database in $((\bbR^d)^k)^*$. We say that $\Ac$ is an \emph{$(n,\alpha,r_{\min},\beta,\Delta,\Lambda)$-averages-estimator\Hnote{I do not think these carrying all these parameters all the time is good.
You can fix n to denote the size of the dataset and the diameter of the domain
maybe even $\Delta$. Do it once and don't carry them in the notation if the are fixed throughout} for $k$-tuple clustering}, if for every $\cT \in
	(B(\pt{0},\Lambda)^k)^n \subseteq ((\bbR^d)^k)^n$ that is partitioned by $\Delta$-far balls, $\Ac(\cT)$ outputs w.p. $1-\beta$ an $(\alpha,r_{\min})$-good-averages solution $Y \in (\bbR^d)^k$ for clustering $\cT$.
\end{definition}

Note that we allow the algorithm to handle only tuples over $B(\pt{0},\Lambda)$. If the algorithm can handle arbitrary tuples over $\bbR^d$, we omit the last parameter $\Lambda$.

\subsection{Additional Facts}

In this section we prove some facts about $\Delta$-far balls for $\Delta > 6$.

The following proposition states that if $\cB=\set{B_i}_{i=1}^k$ and $\cB' = \set{B_i'}_{i=1}^k$ are two sets of $\Delta$-far balls for $\Delta > 6$, and $\px$ is a point that belongs to either $B_i$ or $B_i'$ for some $i \in [k]$, such that $B_i$ and $B_i'$ intersect each other, then the center of $B_i$ is closest to $\px$ among all the centers of all the balls in $\cB$ (and the same holds w.r.t. $B_i'$ and $\cB'$).



\begin{proposition}\label{prop:close-sets-of-far-balls}
	Let $\cB = \set{B_i = B(\pc_i,r_i)}_{i=1}^k$ and $\cB' = \set{B_i' = B(\pc_i',r_i')}_{i=1}^k$ be two sets of $\Delta$-far balls for $\Delta > 6$ s.t. for every $i \in [k]$ it holds that $B_i \cap B_i' \neq \emptyset$.
	Then for every $i \in [k]$ and every $\px \in B_i \cup B_i'$, it holds that $i = \argmin_{j \in [k]} \norm{\px - \pc_j}$.
\end{proposition}
\begin{proof}
	Fix $i \in [k]$ and $\px \in B_i \cup B_i'$.
	If $\px \in B_i$, the proof trivially follows. Therefore, in the following we assume that $\px \notin B_i$, and therefore, $\px \in B_i'$.
	
	Note that on the one hand, it holds that
	 \begin{align}\label{eq:upbound-px-pc_i}
		\norm{\px - \pc_i}
		\leq \norm{\px - \pc_i'} + \norm{\pc_i' - \pc_i}
		\leq r_i' + (r_i + r_i')
		= 2r_i' + r_i
	\end{align}
	On the other hand, fix $j \neq i$, and note that 
	\begin{align}\label{eq:lowbound-px-pc_j-1}
		\norm{\px - \pc_j} 
		&\geq \norm{\pc_i - \pc_j} - \norm{\px - \pc_i}\\
		&> 6\max\set{r_i,r_j} - (2r_i' + r_i)\nonumber\\
		&\geq 5\max\set{r_i,r_j} - 2r_i',\nonumber
	\end{align}
	where the second inequality holds by \cref{eq:upbound-px-pc_i} along with the fact that $\cB$ are $\Delta$-far balls for $\Delta > 6$.
	Therefore, if $\max\set{r_i,r_j} \geq r_i'$, we deduce by \cref{eq:upbound-px-pc_i,eq:lowbound-px-pc_j-1} that $\norm{\px - \pc_i} < \norm{\px - \pc_j}$. Otherwise (i.e., $\max\set{r_i,r_j} < r_i'$), note that
	\begin{align}\label{eq:lowbound-px-pc_j-2}
		\norm{\px - \pc_j} 
		&\geq \norm{\pc_i' - \pc_j'} - \norm{\px - \pc_i'} - \norm{\pc_j' - \pc_j}\\
		&> 6\max\set{r_i',r_j'} - r_i' - (r_i' + r_i)\nonumber\\
		&> 3r_i'.\nonumber
	\end{align}
	Hence, we deduce by \cref{eq:upbound-px-pc_i,eq:lowbound-px-pc_j-2} that  $\norm{\px - \pc_i} < \norm{\px - \pc_j}$ also in this case, which concludes the proof of the proposition.
\end{proof}

We next prove that if $\cT$ is partitioned by $\Delta$-far balls for $\Delta > 6$, and $\cB$, a set of $\Delta$-far balls, partitions at least one tuple in $\cT$, then by partitioning the points in $\Points(\cT)$ w.r.t. the centers of the balls in $\cB$, we obtain exactly $\Partition(\cT)$.

\begin{proposition}\label{prop:from-almost-par-to-evenly-par}
	Let $\cT \in ((\bbR^d)^k)^n$ be a multiset that is partitioned by $\Delta$-far balls for $\Delta > 6$, let $\cB = \set{B_1,\ldots,B_k}$ be a set of $\Delta$-far balls that partitions at least one $k$-tuple of $\cT$, and let $\pc_1,\ldots,\pc_k$ be the centers of $B_1,\ldots,B_k$, respectively. In addition, for every $i \in [k]$  let $\cQ_i = \set{\px \in \Points(\cT) \colon i = \argmin_{j \in [k]} \norm{\px - \pc_j}}$. Then $\set{\cQ_1,\ldots,\cQ_k} = \Partition(\cT)$.
\end{proposition}

\begin{proof}

    Let $X = \set{\px_1,\ldots,\px_k} \in \cT$ be the assumed $k$-tuple that is partitioned by $\cB$, let $\cB^* = \set{B_1^*,\ldots,B_k^*}$ be a set of $\Delta$-far balls that partitions (all of) $\cT$, and assume w.l.o.g. that $\px_i \in B_i \cap B_i^*$ for every $i \in [k]$. \cref{prop:close-sets-of-far-balls} yields that for every $i \in [k]$ and $\px \in B_i^*$ it holds that $\px \in \cQ_i$, yielding that $B_i^* \cap \Points(\cT) \subseteq \cQ_i$. Since both sets $\set{B_i^*}_{i=1}^k$ and $\set{\cQ_i}_{i=1}^k$ consist of disjoints sets that cover all the points in $\Points(\cT)$, we conclude that $\set{\cQ_1,\ldots,\cQ_k} = \Partition(\cT)$.
\end{proof}

\remove{
In the following we describe a simple algorithm $\AlgIsPartitioned$ that takes a multiset $\cT \in ((\bbR^d)^k)^n$ and distinguishes between the case that $\cT$ is partitioned by \textbf{very} far balls, and the case that $\cT$ is \textbf{not} partitioned by $k$ far balls. In the former case it also outputs a set of $k$ far balls that partitions $\cT$. The algorithm is described in \cref{fig:IsPartitioned}.

\begin{figure}[thb!]
	\begin{center}
		\noindent\fbox{
			\parbox{.95\columnwidth}{
				\begin{center}{ \bf Algorithm $\AlgIsWeaklyPartitioned$}\end{center}
				\textbf{Input:} A multiset $\cT \in ((\bbR^d)^k)^*$.
				
				\begin{enumerate}
					\item Choose an arbitrary tuple $(\px_1,\ldots,\px_k) \in \cT$, and initialize $\cT_1,\ldots,\cT_k = \emptyset$.
					
					\item For $(\px_1',\ldots,\px_k') \in \cT$:\label{step:for-loop-tuple}
					
					\begin{enumerate}
						\item For $i=1$ to $k$:\label{step:for-loop-i}
						\begin{enumerate}
							\item Define $\pi(i) = \argmin_{j \in [k]} \norm{\px_i - \px_j'}$.
							\item $\cT_i = \cT_i \cup \set{\px_{\pi(i)}'}$ (a multiset union).
						\end{enumerate}
						\item If $\pi$ is not a permutation of $[k]$, output $(\text{``NO''},\perp)$.
					\end{enumerate}
				
					\item Output $(\text{``YES''},(\px_1,\ldots,\px_k))$.
					
%
%
%
					
				\end{enumerate}
		}}
	\end{center}
	\caption{A non-private algorithm that tests whether $\cT$ can be partitioned.\label{fig:IsPartitioned}}
\end{figure}

The following proposition summarizes the properties of $\AlgIsPartitioned$.

\def\propIsPartitioned{
	On inputs $\cT \in ((\bbR^d)^k)^n$, Algorithm $\AlgIsPartitioned$ satisfies the following guarantees:
	\begin{itemize}
		\item If $\cT$ is partitioned by \textbf{very} far balls, then the algorithm outputs $(\text{``YES''},\cB)$ where $\cB$ is a set of $k$ far balls that partitions $\cT$. 
		\item If $\cT$ is \textbf{not} partitioned by far balls, then the algorithm outputs $(\text{``NO''},\perp)$.
	\end{itemize}
	The algorithm runs in time $O(d n k^2)$.

}
\begin{proposition}\label{prop:IsPartitioned}
	\propIsPartitioned
\end{proposition}

\begin{proof}
	Consider an execution of $\AlgIsPartitioned(\cT)$.  If $\cT$ is \textbf{not} partitioned by far balls, then by definition, the output of the execution is $(\text{``NO''},\perp)$. In the rest of the analysis we assume that $\cP$ is partitioned by $k$ \textbf{very} far balls.
	
	By construction, the resulting multisets $\set{\cT_1,\ldots,\cT_k}$ in the for-loop (Step~\ref{step:for-loop-tuple}) are exactly $\Partition(\cT)$.
	Let $\cB = \set{B_i = B(\px_i, r_i)}_{i=1}^k$ be the set of $k$ balls from Step~\ref{step:resulting-balls}, and let $\cB^* = \set{B_i^* = B(\px_i^*, r_i^*)}_{i=1}^k$ be a set of $k$ \textbf{very} far balls that partitions $\cT$. The proof of the proposition now follows since for every $i \neq j$ it holds that
	\begin{align*}
		\norm{\px_i - \px_j}
		&\geq \norm{\px_i^* - \px_j^*} - \norm{\px_i - \px_i^*} - \norm{\px_j - \px_j^*}\\
		&> 14 \max\set{r_i^*,r_j^*} - r_i^* - r_j^*\\
		&\geq 12 \max\set{r_i^*,r_j^*}\\
		&\geq 6 \max\set{r_i,r_j},
	\end{align*} 
	where the last inequality holds since $r_i \leq 2 r_i^*$ and $r_j \leq 2 r_j^*$ by construction.
	
\end{proof}

}

\section{Our Algorithms}\label{sec:finding-averages}

In this section we present two $(\eps,\delta)$-differentially private algorithms for the $k$-tuple clustering problem:  $\AlgPrivatekAverages$ and $\AlgPrivateNoisykCenters$. Algorithm $\AlgPrivatekAverages$ attempts to solve the problem by determining the clusters in $\Partition(\cT)$ and then privately estimating the average of each cluster using the algorithm from \cref{prop:approx-aver-Rd}. Algorithm $\AlgPrivateNoisykCenters$, on the other hand, does not operate by averaging clusters. Instead, it first selects one of the input tuples $X \in \cT$ (in a special way), and then adds a (relatively small) Gaussian noise to this tuple.\footnote{We remind that all the tuples in this work are \emph{unordered}, and indeed the privacy analysis of our algorithms relies on it (i.e., the domain of outputs is all the unordered $k$-tuples, and $(\eps,\delta)$-indistinguishability holds for each subset of this domain).}

Both algorithms share the same first step, which is to call $\AlgTestPartition$ (\cref{alg:TestPar}) that privately decides whether $\cT$ is $\ell$-nearly partitioned by $\Delta$-far balls or not (for small $\ell$), and if so, determines (non-privately) a set of $\Delta$-far balls $\cB = \set{B_1,\ldots,B_k}$ that $\ell$-nearly partitions $\cT$. In \cref{sec:alg-test-close-tuples} we describe Algorithm $\AlgTestCloseTuples$, which is the main component of $\AlgTestPartition$. 
In \cref{sec:alg-test} we describe $\AlgTestPartition$ and state its properties. Then, in \cref{sec:alg-find-averages} we describe $\AlgPrivatekAverages$ and prove its guarantees, and in \cref{sec:alg-prac-find-averages} we describe $\AlgPrivateNoisykCenters$ and prove its guarantees.


\subsection{Algorithm $\AlgTestCloseTuples$}\label{sec:alg-test-close-tuples}

In this section we describe $\AlgTestCloseTuples$ (\cref{alg:TestCloseTuples}), which given two multisets of $k$-tuples $\cT_1$ and $\cT_2$, privately checks whether the tuples in $\cT_1$ are close to the tuples in $\cT_2$.

\begin{algorithm}[$\AlgTestCloseTuples$]\label{alg:TestCloseTuples}
	\item Input: Multisets  $\cT_1 \in ((\bbR^d)^k)^m$ and $\cT_2 \in ((\bbR^d)^k)^n$, a privacy parameter $\eps_1 \in (0,1]$ for $\cT_1$, a privacy parameters $\eps_2 \in (0,1]$ for $\cT_2$, a confidence parameter $\beta \in (0,1]$, and a separation parameter $\Delta > 6$.
	
	\item Operation:~
	\begin{enumerate}
		
		\item For each $X = \set{\px_1,\ldots,\px_k} \in \cT_1$:\label{step:loop-over-R}
		\begin{enumerate}
			\item Let $\cB_{X} = \set{B_i^{X} = B(\px_i, r_i^{X})}_{i=1}^k$, where $r_i^X = \frac1{\Delta}\cdot \min_{j \neq i} \norm{\px_i - \px_j}$.\label{step:B_x}
			\item Let $\ell_{X} = \size{\set{Y \in \cT_2 \colon Y\text{ is \textbf{not} partitioned by }\cB_{X}}}$.\label{step:l_x}
			
			\item Let $\hat{\ell}_{X} = \ell_{X} + \Lap\paren{\frac{m}{\eps_2}}$.\label{step:hl_x}
			
			\item Set $\pass_{X} = \begin{cases} 1 & \hat{\ell}_{X} \leq \frac{m}{\eps_2} \cdot \log\paren{\frac{m}{\beta}}\\ 0 &\text{otherwise} \end{cases}$.\label{step:pass}
		\end{enumerate}

		\item Let $s = \sum_{X \in \cT_1} \pass_{X}$ and compute $\hat{s} \la s + \Lap\paren{\frac1{\eps_1}}$.\label{step:s}
		\item If $\hat{s} <  m - \frac1{\eps_1} \log\paren{\frac1{\beta}}$, set $\Status
		= \text{"Failure"}$. Otherwise, set $\Status
		= \text{"Success"}$.\label{step:status}
		
		\item If $\Status
		= \text{"Success"}$ and $\pass_{X} = 1$ for at least one $X \in \cT_1$,
		let $X^*$ be the first tuple in $\cT_1$ with $\pass_{X^*} = 1$ and set $\cB = \cB_{X^*}$.
		Otherwise, set $\cB$ to be a set of $k$ empty balls.\label{step:x-star}
		
		\item Output $(\Status,\cB)$.
	\end{enumerate}
\end{algorithm}

\subsubsection{Properties of $\AlgTestCloseTuples$}

The properties of $\AlgTestCloseTuples$ are summarized by the following claims.

\begin{claim}[Correctness]\label{claim:correctness-closeTuples}
    Assume that $\cT = \cT_1 \cup \cT_2$ is partitioned by $(2\Delta+2)$-far balls. Then with probability $1-\beta$, when executing $\AlgTestCloseTuples$ on input $\cT_1,\cT_2,\eps_1,\eps_2,\beta$,$\Delta$, it outputs $(\text{``Success"},\cB)$, where $\cB$ is a set of $\Delta$-far balls that partitions $\cT$.
\end{claim}
\begin{proof}
	We first prove that for every $X \in \cT_1$, the set of balls $\cB_{X} = \set{B_i^{X} = B(\px_i, r_i^{X})}_{i=1}^k$ from Step~\ref{step:B_x} is a set of $\Delta$-far balls that partitions $\cT_2$. Fix $X = \set{\px_1,\ldots,\px_k} \in \cT_1$, let $\cB = \set{B_i = B(\pc_i, r_i)}_{i=1}^k$ be a set of $(2\Delta+2)$-far balls that partitions $\cT$ (such a set exists by assumption), and assume w.l.o.g. that $\forall i \in [k] \colon \px_i \in B_i$. 
	In addition, recall that $r_i^{X} = \frac1{\Delta}\cdot \min_{j \neq i} \norm{\px_i - \px_j}$ (Step~\ref{step:B_x}), and therefore, by definition it holds that $\cB_{X}$ is a set of $\Delta$-far balls. It is left to prove that it partitions $\cT$. Note that for every $i \neq j$ it holds that
	\begin{align*}
		\norm{\px_i - \px_j}
		&\geq \norm{\pc_i - \pc_j} - \norm{\px_i - \pc_i} - \norm{\px_j - \pc_j}\\
		&> (2\Delta+2) \cdot \max\set{r_i,r_j} - r_i - r_j\\
		&\geq 2\Delta \cdot \max\set{r_i,r_j}
	\end{align*} 
	Therefore, for every $i \in [k]$, $r_i^{X} = \frac1{\Delta}\cdot \min_{j \neq i} \norm{\px_i - \px_j} >  2\cdot r_i$. Since $\px_i \in B_i$, we conclude that $B_i \subseteq B_i^{X}$, which yields that $\cB_{X}$ partitions $\cT$.
	
	Therefore, for every $X = \set{\px_1,\ldots,\px_k} \in \cT_1$ it holds that $\ell_{X}$, the value from Step~\ref{step:l_x}, is $0$. Hence, by \cref{fact:laplace-concent} and the union bound, with probability $1-\frac{\beta}2$ it holds that $\forall X \in \cT_1:\text{ }\pass_{X} = 1$, which yields that $s = m$ (where $m = \size{\cT_1}$). When $s = m$, we obtain by \cref{fact:laplace-concent} that with probability $1- \frac{\beta}{2}$ it holds that  $\hat{s} \geq s - \frac1{\eps_1} \log(1/\beta) = m - \frac1{\eps_1} \log(1/\beta)$, i.e., $\Status = \text{``Success''}$. This concludes the proof of the claim.
\end{proof}

\begin{claim}[$\Status$ is $\eps_1$-DP w.r.t. $\cT_1$]\label{claim:status-is-private-T1}
	Let $\cT_1,\cT_1' \in ((\bbR^d)^k)^m$ be two neighboring databases, let $\cT_2 \in ((\bbR^d)^k)^n$, and consider two independent executions $\AlgTestCloseTuples(\cT_1, \cT_2)$ and $\AlgTestCloseTuples(\cT_1', \cT_2)$ (with the same parameters $\eps_1,\eps_2,\beta, \Delta$). Let $\Status$ and $\Status'$ be the status outcomes of the two executions (respectively). Then $\Status$ and $\Status'$ are $\eps_1$-indistinguishable.
\end{claim}
\begin{proof}
    Note that each $k$-tuple $X \in \cT_1$ can affect only the bit $\pass_{X}$. Therefore,
    by the properties of the Laplace mechanism (\cref{fact:laplace}) and post-processing (\cref{fact:post-processing}), it holds that $\Status$ and $\Status'$ are $\eps_1$-indistinguishable.
\end{proof}

\begin{claim}[$\Status$ is $\eps_2$-DP w.r.t. $\cT_2$]\label{claim:status-is-private-T2}
	Let $\cT_2,\cT_2' \in ((\bbR^d)^k)^n$ be two neighboring databases, let $\cT_1 \in ((\bbR^d)^k)^m$, and consider two independent executions $\AlgTestCloseTuples(\cT_1, \cT_2)$ and $\AlgTestCloseTuples(\cT_1, \cT_2')$ (with the same parameters $\eps_1,\eps_2,\beta$). Let $\Status$ and $\Status'$ be the status outcomes of the two executions (respectively). Then $\Status$ and $\Status'$ are $\eps_2$-indistinguishable.
\end{claim}
\begin{proof}
    For each $X \in \cT_1$, let $\ell_{X}, \pass_{X}$ and $\ell_{X}', \pass_{X}'$ be the values computed in the loop \ref{step:loop-over-R} in the two executions (respectively). Since $\size{\ell_{X} - \ell_{X}'} \leq 1$, we obtain by the properties of the Laplace mechanism, along with post-processing, that $\pass_{X}$ and $\pass_{X}'$ are $\frac{\eps_2}{m}$-indistinguishable. Hence, by basic composition (\cref{thm:composition1}) we deduce that $\set{\pass_{X}}_{X \in \cT_1}$ and $\set{\pass_{X}'}_{X \in \cT_1}$ are $\eps_2$-indistinguishable, and we conclude by post-processing that $\Status$ and $\Status'$ are $\eps_2$-indistinguishable.
\end{proof}

The following claim states that when $\AlgTestCloseTuples(\cT_1,\cT_2)$ outputs $(\text{``Success''},\cB)$, then with high probability, $\cT_2$ is almost partitioned by $\cB$.

\begin{claim}[On success, $\cB$ almost partitions $\cT_2$]\label{claim:main-T2}
    Let $\delta > 0$, let $\cT_1 \in ((\bbR^d)^k)^m$ and $\cT_1 \in ((\bbR^d)^k)^n$, and assume that $m > \frac1{\eps_1}\cdot \paren{2 \log(1/\delta) + \log(1/\beta)}$.
    Consider a random execution of \\$\AlgTestCloseTuples(\cT_1,\cT_2,\eps_1,\eps_2,\beta)$, and let $(\Status,\cB)$ be the outcome of the execution. Let $S$ be the event that $\Status = \text{"Success"}$,
    and let $E \subseteq S$ be the event that $\cT_2$ is $\ell$-nearly partitioned by $\cB$, where $\ell = \frac{m}{\eps_2} \cdot \log\paren{\frac{m}{\beta \delta}}$. Then the following holds:
    If $\pr{S} \geq \delta$, then $\pr{E \mid S} \geq 1 - \delta$.
\end{claim}

\begin{proof}
    Let $\set{\pass_{X}}_{X \in \cT_1}$ be the values from \cref{alg:TestPar} in the execution  $\AlgTestCloseTuples(\cT_1,\cT_2,\eps_1,\eps_2,\beta)$, and let $W$ be the event that there exists $X \in \cT_1$ with $\pass_{X} = 1$. Note that
    \begin{align*}
        \pr{\neg W \mid S}
        \leq \frac{\pr{S \mid \neg W}}{\pr{S}}
        \leq \frac{\pr{\Lap(1/\eps_1) > \frac{2}{\eps_1}\cdot \log\paren{\frac1{\delta}}}}{\delta}
        \leq \frac{\delta^2}{2 \delta}
        \leq \frac{\delta}2,
    \end{align*}
	where the second inequality holds since $\pr{S} \geq \delta$ and since $m - \frac1{\eps_1} \log\paren{\frac1{\beta}} > \frac{2}{\eps_1}\cdot \log\paren{\frac1{\delta}}$, and the third one holds by \cref{fact:laplace-concent}.
    Therefore, in the following we prove the claim by showing that
    \begin{align}\label{eq:E-mid-W-goal}
        \pr{E \mid W \land S} \geq 1 -\frac{\delta}2
    \end{align}
	Let $X^*$ be the tuple from Step~\ref{step:x-star} (it exists when $W \land S$ occurs), and recall that $\cB = \cB_{X^*}$ and that $\ell_{X^*}$ is the minimal value such that $\cT_2$ is $\ell_{X^*}$-nearly partitioned by $\cB$. Since $\pass_{X^*} = 1$, it holds that $\hat{\ell}_{X^*} = \ell_{X^*} + \Lap(m/\eps_2) \leq \frac{m}{\eps_2}\cdot \log\paren{\frac{m}{\beta}}$. \cref{eq:E-mid-W-goal} now follows by the following calculation.
	\begin{align*}
		\pr{E \mid W \land S}
		&= \pr{\ell_{X^*} > \frac{m}{\eps_2} \cdot \log\paren{\frac{m}{\beta \delta}} \mid \hat{\ell}_{X^*} \leq \frac{m}{\eps_2}\cdot \log\paren{\frac{m}{\beta}}}\\
		&\leq \pr{\Lap(m/\eps_2) < -\frac{m}{\eps_2}\cdot \log\paren{\frac{1}{\delta}}}\\
		&\leq \frac{\delta}{2},
	\end{align*}
	where the last inequality holds by \cref{fact:laplace-concent}.
\end{proof}

\subsection{Algorithm $\AlgTestPartition$}\label{sec:alg-test}

In this section we describe $\AlgTestPartition$ (\cref{alg:TestPar}) and state its properties.
In the following, we define $m$ and $\eps_1$ (functions of $n,\eps,\delta,\beta$) that are used by $\AlgTestPartition$. 

\begin{definition}\label{def:m}
	Let $m = m(n,\eps,\delta,\beta)$ be the smallest integer that satisfies $m > \frac1{\eps_1} \cdot \paren{2 \log(1/\delta) + \log(1/\beta)}$, where $\eps_1 = \log(\frac{\eps n}{2 m} - 3)$.
\end{definition}
The dependence between $m$ and $\eps_1$ for Algorithm $\AlgTestPartition$ is due to the choice of $\cT_1$ as an $m$-size random sample of $\cT$. A smaller $m$ allows for a larger value of $\eps_1$ for the same overall privacy, by a sub-sampling argument (e.g., \cref{lem:subsampling}).
We note that for $n \gg 1/\eps$ and $\beta,\delta \geq \frac{1}{\poly(n)}$, we have $\eps_1 = \Theta(\log n)$, which yields that $m = O(1)$. For smaller values of $\delta$, we obtain that $m = O\paren{\frac{\log(1/\delta)}{\log n}}$.

\begin{algorithm}[$\AlgTestPartition$]\label{alg:TestPar}
	\item Input: A multiset  $\cT \in ((\bbR^d)^k)^n$, privacy parameters $\eps,\delta \in (0,1]$, confidence parameter $\beta \in (0,1]$, and separation parameter $\Delta > 6$.
	
	\item Operation:~
	\begin{enumerate}
		
		\item Let $m$ and $\eps_1$ be the values from \cref{def:m} w.r.t. $n,\eps,\delta,\beta$, and let $\eps_2 = \eps/2$.
		
		\item Let $\cT_1$ be a uniform sample of $m$ $k$-tuples from $\cT$ (without replacement), and let $\cT_2 = \cT$.
		
		\item Output $(\Status,\cB) = \AlgTestCloseTuples(\cT_1,\cT_2,\eps_1,\eps_2,\beta,\Delta)$.
		
	\end{enumerate}
\end{algorithm}

%
%
%
%

\subsubsection{Properties of $\AlgTestPartition$}

The following claim is an immediate corollary of \cref{claim:correctness-closeTuples}
\begin{claim}[Correctness]\label{claim:correctness}
    Assume that $\cT$ is partitioned by $(2\Delta+2)$-far balls. Then with probability $1-\beta$, when executing $\AlgTestCloseTuples$ on input $\cT,\eps,\delta,\beta,\Delta$, it outputs $(\text{``Success"},\cB)$, where $\cB$ is a set of $\Delta$-far balls that partitions $\cT$.
\end{claim}

The following claim is a corollary of \cref{claim:status-is-private-T1,claim:status-is-private-T2}.

\begin{claim}[$\Status$ is private]\label{eq:status-is-private}
	Let $\cT$ and $\cT'$ be two neighboring databases, and consider two independent executions $\AlgTestPartition(\cT)$ and $\AlgTestPartition(\cT')$ (with the same parameters $\eps,\delta,\beta$). Let $\Status$ and $\Status'$ be the status outcomes of the two executions (respectively). Then $\Status$ and $\Status'$ are $\eps$-indistinguishable.
\end{claim}
\begin{proof}
As a first step, assume that we have two (different) copies of $\cT$, call them $\tilde{\cT}_1$ and $\tilde{\cT}_2$, where $\cT_1$ is chosen from the copy $\tilde{\cT}_1$, and $\cT_2$ is chosen from the copy $\tilde{\cT}_2$, and let $(\tilde{\cT}_1',\tilde{\cT}_2')$ be a neighboring database of $(\tilde{\cT}_1,\tilde{\cT}_2)$. If $\tilde{\cT}_2$ and $\tilde{\cT}_2'$ are neighboring (and $\tilde{\cT}_1 = \tilde{\cT}_1'$), we obtain by \cref{claim:status-is-private-T2} that $\Status$ and $\Status'$ are $\eps/2$-indistinguishable. Therefore, assume that $\tilde{\cT}_1$ and $\tilde{\cT}_1'$ are neighboring (and $\tilde{\cT}_2 = \tilde{\cT}_2'$). By \cref{claim:status-is-private-T1}, $\Status$ and $\Status'$ are $\eps_1$-indistinguishable if the resulting samples $\cT_1$ and $\cT_1'$ in the two executions are neighboring. Since $\cT_1$ is just an $m$-size sample from $\tilde{\cT}_1$, and since $\eps_1 = \log(\frac{\eps n}{2 m} - 3)$, we obtain by subsampling argument (\cref{lem:subsampling}) that $\Status$ and $\Status'$ are $\eps/2$-indistinguishable also in this case.

Finally, going back to our case where $\tilde{\cT}_1 = \tilde{\cT}_2 = \cT$, we deduce by the above analysis along with group privacy (of $2$) that $\Status$ and $\Status'$ are $\eps$-indistinguishable.
\end{proof}

The following claim is an immediate corollary of \cref{claim:main-T2}. It states that when the tests succeed, then w.h.p., $\cT$ is $\ell$-nearly partitioned by $\cB$, for the value of $\ell$ defined below.

\begin{definition}\label{def:ell}
    Let $\ell = \ell(n,\eps,\delta,\beta) = \frac{2 m}{\eps} \cdot \log\paren{\frac{m}{\beta \delta}}$, where $m = m(n,\eps,\delta,\beta)$ is the value from \cref{def:m}.
\end{definition}

We note that $\ell = O\paren{\frac{\log^2(1/\delta)}{\eps \log n}}$. When $\beta,\delta \geq 1/\poly(n)$, we have that $\ell = O\paren{\frac1{\eps} \log n}$.

\begin{claim}[On success, $\cB$ almost partitions $\cT$]\label{claim:main}
    Let $\cT \in ((\bbR^d)^k)^n$ and $\delta > 0$.
    Consider a random execution of $\AlgTestPartition(\cT,\eps,\delta,\beta,\Delta)$, and let $(\Status,\cB)$ be the outcome of the execution. Let $S$ be the event that $\Status = \text{"Success"}$,
    and let $E \subseteq S$ be the event that $\cT$ is $\ell$-nearly partitioned by $\cB$, where $\ell = \ell(n,\eps,\delta,\beta)$ is the value from \cref{def:ell}. Then the following holds:
    If $\pr{S} \geq \delta$, then $\pr{E \mid S} \geq 1 - \delta$.
\end{claim}
\begin{proof}
    Immediately holds by \cref{claim:main-T2} since $\ell = \frac{m}{\eps_2}\cdot \log\paren{\frac{m}{\beta \delta}}$, and since it holds that $m > \frac1{\eps_1}\cdot \paren{2 \log(1/\delta) + \log(1/\beta)}$ (by definition), as required by \cref{claim:main-T2}.
\end{proof}

Recall that Algorithm $\AlgTestPartition$ has two outputs: A bit $Status$ and a set of balls $\cB$.
As we stated in Claim~\ref{eq:status-is-private}, the bit $Status$ preserves privacy. The set of balls $\cB$, however, does {\em not}. Still, in the following sections we use Algorithm $\AlgTestPartition$ as a subroutine in our two main algorithms $\AlgPrivatekAverages$ and $\AlgPrivateNoisykCenters$. To argue about the privacy properties of these algorithms, we rely on the following key property of algorithm $\AlgTestPartition$.

\begin{claim}\label{claim:privacy-framework}
	Let $\Ac^*$ be an algorithm that gets as input a multiset $\cT \in ((\bbR^d)^k)^n$ and a set of balls $\cB = \set{B_1,\ldots,B_k}$, and let $\ell = \ell(n,\eps/2,\delta/4,\beta/2)$ be the value from \cref{def:ell}. Assume that $\Ac^*$
	has the property that for any neighboring multisets $\cT,\cT'$ and any sets of $\Delta$-far balls $\cB,\cB'$ that $\ell$-nearly partitions $\cT$ and $\cT'$ (respectively), it holds that $\Ac^*(\cT,\cB)$ and $\Ac^*(\cT',\cB')$ are $(\eps^*,\delta/4)$-indistinguishable. Let $\Ac$ be the algorithm that on input $\cT$, does the following steps: (1) Compute $(\Status,\cB)  = \AlgTestPartition\paren{\cT,\eps/2,\delta/4,\beta/2,\Delta}$, and (2) If $\Status = \text{``Failure''}$, output $\perp$ and abort, and otherwise output $\Ac^*(\cT,\cB)$.  Then $\Ac$ is $(\eps/2 + \eps^*,\delta)$-differentially private. 
\end{claim}
\begin{proof}
	Let $\cT$ and $\cT'$ be two neighboring multisets of size $n$. In the following we consider two independent executions: $\Ac(\cT)$ and $\Ac(\cT')$. In $\Ac(\cT)$, let  $O$ be the outcome, let $S,E$ be the events from \cref{claim:main} w.r.t. the execution of $\AlgTestPartition$ in step (1), and let $(\Status,\cB)$ be the resulting output of $\AlgTestPartition$. Similarly, let $O',S',E',\Status',\cB'$ be the events and random variables w.r.t. the execution $\Ac(\cT')$. Let $q = \pr{S}$ and $q' = \pr{S'}$. By \cref{eq:status-is-private} and by group privacy (\cref{fact:group-priv}), $\Status$ and $\Status'$ are  $\frac{\eps}{2}$-indistinguishable. Therefore, $q \in e^{\pm \eps/2} \cdot q'$. Recall that $\Ac$ outputs $\perp$ and aborts whenever $\Status = \text{"Failure"}$, and therefore, $\pr{O = \perp} = 1-q$ and  $\pr{O'  = \perp} = 1-q'$. If $q < \frac{\delta}2$ then $q' < e^{\eps/2}  \cdot \frac{\delta}2 \leq \delta$ (recall that $\eps\leq 1$), and therefore, $\pr{O = \perp}, \pr{O' = \perp} \geq 1 - \delta$. This means that $O$ and $O'$ are $(0,\delta)$-indistinguishable in the case that $q < \frac{\delta}2$ (by \cref{lem:indis}). Similarly, it holds that $O$ and $O'$ are $(0,\delta)$-indistinguishable when $q' < \frac{\delta}2$. 
	Hence, in the rest of the analysis we assume that $q,q' \geq \frac{\delta}2$.
	
	By \cref{prop:similar-E}, since $O|_{\neg S} \equiv O'|_{\neg S'}$ (both outcomes equal to $\perp$ when $\Status = \Status' = "\text{Failure}"$)
	and since $\pr{S} \in e^{\pm \eps/2} \cdot \pr{S'}$, it is enough to prove that $O|_S$ and $O'|_{S'}$ are $(\eps^*,\frac{\delta}2)$-indistinguishable.
	Furthermore, since $\pr{E \mid S},\pr{E' \mid S'} \geq 1 - \frac{\delta}{4}$ (by \cref{claim:main}), we deduce by \cref{fact:indis-cor} that it is enough to prove that $O|_{E}$ and $O'|_{E'}$ are $(\eps^*,\frac{\delta}4)$-indistinguishable, meaning that we only need to prove indistinguishability in the case that $\cT$ and $\cT'$ are $\ell$-nearly partitioned by $\cB$ and $\cB'$, respectively. The proof of the claim now follows since $\Ac^*(\cT,\cB)|_{E}$ and $\Ac^*(\cT',\cB')|_{E'}$ are $(\eps^*,\delta/4)$-indistinguishable by the assumption on the algorithm $\Ac^*$.
\end{proof}

\begin{remark}\label{remark:Test-runtime}
	Note that $\AlgTestPartition$ runs in time $O(m d k^2 n) = \tilde{O}(d k^2 n)$ since for each iteration $X \in \cT_1$ in $\AlgTestCloseTuples$, Step~\ref{step:B_x} takes $O(dk^2)$ time, and Step~\ref{step:l_x} takes $O(d k^2 n)$ times.
\end{remark}

\subsection{Algorithm $\AlgPrivatekAverages$}\label{sec:alg-find-averages}

In this section we describe and state the properties of $\AlgPrivatekAverages$ (\cref{alg:FindAverages}) which is our first algorithm for $k$-tuple clustering.

\begin{algorithm}[$\AlgPrivatekAverages$]\label{alg:FindAverages}
	\item Input: A multiset  $\cT \in \paren{B(0,\Lambda)^k}^n \subseteq ((\bbR^d)^k)^n$, privacy parameters $\eps,\delta \in (0,1]$,  a confidence parameter $\beta \in (0,1]$, and a lower bound on the radii $r_{\min} \in [0,\Lambda]$.
	
	\item Operation:~
	\begin{enumerate}
		
		
		\item Compute $(\Status,\cB = \set{B_1,\ldots,B_k}) = \AlgTestPartition(\cT,\eps/2,\delta/4,\beta/2, \Delta)$ for $\Delta = 7$.\label{step:call-testSeparation}
		
		\item If $\Status = \text{"Failure"}$, output $\perp$ and abort.

		\item Let $\pc_1,\ldots,\pc_k$ be the centers of $B_1,\ldots,B_k$ (respectively), and let
		
		$\cQ_i = \set{\px \in \Points(\cT) \colon i = \argmin_{j \in [k]} \norm{\px - \pc_j}}$.\label{step:compute-clusters}
		
		\item Let $\ell = \ell(n,\eps/2,\delta/4,\beta/2)$ be the value from \cref{def:ell}.
		
		\item For $i=1$ to $k$:
		
		\begin{enumerate}
			\item Compute a noisy average $\hat{\pa}_i$ of $\cQ_i$ by executing the algorithm from \cref{prop:approx-aver-Rd} with parameters $\Lambda, r_{\min}, \hat{\beta} = \frac{\beta}{2k}, \hat{\eps} = \frac{\eps}{4k(\ell+1)}, \hat{\delta} = \frac{\delta}{8 k \exp(\eps/2)(\ell + 1)}$.\label{step:computing-noisy-bound-aver}
		\end{enumerate}
		
		\item Output $\hat{A} = \set{\hat{\pa}_1,\ldots,\hat{\pa}_k}$.\label{step:kAverages:output}

	\end{enumerate}
\end{algorithm}

\subsubsection{Properties of $\AlgPrivatekAverages$}
The properties of $\AlgPrivatekAverages$ are given in the following theorems.

\begin{theorem}[Privacy of $\AlgPrivatekAverages$]\label{claim:privacy}
	Let $d, k, \Lambda > 0$, $r_{\min} \in [0,\Lambda]$, $\eps,\delta, \beta \in (0,1]$. Then for any integer $n \geq 2\cdot \ell(n,\eps/2,\delta/4,\beta/2) + 2$ (where $\ell$ is the function from \cref{def:ell}), 
	algorithm $\AlgPrivatekAverages(\cdot,\eps,\delta,\beta,r_{\min})$ is $(\eps,\delta)$-differentially private for databases $\cT \in (B(\pt{0},\Lambda)^k)^n \subseteq ((\bbR^d)^k)^n$.
\end{theorem}
\begin{proof}
	Let $\cT$ and $\cT'$ be two neighboring multisets of size $n$. In the following we consider two independent executions: $\AlgPrivatekAverages(\cT)$ and $\AlgPrivatekAverages(\cT')$ (both with the same parameters $r_{\min},\eps, \delta, \beta$). In $\AlgPrivatekAverages(\cT)$, let $O$ be the output, and let $\cB = \set{B_1,\ldots,B_k},\cQ_1,\ldots,\cQ_k$ be the values from the execution $\AlgPrivatekAverages(\cT)$. Similarly, we let
	$O', \cB' = \set{B_1',\ldots,B_k'},\cQ_1',\ldots,\cQ_k'$ be the these values w.r.t. the execution $\AlgPrivatekAverages(\cT')$. By \cref{claim:privacy-framework}, if we treat Step~\ref{step:compute-clusters} to \ref{step:kAverages:output} as algorithm $\Ac^*$ of the claim, it is enough to prove that $O = \hat{A}$ and $O'= \hat{A}'$ are $(\eps/2,\delta/4)$-indistinguishable only in the case that $\cT$ and $\cT'$ are $\ell$-nearly partitioned by $\cB$ and $\cB'$, respectively. In addition, note that since $\cT$ and $\cT'$ are neighboring, and since $n \geq 2\ell + 2$, there exists at least one $k$-tuple that is partitioned by both $\cB$ and $\cB'$, yielding that for each ball $B_i \in \cB$, there exists a balls in $\cB'$ (call it $B_i'$), such that $B_i \cap B_i' \neq \emptyset$. Since $\cB$ and $\cB'$ are sets of $\Delta$-far balls for $\Delta = 7$, \cref{prop:close-sets-of-far-balls} yields that for every $\px \in B_i$ (or $B_i'$), it holds that $i = \argmin_{j \in [k]}\norm{\px - \pc_j} = \argmin_{j \in [k]}\norm{\px - \pc_j'}$. Therefore, in the two executions, $\set{\cQ_1,\ldots,\cQ_k}$ and $\set{\cQ_1',\ldots,\cQ_k'}$ agree on all the points of all the common $(n-1)$ $k$-tuples of $\cT$ and $\cT'$ that are partitioned by $\cB$ or $\cB'$. Since there are at least $k \cdot (n-1-\ell)$ such points, we deduce that there are at most $k (\ell + 1)$ points that the partitions  $\set{\cQ_1,\ldots,\cQ_k}$ and $\set{\cQ_1',\ldots,\cQ_k'}$ disagree on.
	
	In the following, let $s_i$ be the number of points that the multisets $\cQ_i$ and $\cQ_i'$ differ by. Note that each point that the partitions disagree on contributes at most $1$ to at most two of the $s_i$'s. Hence, $\sum_{i=1}^k s_i \leq 2k(\ell + 1)$.
	
	By the privacy guarantee of \cref{prop:approx-aver-Rd} (see \cref{remark:bound-aver-add-del}) along with group privacy (\cref{fact:group-priv}),  for each $i \in [k]$, the resulting noisy averages $\hpa_i$ of the execution $\AlgPrivatekAverages(\cP)$, and the resulting $\hpa_i'$  of the execution $\AlgPrivatekAverages(\cP')$, which computed in Step~\ref{step:computing-noisy-bound-aver}, are 
	$(\frac{\eps s_i}{4k(\ell + 1)},\frac{\delta s_i}{8k(\ell + 1)})$-indistinguishable. Thus, by basic composition (\cref{thm:composition1}) we deduce that $\set{\hpa_1,\ldots,\hpa_k}$ and $\set{\hpa_1'\ldots,\hpa_k'}$ are $( \frac{\eps \sum_{i=1}^k s_i}{4k(\ell + 1)},\frac{\delta \sum_{i=1}^ks_i}{8 k(\ell + 1)}) = (\frac{\eps}{2}, \frac{\delta}{4})$-indistinguishable, as required.
	
\end{proof}

\begin{theorem}[Utility of $\AlgPrivatekAverages$]\label{claim:utility-kAverg}
	There exists a universal constant $\lambda > 0$ such that the following holds:
	Let $n,d, k \in \bbN$, $\Lambda > 0$, $r_{\min} \in [0,\Lambda]$, $\eps,\delta,\beta \in (0,1]$, let $\ell = \ell(n,\frac{\eps}2,\frac{\delta}4,\frac{\beta}2)$ be the value from \cref{def:ell}. If $n \geq \frac{\lambda k \ell \sqrt{d \log(d k \ell/\delta)} \log\paren{\frac{\Lambda d k}{r_{\min} \beta}}}{\eps}$, then algorithm $\AlgPrivatekAverages(\cdot, \eps, \delta, \beta, r_{\min})$ is an $(\alpha,r_{\min},\beta,\Delta=16,\Lambda)$-averages-estimator for $k$-tuple clustering (\cref{def:averages-estimator}), for
	\begin{align*}
		\alpha = \alpha(n,d,k,\eps,\delta,\beta,\Lambda,r_{\min}) \eqdef \frac{ \lambda d k \ell \sqrt{\log\paren{\frac{ k \ell}{\delta}}}}{\eps n} \paren{\sqrt{\log\paren{\frac{d k \ell}{\delta}}\log\paren{\frac{d k \ell}{\beta}}} + \log \paren{\frac{\Lambda d k}{r_{\min} \beta}}}.
	\end{align*}
\remove{
	and let $\cT \in \paren{B(0,\Lambda)^k}^n \subseteq ((\bbR^d)^k)^n$. 
	Assume that $n \geq \frac{\lambda k \ell \sqrt{d \log(d k \ell/\delta)} \log\paren{\frac{\Lambda d k}{r_{\min} \beta}}}{\eps}$
	and that 
	$\cT$ is partitioned by $\Delta$-far balls for $\Delta = 16$.
	Then w.p. $\geq 1-\beta$ over a random execution of $\AlgPrivatekAverages(\cT, \eps, \delta, \beta, r_{\min})$, the output $\hat{A} = \set{\hat{\pa}_1,\ldots,\hat{\pa}_k}$ of the execution is an \emph{$(\alpha,r_{\min})$-good-average} solution for clustering $\cT$ (according to \cref{def:gamma-good}), where 
	\begin{align*}
		\alpha = \alpha(n,d,k,\eps,\delta,\beta,\Lambda,r_{\min}) \eqdef \frac{ \lambda d k \ell \sqrt{\log\paren{\frac{ k \ell}{\delta}}}}{\eps n} \paren{\sqrt{\log\paren{\frac{d k \ell}{\delta}}\log\paren{\frac{d k \ell}{\beta}}} + \log \paren{\frac{\Lambda d k}{r_{\min} \beta}}}.
	\end{align*}
}
\end{theorem}

We remind that $\ell = O\paren{\frac{\log^2(1/\delta)}{\eps \log n}}$. Therefore, by ignoring $\polylog(n,d,k,\ell)$ factors, we obtain that for $n = \tilde{\Omega}\paren{\frac{d k \cdot  \log^{2.5}(1/\delta)\paren{\sqrt{\log(1/\delta)} +  \log\paren{\Lambda/r_{\min}}}}{\eps^2}}$, it holds that $\alpha = O(1)$.

\begin{proof}
	Let $\cT \in (\cB(\pt{0},\Lambda)^k)^n \subset  ((\bbR^d)^k)^n$ that is partitioned by $(\Delta=16)$-far balls. Consider
    a random execution of $\AlgPrivatekAverages(\cT,\eps,\delta,\beta,r_{\min})$, and let $\tilde{\beta} = \frac{\beta}{2}$ be the value from Step \ref{step:call-testSeparation}.
    Since $\cT$ is partitioned by $(2\cdot 7 + 2)$-far balls, \cref{claim:correctness} yields that with probability $1 - \beta/2$,
    the set $\cB = \set{B_1,\ldots,B_k}$ (computed in Step~\ref{step:call-testSeparation}) partitions $\cT$. In the following we assume that this event occurs. Let $\set{\cQ_1,\ldots,\cQ_k}$ be the clusters that were computed in Step~\ref{step:compute-clusters} of $\AlgPrivatekAverages$. By \cref{prop:from-almost-par-to-evenly-par}, it holds that $\set{\cQ_1,\ldots,\cQ_k} = \Partition(\cT)$. The proof now follows by the utility guarantee of \cref{prop:approx-aver-Rd} for each $i \in [k]$ with the parameters defined in Step~\ref{step:computing-noisy-bound-aver} of the algorithm.
\end{proof}

\begin{remark}[Run time of $\AlgPrivatekAverages$]\label{remark:kAver-runtime}
	Step~\ref{step:call-testSeparation} of $\AlgPrivatekAverages$ takes $\tilde{O}(dk^2 n)$ time (see \cref{remark:Test-runtime}). By \cref{prop:approx-aver-Rd}, the $k$ executions of Step~\ref{step:computing-noisy-bound-aver} takes time $\sum_{i=1}^k \tilde{O}(\size{\cT_i}) = \tilde{O}(dkn)$ (ignoring logarithmic factors). Overall, the running time of $\AlgPrivatekAverages$ is $\tilde{O}(dk^2n)$.
\end{remark}

\subsubsection{Reducing the dependency in the dimension $d$}\label{sec:reducing-by-JL}

When the dimension $d$ is large, algorithm $\AlgPrivatekAverages$ is an averages-estimator for $\alpha = \tilde{O}\paren{d/n}$
(ignoring $\poly(k,1/\eps)$ and $\polylog(n,\delta,\beta,\Lambda,1/r_{\min})$ factors). This means that if we aim for $\alpha = O(1)$, we must take $n = \tilde{\Omega}(d)$, and in some settings, such a dependency in the dimension might be expensive. Yet, we can easily reduce the $d$ into $\sqrt{d}$  by replacing in Step~\ref{step:computing-noisy-bound-aver} the average algorithm of  \cref{prop:approx-aver-Rd} by the average algorithm of  \cite{NSV16} that uses the JL transform for saving a factor of $\sqrt{d}$ (see the last paragraph in \cref{sec:prelim:est-aver} for more details).

\subsection{Algorithm $\AlgPrivateNoisykCenters$}\label{sec:alg-prac-find-averages}

In this section we describe Algorithm $\AlgPrivateNoisykCenters$ (\cref{alg:noisy-tuple}) which is our second algorithm for $k$-tuple clustering.

\begin{algorithm}[$\AlgPrivateNoisykCenters$]\label{alg:noisy-tuple}
	\item Input: A multiset  $\cT \in ((\bbR^d)^k)^n$, privacy parameters $\eps \in (0,1]$, $\delta \in (0,1/2]$,  confidence parameter $\beta \in (0,1]$, and a separation parameter $\Delta \gg 6$.
	
	\item Operation:~
	\begin{enumerate}
		
		
		\item Compute $(\Status,\cB = \set{B_1,\ldots,B_k}) = \AlgTestPartition(\cT,\eps/2,\delta/4,\beta/2, \Delta)$.\label{step:call-testSeparation-prac}
		
		\item If $\Status = \text{"Failure"}$, output $\perp$ and abort.

		\item Let $\pc_1,\ldots,\pc_k$ be the centers of $B_1,\ldots,B_k$ (respectively).\label{step:centers}
		
		\item For $i=1$ to $k$:\label{step:for-loop-prac}
		
		\begin{enumerate}
			\item Let $\lambda_i = \frac{2}{\Delta} (1+\gamma_i) \min_{j \neq i} \norm{\pc_i - \pc_j}$ where $\gamma_i = \frac4{\Delta-2} \cdot \paren{\Lap(4k/\eps) + \frac{4k}{\eps} \log(4k/\delta) + 1}$.
			
			\item Let $\hat{\pc}_i = \pc_i + (\cN(\pt{0}, \sigma_i^2))^d$, where $\sigma_i = \frac{4 k \lambda_i}{\eps} \sqrt{2 \log(10k/\delta)}$.
			
		\end{enumerate}
		
		\item Output $\hat{C} = \set{\hat{\pc}_1,\ldots,\hat{\pc}_k}$.\label{step:kAverages:output-prac}

	\end{enumerate}
\end{algorithm}

\subsubsection{Properties of $\AlgPrivateNoisykCenters$}
The properties of $\AlgPrivateNoisykCenters$ are given in the following theorems.

\begin{theorem}[Utility of $\AlgPrivateNoisykCenters$]\label{thm:util-NoisykCen}
    Let $d,k > 0$, $\eps,\beta,\delta \in (0,1]$ with $\delta < \beta$, let $\cT \in ((\bbR^d)^k)^n$, and assume that $\cT$ is partitioned by $(2\Delta+2)$-far balls, for
    $\Delta = \Omega\paren{\frac{k \log\paren{k/\delta} \sqrt{\log(k/\beta)}}{\eps}}$. Then when executing $\AlgPrivateNoisykCenters(\cT,\eps,\delta,\beta,\Delta)$, with probability $1-\beta$, the output $\hat{C} = \set{\hat{\pc}_1,\ldots,\hat{\pc}_k}$
    satisfy for every $i$ and $j \neq i$ that
    $\norm{\hat{\pc}_i - \pc_i} < \norm{\hat{\pc}_i - \pc_j}$, where $\pc_1,\ldots,\pc_n$ are the centers from Step~\ref{step:centers}.
\end{theorem}
We remark that the
 $k$ factor in the $\Delta$
in \cref{thm:util-NoisykCen},  comes from  applying basic composition (\cref{thm:composition1}) over the $k$ noisy centers $\hat{C}$. This however can be reduced to $\tilde{O}(\sqrt{k})$ factor by applying advanced composition (\cref{thm:composition2}).

\begin{proof}
    By the union bound on all the choices of $\gamma_i$, w.p. $1-\delta/8 \geq 1 - \beta/8$, for each $i \in [k]$ it holds that $\min_{j \neq i} \norm{\pc_i - \pc_j} \geq \Omega\paren{\sqrt{\log(k/\beta)}} \cdot \sigma_i$. Therefore, for every $i\neq j$ we can apply \cref{prop:separation-spherical-case} with $\mu = \pc_i$ and $\py = \pc_j$ to obtain that with proper choices of the constants in $\Delta$, with probability $1-\frac{\beta}{2k^2}$ it holds that $\norm{\hat{\pc}_i - \pc_i} < \norm{\hat{\pc}_i - \pc_j}$. By the union bound over all $i \neq j$ we deduce that with probability $1 - \beta/2$ this holds for every $i \neq j$, as required.
\end{proof}

\begin{theorem}[Privacy of $\AlgPrivateNoisykCenters$]\label{thm:priv-NoisykCen}
    Let $d,k > 0$, $\eps,\beta \in (0,1]$, $\delta \in (0,1/2]$, $\Delta > 6$. Then for any integer $n \geq 2\cdot \ell(n,\eps/2,\delta/4,\beta/2) + 2$ (where $\ell$ is the function from \cref{def:ell}),  $\AlgPrivateNoisykCenters(\cdot,\eps,\delta,\beta,\Delta)$ is $(\eps + \delta/4, \delta)$-differentially private for databases $\cT \in ((\bbR^d)^k)^n$.
\end{theorem}
\begin{proof}
    Let $\cT$ and $\cT'$ be two neighboring multisets of size $n$. In the following we consider two independent executions: $\AlgPrivateNoisykCenters(\cT)$ and $\AlgPrivateNoisykCenters(\cT')$ (both with the same parameters $r_{\min},\eps, \delta, \beta$). In $\AlgPrivateNoisykCenters(\cT)$, let $O$ be the output, and let $\cB = \set{B_i = B(\pc_i,r_i)}_{i=1}^k$ be the $\Delta$-far balls in the execution $\AlgPrivateNoisykCenters(\cT)$. Similarly, we let
    $O', \cB' = \set{B_i' = B(\pc_i',r_i')}_{i=1}^k$ be the these values w.r.t. the execution $\AlgPrivateNoisykCenters(\cT')$. By \cref{claim:privacy-framework}, it is enough to prove that the resulting outputs $O = \tilde{C}$ and $O' = \tilde{C}'$ of Steps \ref{step:centers} to \ref{step:kAverages:output-prac}
    are $(\eps/2 + \delta/4,\delta/4)$-indistinguishable only in the case that $\cT$ and $\cT'$ are $\ell$-nearly partitioned by $\cB$ and $\cB'$, respectively. Since $2 \ell \leq n-2$ and since $\cT$ and $\cT'$ are neighboring, there must exists a $k$-tuple $X = \set{\px_1,\ldots,\px_k} \in \cT$ that is partitioned by both $\cB$ and $\cB'$. In the rest of the analysis we assume (w.l.o.g.) that $\px_i \in B_i \cap B_i'$ for every $i \in [k]$.
    
    In the following, we prove that for every $i \in [k]$ it holds that $\min_{j \neq i} \norm{\pc_i - \pc_j}$ is close to $\min_{j \neq i} \norm{\pc_i' - \pc_j'}$. For every $i\neq j$ it holds that
    \begin{align*}
        \norm{\pc_i - \pc_j} 
		&\leq \norm{\pc_i - \pc_i'} +  \norm{\pc_j - \pc_j'} +  \norm{\pc_i' - \pc_j'}\\
		&\leq \norm{\pc_i - \px_i} + \norm{\pc_i' - \px_i} + \norm{\pc_j - \px_j} + \norm{\pc_j' - \px_j} +  \norm{\pc_i' - \pc_j'}\\
		&\leq r_i + r_i' + r_j + r_j' +  \norm{\pc_i' - \pc_j'}\\
		&\leq \frac2{\Delta} \norm{\pc_i - \pc_j}  + \frac2{\Delta} \norm{\pc_i' - \pc_j'} + \norm{\pc_i' - \pc_j'}.
    \end{align*}
    Therefore,
    \begin{align*}
		\norm{\pc_i - \pc_j}  \leq \frac{\Delta+2}{\Delta-2} \norm{\pc_i' - \pc_j'} = \paren{1 + \frac4{\Delta-2}} \norm{\pc_i' - \pc_j'}.
	\end{align*}
	Now let $i \in [k]$, and let $s = \argmin_{j \neq i} \norm{\pc_i - \pc_j}$ and $t = \argmin_{j \neq i} \norm{\pc_i' - \pc_j'}$. We deduce that
	\begin{align}\label{eq:compare-minimums}
		\min_{j \neq i} \norm{\pc_i - \pc_j} = \norm{\pc_i - \pc_s}  \leq \norm{\pc_i - \pc_t} \leq \paren{1 + \frac4{\Delta-2}} \norm{\pc_i' - \pc_t'} = \paren{1 + \frac4{\Delta-2}} \cdot \min_{j \neq i} \norm{\pc_i' - \pc_j'}
	\end{align}
	Similarly, it holds that $\min_{j \neq i} \norm{\pc_i' - \pc_j'} \leq \paren{1 + \frac4{\Delta-2}} \cdot \min_{j \neq i} \norm{\pc_i - \pc_j}$.
	Therefore, by the properties of the laplace mechanism, we deduce that for each $i$, the values of $\lambda_i$ and $\lambda_i'$ are $\frac{\eps}{4k}$-indistinguishable, and by basic composition we deduce that $\set{\lambda_i}_{i=1}^k$ and $\set{\lambda_i'}_{i=1}^k$ are all together $\eps/4$-indistinguishable. 
	
	In the following, let $L$ be the event that $\forall i \in [k]:\text{ }\gamma_i \geq \frac{4}{\Delta-2}$, and $L'$ be the event that $\forall i \in [k]:\text{ }\gamma_i' \geq \frac{4}{\Delta-2}$. By \cref{fact:laplace-concent} and the union bound, it holds that $\pr{L},\pr{L'} \leq \delta/8$. Therefore, by \cref{fact:indis-cor}, it is enough to prove that $\tilde{C}|_{L}$ and $\tilde{C}'|_{L'}$ 
	are $(\eps/2 + \delta/4,\delta/8)$-indistinguishable.
	
	First, by \cref{fact:conditioning}, we deduce that $\set{\lambda_i}_{i=1}^k|_{L}$ and $\set{\lambda_i'}_{i=1}^k|_{L'}$ are $(\eps/4 + \delta/4)$-indistinguishable\Enote{Missing here a delta error of $e^{\eps} \delta/(8-\delta) \leq \delta/4$.}. We now continue with the analysis assuming that $\lambda_i = \lambda_i'$ for all $i \in [k]$. Note that for every $i$ it holds that 
	\begin{align*}
	    \norm{\pc_i - \pc_i'} 
	    &\leq \norm{\pc_i - \px_i} + \norm{\pc_i' - \px_i}\\
	    &\leq r_i + r_i'\\
	    &\leq \frac1{\Delta} \cdot \paren{\min_{j \neq i} \norm{\pc_i - \pc_j} + \min_{j \neq i} \norm{\pc_i' - \pc_j'}}\\
	    &\leq \lambda_i,
	\end{align*}
	where the last inequality holds by \cref{eq:compare-minimums} (assuming that $L$ occurs). Therefore, by the properties of the Gaussian Mechanism (\cref{fact:Gaus}), we deduce that for each $i$, $\hpc_i$ and $\hpc_i'$ are $(\frac{\eps}{4k},\frac{\delta}{8k})$-indistinguishable, and by basic composition (\cref{thm:composition1}) we deduce that $\hat{C}$ and $\hat{C}'$ are $(\frac{\eps}{4},\frac{\delta}{8})$-indistinguishable (assuming that $\lambda_i = \lambda_i'$ for all $i \in [k]$). Finally, recall that $\set{\lambda_i}_{i=1}^k|_{L}$ and $\set{\lambda_i'}_{i=1}^k|_{L'}$ are $(\eps/4 + \delta/4)$-indistinguishable, and therefore, we conclude by adaptive composition (\cref{thm:composition1}) that $\hat{C}$ and $\hat{C}'$ are $(\eps/2 + \delta/4,\delta/8)$-indistinguishable.
\end{proof}

\begin{remark}[Run time of $\AlgPrivateNoisykCenters$]\label{remark:kNoisyCenters-runtime}
	Step~\ref{step:call-testSeparation-prac} of $\AlgPrivateNoisykCenters$ takes $\tilde{O}(dk^2 n)$ time (see \cref{remark:Test-runtime}). The foor-loop in Step~\ref{step:for-loop-prac} only takes $O(d k n)$ time. Overall, the running time of $\AlgPrivateNoisykCenters$ is $\tilde{O}(dk^2n)$.
\end{remark}
\section{$k$-Means Clustering}\label{sec:kMeans}

In this section we present our first application of $k$-tuples clustering, which is an $(\eps,\delta)$-differentially private $k$-means approximation algorithm $\AlgPrivatekMeans$ with utility guarantee that holds when the input is stable in the sense that we will define.
We first start with preliminaries about $k$-means clustering.

\subsection{Preliminaries}\label{sec:kMeans:Preliminaries}

For a multiset $\cP \in (\bbR^d)^*$ and a $k$-tuple of centers $C = \set{\pc_1,\ldots,\pc_k} \in (\bbR^d)^k$, we denote $\COST_{\cP}(C) \eqdef \sum_{\px \in \cP} \min_{i \in [k]}\norm{\px - \pc_i}^2$ and denote $\OPT_k(\cP) \eqdef \min_{C \in (\bbR^d)^k} \COST_{\cP}(C)$.

The following proposition states that given a multiset $\cP \in (\bbR^d)^n$ and an $\omega$-approximation algorithm $\Ac$ for $k$-means, then when sampling $s$ i.i.d.\ points from $\cP$ and executing $\Ac$ on these points, then with probability $1-\beta$ we obtain $k$ centers with cost $\approx \omega \OPT_{k}(\cP)$ (up to a small additive error that depends on $s$ and $\beta$). The proof appears at \cref{missing-proof:cost-of-sample-is-good}.

\def\propCostOfSampleIsGood{
	Let $\cP$ be a multiset of $n$ points in $\cB(\pt{0},\Lambda) \subseteq \bbR^d$ and let $\cA$ be an $\omega$-approximation algorithm for $k$-means. Consider the following random execution: (1) Construct a multiset $\cS$ of $s$ i.i.d.\ samples from $\cP$, (2) Compute $\tilde{C} = \cA(\cS,k)$. Then for every $\beta > 0$, with probability $1-\beta$ it holds that
	\begin{align*}
	\COST_{\cP}(\tilde{C}) \leq \omega\cdot \OPT_k(\cP) + \xi(s,\beta),
	\end{align*}
	where $\xi(s,\beta) \eqdef 4\paren{M(s,\beta)+  \sqrt{M(s,\beta) \cdot \omega \OPT_k(\cP)}}$ for $M(s,\beta) \eqdef 25  \Lambda^2 k d \log \paren{\frac{2nd}{\beta}} \cdot \frac{n}{s}$.
}

\begin{proposition}\label{prop:cost-of-sample-is-good}
	\propCostOfSampleIsGood
\end{proposition}

The following proposition states that given a multiset of points $\cP$ and given two $k$-tuples of centers $C = \set{\pc_1,\ldots,\pc_k}$ and $C' = \set{\pc_{1}', \ldots, \pc_{k}'}$ such that each $\pc_i'$ is relatively close to a unique  center $\pc_i$ in $C$, then by clustering the points according to $C'$ and performing a single Lloyd step, we get new centers whose $k$-means cost is almost bounded by $\COST_{\cP}(C)$. The proof appears at \cref{missing-proof:close-centers-have-similar-cost}.

\def\propCloseCentersHaveSimilarCost{
	Let $k \in \N $ and $\gamma \in [0,1/8]$.
	Let $\cP \in (\bbR^d)^*$, let $C = \set{\pc_1,\ldots,\pc_k}$ and $C' = \set{\pc_{1}', \ldots, \pc_{k}'}$ be two $k$-tuples of centers in $\bbR^d$ such that
	for every $i \in [k]$ it holds that $\norm{\pc_i' - \pc_i} \leq \gamma\cdot D_i$, where $D_i = \min_{j \neq i}\norm{\pc_i - \pc_j}$. In addition, for every $i \in [k]$ let $\cP_i$ be the multiset of all points in $\cP$ that $\pc_i'$ is closest to them in $C'$.
	Then
	$$\sum_{i=1}^k \OPT_1(\cP_i)  \leq (1 + 32\gamma) \COST_{\cP}(C).$$
}

\begin{proposition}\label{prop:close-centers-have-similar-cost}
	\propCloseCentersHaveSimilarCost
\end{proposition}

\subsection{Private $k$-Means Under Stability Assumption}\label{sec:kMeans:Alg}

In this section we describe our private algorithm $\AlgPrivatekMeans$ for approximation the  $k$-means when the input is stable in the sense that we will define next.
The idea is the following: Fix a database $\cP \in (\bbR^d)^n$, parameters $s,t \in \bbN$ and a (non-private) $k$-means approximation algorithm $\Ac$. Now execute $\Ac$ on $s$ i.i.d.\ samples from $\cP$, and repeat this process $t$ times. Consider the event (over this process) that all the $t$ sets of $k$ centers are almost located at the same positions. 
More formally, consider a random execution of $\AlgGenerateCenters^{\Ac}(\cP,k,s,t)$ (\cref{alg:generateCenters}). For a $k$-tuple of centers $C = \set{\pc_1,\ldots,\pc_k} \in (\bbR^d)^k$ and a small stability parameter $\gamma > 0$ (say, $\gamma = 0.01$), let $E_C^{\gamma}$ be the event that is defined below.

\begin{definition}[Event $E_{C}^{\gamma}$ over a random sampling of $\AlgGenerateCenters$]\label{def:event-ECgamma}
	Let $E_{C}^{\gamma}$ be the event that for every $j \in [t]$ and $i \in [k]$, there exists a center in $\tilde{C}_j$ (call it $\tpc_i^j$)  such that $\norm{\tpc_i^j - \pc_i} \leq \gamma\cdot D_i$, where $D_i = \min_{j \neq i}\norm{\pc_i - \pc_j}$.
\end{definition}

Namely, event $E_{C}^{\gamma}$ implies that the output $\tilde{\cC} \in ((\bbR^d)^k)^t$ of $\AlgGenerateCenters$ is partitioned by $\Delta$-far balls for $\Delta = 1/\gamma$, where $\Partition(\tilde{\cC})$ (according to \cref{def:clusters-rel}) is exactly $\set{\set{\tpc_1^j}_{j=1}^t, \ldots, \set{\tpc_k^j}_{j=1}^t}$ (i.e., for each $i \in [k]$, the centers $\set{\tpc_i^j}_{j=1}^t$ are very close to each other, compared to the distance from the other centers). 

In this section, we describe our general $(\eps,\delta)$-differentially private algorithm $\AlgPrivatekMeans$, that uses oracle accesses to a non-private $k$-means algorithm $\Ac$ and a private good-average $k$-tuple clustering algorithm $\Bc$, such that the following utility guarantee it achieved: 
For any $k$-centers $C$ and a small enough $\gamma$, 
when the event $E_{C}^{\gamma}$ occurs over $\AlgGenerateCenters^{\Ac}(\cP,k,s,t)$, then with probability $1-\beta$, algorithm $\AlgPrivatekMeans$ outputs $\hat{C} = \set{\hat{\pc}_1,\ldots,\hat{\pc}_k}$ with $\COST_{\cP}(\hat{C}) \leq (1 + O(\gamma)) \COST_{\cP}(C)$ (plus some small additive error). $\AlgPrivatekMeans$ is described in \cref{alg:kMeans} and its properties are proven in \cref{sec:kMeans:properties}. In \cref{sec:application} we show that a variant of the separation assumption in \cite{OstrovskyRSS12} implies that event $E_{C^*}^{\gamma}$ holds with high probability, where $C^*$ are the optimal $k$ means for $\cP$.

%
%
%

\begin{algorithm}[$\AlgGenerateCenters$]\label{alg:generateCenters}
	\item Input: A multiset  $\cP$ of points in $B(\pt{0},\Lambda) \subseteq \bbR^d$ and parameters $k, s,t \in \bbN$.
	
	\item Oracle: A (non-private) $k$-means algorithm $\Ac$.
	
	\item Operation:~
	\begin{enumerate}
		\item For each $j \in [t]$:
		
		\begin{enumerate}
			\item Let $\cS_j$ be a database containing $s$ i.i.d.\ samples from $\cP$ (with replacement).\label{step:sample}
			
			\item Compute the $k$-tuple of centers $\tilde{C}_j = \Ac(\cS_j)$.\label{step:non-priv-centers}
		\end{enumerate}
		
		\item Output $\cT = \set{\tilde{C}_1,\ldots,\tilde{C}_t}$.\label{step:new-multisets}
	\end{enumerate}
\end{algorithm}

%
%
%
%
%

\begin{algorithm}[$\AlgPrivatekMeans$]\label{alg:kMeans}
	\item Input: A multiset  $\cP$ of $n$ points in $B(\pt{0},\Lambda) \subseteq \bbR^d$, parameters $k, s, t \in \bbN$, privacy parameters $\eps,\delta \in (0,1]$, confidence parameter $\beta \in (0,1]$, and a stability parameter $\gamma > 0$. 
	
	\item Oracle: A (non-private) $k$-means algorithm $\Ac$, a private average algorithm $\Ac'$,
	and a private $k$-tuple clustering algorithm $\Bc$.
	
	\item Operation:~
	\begin{enumerate}
		
		\item Compute $\cT = \AlgGenerateCenters^{\Ac}(\cP,k,s,t)$.\label{step:gen}

		\item Compute $\set{\hat{\pa}_1,\ldots,\hat{\pa}_k} = \Bc(\cT)$.\label{step:the-aver-estim}
		
		\item For each $i \in [k]:$
		\begin{itemize}
			\item Let $\cP_i$ be the points in $\cP$ that $\hpa_i$ is the closest point to them among $\set{\hat{\pa}_1,\ldots,\hat{\pa}_k}$. \label{step:compute-clusters-kmeans}
			
			\item Compute $\hpc_i = \Ac'(\cP_i)$.\label{step:add-gaus-noise}
			
		\end{itemize}

		\item Output $\set{\hat{\pc}_1,\ldots,\hat{\pc}_k}$.
		
		
		%
		%
		%
		%
		%
	\end{enumerate}
\end{algorithm}

\subsection{Properties of $\AlgPrivatekMeans$}\label{sec:kMeans:properties}

The following theorem captures the privacy guarantee of $\AlgPrivatekMeans$.

\begin{theorem}[Privacy of $\AlgPrivatekMeans$]\label{claim:kMeans-privacy}
	 Let $n,s,t,d,k \in \bbN$, $\Lambda > 0$, $\beta,\eps,\delta, \gamma \in (0,1]$,
	 let $\Ac$ be an (arbitrary) algorithm that outputs $k$ centers in $B(\pt{0},\Lambda) \subset \bbR^d$, let $\Ac'$ be an $(\frac{\eps}{12},\frac{\delta}{8 e^{\eps}})$-DP algorithm for databases over $B(\pt{0},\Lambda)$, and let 
	 $\Bc$ be an $\left(\frac{\eps}{6},\frac{\delta}{4 e^{\eps}}\right)$-DP algorithm for databases $\cT \in (B(\pt{0},\Lambda)^k)^t$ (i.e., of size $t$). If $n \geq 2st$, then algorithm $\AlgPrivatekMeans^{\Ac, \Ac', \Bc}(\cdot, k,s,t, \eps,\delta,\beta,\gamma)$
	 is $(\eps,\delta)$-differentially private for databases $\cP$ over $B(\pt{0},\Lambda) \subset \bbR^d$.
\end{theorem}
\begin{proof}
	The proof builds on the fact that switching between sampling
	with replacement and without replacement has only a small effect on
	the privacy, as stated in  \cref{lem:DP-with-replacement}.
	
	Consider a different variant $\tAlgGenerateCenters$ of the procedure $\AlgGenerateCenters$, in which the sampling of the $\approx s \cdot t$ points in all the iterations of Step~\ref{step:sample} is done without replacement, and consider a variant $\tAlgPrivatekMeans$ of $\AlgPrivatekMeans$ in which it executes $\tAlgGenerateCenters$ in Step~\ref{step:gen} rather than $\AlgGenerateCenters$.
	Let $\cP = \set{\px_1,\ldots,\px_n}$ and $\cP' = \set{\px_1',\ldots,\px_n'}$ be two neighboring databases of points. In the following we consider two independent executions $\tAlgPrivatekMeans(\cP)$ and $\tAlgPrivatekMeans(\cP')$ (both with the same parameters $k,\eps,\delta,\beta$ and oracles $\Ac,\Bc$). For $j \in [t]$ let $\cJ_j \subseteq [n]$ be the $s$ chosen indices of the points $\cS_j$ in Step~\ref{step:sample} of $\tAlgGenerateCenters$ (i.e., $\cS_j = \set{\px_i}_{i \in \cJ_j}$), and let $\cJ_j'$ be the same indices in the execution $\tAlgPrivatekMeans(\cP')$. Since $\cJ_j$ and $\cJ_j'$ only depend on $n$ and not on the content of $\cP$ and $\cP'$, it is enough to prove that the output of both executions is $(\eps,\delta)$-indistinguishable conditioned on the event that $\cJ_j = \cJ_j'$ for every $j \in [t]$. In the following, we assume that this event occurs.
	
	Since $\cP$ and $\cP'$ are neighboring, there exists at most one index $j \in [t]$ such that $\cS_j$ of the execution $\tAlgPrivatekMeans(\cP)$ is different than the corresponding set in $\tAlgPrivatekMeans(\cP')$, and therefore, the outputs $\hat{\cC}$ of $\tAlgGenerateCenters$ are different by at most one $k$-tuple. 
	Therefore, by the assumption over algorithm $\Bc$, we deduce that the outcome of Step~\ref{step:the-aver-estim} is $(\frac{\eps}{6}, \frac{\delta}{4 e^{\eps}})$-differentially private. 
	
	In the following, we prove that for any fixing of $k$ averages $\tpa_1,\ldots,\tpa_k$, Step~\ref{step:add-gaus-noise} is $(\frac{\eps}{6}, \frac{\delta}{4 e^{\eps}})$-differentially private. Given that, we deduce that $\tAlgPrivatekMeans$ is $(\frac{\eps}{3}, \frac{\delta}{2 e^{\eps}})$-differentially private by (adaptive) composition of Steps~\ref{step:the-aver-estim} and \ref{step:add-gaus-noise} (\cref{thm:composition1}). Hence, we conclude that the original algorithm $\AlgPrivatekMeans$, that chooses the points with replacement, is $(\eps,\delta)$-differentially private by applying \cref{lem:DP-with-replacement} with $m = s t \leq \frac{n}2, \frac{\eps}{3}, \frac{\delta}{2 e^{\eps}}$.
	
	It is left to prove the privacy guarantee of Step~\ref{step:add-gaus-noise}. For that, fix $k$ averages $\hat{A}=\set{\tpa_1,\ldots,\tpa_k}$, let $\cP_1,\ldots,\cP_k$ be the $k$ multisets in Step~\ref{step:add-gaus-noise} w.r.t $\cP$ and $\hat{A}$, and  let $\cP_1',\ldots,\cP_k'$ be the same multisets w.r.t $\cP'$ and $\hat{A}$. Since $\cP$ and $\cP'$ are neighboring, there exist at most two indices $i \in [k]$ such that $\cP_i \neq \cP_i'$, and for each one of them, $\cP_i$ and $\cP_i'$ are neighboring. Therefore,  by the privacy guarantee of $\Ac'$ along with basic composition (\cref{thm:composition1}), Step~\ref{step:add-gaus-noise}  is $(2 \cdot \frac{\eps}{12}, 2\cdot \frac{\delta}{8 e^{\eps}})$-differentially private, as required.

\end{proof}

The following theorem, which captures the utility guarantee of $\AlgPrivatekMeans$, states that when event $E^{\gamma}_C$ (\cref{def:event-ECgamma}) occurs by $\AlgGenerateCenters$ in Step~\ref{step:gen}, then with probability at least $1-\beta$, the output $\hat{C} = \set{\hat{\pc}_1,\ldots,\hat{\pc}_k}$ has $\COST_{\cP}(\hat{C}) \leq (1 + O(\gamma)) \COST_{\cP}(C) + O\paren{\zeta^2 k}$, where $\zeta/m$ is the additive error of $\Ac'$ over database of size $m$ with confidence $1- O(\beta/k)$.

We remark that by setting $\Ac'$ as the Gaussian mechanism (\cref{fact:Gaus}) with privacy parameter $\paren{O(\eps),O(\delta)}$, we obtain that $\zeta = O\paren{\frac{\Lambda \sqrt{\log(1/\delta)}}{\eps} \paren{\sqrt{d} + \sqrt{\log(k/\beta)}}}$.

   
\begin{theorem}[Utility of $\AlgPrivatekMeans$]\label{claim:kMeans-utility}
	Let $n,s,t,d,k,\Lambda > 0$, $\beta,\eps,\delta \in (0,1]$, $\gamma \in (0,\frac1{16}]$, and let $\cP$ be a multiset of $n$ points in $B(\pt{0},\Lambda) \subseteq \bbR^d$. Let $\Ac$ be an algorithm that outputs $k$ centers in $\bbR^d$. Let $\Ac'$ be an algorithm that given a multiset $\cS$ of points over $B(\pt{0},\Lambda)$, w.p. $1-\frac{\beta}{2k}$ estimates $\Avg(\cS)$ up to an additive error $\zeta/\size{\cS}$ for $\zeta = \zeta(d, \Lambda, \frac{\beta}{2k})$.
	Let $\Bc$ be an $(t, \: \alpha=1,\text{ }r_{\min}=\gamma/n,\: \beta/2, \: \Delta=1/\gamma, \: \Lambda)$-averages-estimator $k$-tuple clustering algorithm (\cref{def:averages-estimator}). Finally, let $C = \set{\pc_1,\ldots,\pc_k} \in (\bbR^d)^k$ with $\min_{i \neq j} \norm{\pc_i - \pc_j} \geq 1/n$. Consider a random execution of $\AlgPrivatekMeans^{\Ac,\Ac',\Bc}(\cP,t,k,\eps,\delta,\beta,\gamma)$ conditioned that the event $E_{C}^{\gamma}$ occurs by $\AlgGenerateCenters$ in Step~\ref{step:gen} of the execution.
	Then with probability  $1-\beta$ (over the above conditional execution), the output  $\hat{C} = \set{\hpc_1,\ldots,\hpc_k}$ of $\AlgPrivatekMeans$ satisfies
	\begin{align*}
		COST_{\cP}(\set{\hpc_1,\ldots,\hpc_k}) \leq (1 + 64\gamma) \COST_{\cP}(C) +  \zeta k (\zeta + 2\Delta)
	\end{align*}
\end{theorem}


\begin{proof}
	Consider a random execution of $\AlgPrivatekMeans^{\Ac,\Ac',\Bc}(\cP,t,k,\eps,\delta,\beta,\gamma)$ conditioned on the event $E^{\gamma}_C$. For $j \in [t]$, let $\tilde{C}_j = \set{\tpc_1^j,\ldots,\tpc_k^j}$ be the value from Step~\ref{step:non-priv-centers} of the $j$'th iteration of $\AlgGenerateCenters$, where we denote by $\tpc_i^j$ the center that is close to $\pc_i$, i.e., $\norm{\tpc_i^j - \pc_i} \leq \gamma \cdot D_i$, where $D_i = \min_{j \neq i}\norm{\pc_i- \pc_j}$ (such center exists by event $E_C^{\gamma}$). In addition, for $i \in [k]$, let $\pa_i = \Avg(\set{\tpc_{i}^j}_{j=1}^t)$ and note that
	\begin{align}\label{eq:dist-pai-pci}
		\forall i \in [k]:\text{}\norm{\pa_i - \pc_i}
		&= \norm{\frac1{t} \sum_{j=1}^t \tpc_{i}^j - \pc_i}\\
		&\leq \frac1{t} \sum_{j=1}^t \norm{\tpc_{i}^j - \pc_i}\nonumber\\
		&\leq \gamma \cdot D_i.\nonumber
	\end{align}
	Now, let $\cT=\set{\tilde{C}_1 = \set{\tpc_i^1}_{i=1}^k, \ldots,\tilde{C}_t = \set{\tpc_i^t}_{i=1}^k}$ be the output of $\AlgGenerateCenters$ in Step~\ref{step:gen} of $\AlgPrivatekMeans$, and note that $\cT$ is partitioned by the set of $(\Delta = 1/\gamma)$-far balls $\set{B(\pc_i, r_i = \gamma D_i)}_{i=1}^k$ where $\Partition(\cT) =  \set{\set{\tpc_{1}^j}_{j=1}^t, \ldots, \set{\tpc_{k}^j}_{j=1}^t}$. Therefore, when executing algorithm $\Bc$ in Step~\ref{step:the-aver-estim}, we obtain by assumption a set of $k$ points $\set{\hpa_1,\ldots,\hpa_k}$ such that with probability $1 - \frac{\beta}{2}$ it holds that
	\begin{align}\label{eq:dist-pai-hpai}
		\forall i \in [k]:\text{   }
		\text{}\norm{\pa_i - \hpa_i} \leq \max\set{r_i,r_{\min}} \leq \gamma\cdot  D_i,
	\end{align}
	where in the second inequality we used the fact that $r_i \leq \gamma D_i$ and that $r_{\min} = \gamma/n \leq \gamma D_i$. Therefore, we deduce by \cref{eq:dist-pai-pci,eq:dist-pai-hpai} that with probability $1 - \frac{\beta}2$ it holds that
	\begin{align}\label{eq:hpa-pc-dist}
		\forall i \in [k]:\text{}\norm{\hpa_i - \pc_i} \leq 2\gamma \cdot D_i.
	\end{align}
	Let $\cP_1,\ldots,\cP_k$ be the clusters from Step~\ref{step:compute-clusters-kmeans} of the algorithm. If \cref{eq:hpa-pc-dist} occurs, then by \cref{prop:close-centers-have-similar-cost} we get that 
	\begin{align}\label{eq:avg-cost}
		\sum_{i=1}^k \sum_{\px \in \cP_i} \norm{\px - \Avg(\cP_i)}^2 \leq (1 + 64\gamma) \COST_{\cP}(C). 
	\end{align}
	Since the algorithm computes a noisy estimation $\hpc_i$ of each $\Avg(\cP_i)$ using the oracle $\Ac'$, we get that with probability $1-k\hat{\beta} = 1 - \frac{\beta}{2}$ it holds that
	\begin{align}\label{eq:dist-from-avg}
		\forall i \in [k]:\quad \norm{\hpc_i - \Avg(\cP_i)}  \leq \zeta/\size{\cP_i}
	\end{align}
	
	Finally, since \cref{eq:avg-cost} occurs with probability $1 - \frac{\beta}{2}$, and \cref{eq:dist-from-avg} occurs with probability $1 - \frac{\beta}{2}$, we conculde that with probability $1-\beta$ both of them occurs, which implies that
	\begin{align*}
		\lefteqn{\COST_{\cP}(\set{\hpc_1,\ldots,\hpc_k})}\\
		&\leq \sum_{i=1}^k \sum_{\px \in \cP_i} \norm{\px - \hpc_i}^2\\
		&= \sum_{i=1}^k \sum_{\px \in \cP_i} \paren{\norm{\px - \Avg(\cP_i)}^2 + \norm{\hpc_i - \Avg(\cP_i)}^2 + 2 \norm{\px - \Avg(\cP_i)}\cdot \norm{\hpc_i - \Avg(\cP_i)}}\\
		&\leq (1 + 64\gamma) \COST_{\cP}(C) + k \zeta^2 + 2 \Lambda k \cdot \zeta.
	\end{align*}
	where in the last term of the second inequality we used the fact that $\norm{\px - \Avg(\cP_i)} \leq \Lambda$ for all $i \in [k]$ and $\px \in \cP_i$.\Hnote{We just discard $\size{\cP_i}^2$ in the denominator in the second term of the second inequality ?}\Enote{Yes}\Hnote{can't we gain anything for this ? mention this ?}\Enote{I guess we can and we should!! In general, I believe we can also improve the theoretical bounds here if we do the analysis more carefully.}
\end{proof}

\remove{
\begin{remark}[Run time of $\AlgPrivatekMeans$]
	The algorithm performs $t$ oracle queries to the non-private algorithm $\cA$, each time over a collection of points of size $s = O(n/t)$. Then, the most expensive step is executing $\AlgPrivatekAverages$, which takes $\tilde{O}(d k^2 n)$ time.
\end{remark}
}

%
%
%
%
%

\remove{

	

}




\remove{
}

\remove{
\Enote{Delete the following event}

}

\subsection{Private $k$-Means under Separation Assumption}\label{sec:application}
In this section we show that our stability assumption holds with high probability when the multiset $\cP$ is separated according to \citet{OstrovskyRSS12}.
Formally, a multiset of points $\cP$ is \emph{$\phi$-separated} for $k$-means if $\OPT_k(\cP) \leq \phi^2 \OPT_{k-1}(\cP)$.
In \cref{def:sep-ost} we strength this definition of \cite{OstrovskyRSS12} to include also an additive separating term $\xi$.

\begin{definition}[$(\phi,\xi)$-separated]\label{def:sep-ost}
	A multiset $\cP \in (\bbR^d)^*$ is $(\phi,\xi)$-separated for $k$-means if
	
	\noindent $\OPT_k(\cP) + \xi \leq \phi^2\cdot \OPT_{k-1}(\cP)$. Note that $\cP$ is $\phi$-separated iff it is $(\phi,0)$-separated.
\end{definition}

We use the following theorem from \cite{OstrovskyRSS12} which states that when $\cP$ is $\phi$-separated for $k$-means for sufficiently small $\phi$, then any set of $k$ centers that well approximate the $k$ means cost, must have the property that each of its centers is relatively close to an optimal center.

\begin{theorem}[\cite{OstrovskyRSS12}]\footnote{The statement of this theorem was taken from \cite{ShechnerSS20}.}\label{thm:Ostrovsky}
	Let $\nu$ and $\phi$ be such that $\frac{\nu + \phi^2}{1-\phi^2} < \frac1{16}$. Suppose that $\cP \in (\bbR^d)^*$ is $\phi$-separated for $k$-means. Let $C^* = \set{\pc_1^*,\ldots,\pc_k^*}$ be a set of \emph{optimal} centers for $\cP$, and let $C= \set{\pc_1,\ldots,\pc_k}$ be centers such that $\COST_{\cP}(C) \leq \nu \cdot \OPT_{k-1}(\cP)$. Then for each $\pc_i$ there is a distinct optimal center, call it $\pc_i^*$, such that $\norm{\pc_i - \pc_i^*} \leq 2 \cdot \frac{\nu + \phi}{1 - \phi}\cdot D_i$, where $D_i = \min_{j \neq i}\norm{\pc_i^* - \pc_j^*}$.
\end{theorem}


The following lemma states that for suitable choices of $\phi$ and $\xi$, if $\cP$ is $(\phi,\xi)$-separated for $k$-means, then with high probability, the event $E_{C^*}^{\gamma}$ over a random execution of $\AlgGenerateCenters$ (\cref{def:event-ECgamma}) occurs, where $C^*$ is the optimal $k$-means for $\cP$.

\begin{lemma}[Bounding the stability probability]\label{lem:bounding-stability-under-sep}
	Let $n,d,k,s,t \in \bbN$, $\beta, \phi \in (0,1)$,  $\gamma \in (0,1/16]$, be values such that $\frac{(1 + \omega) \phi^2}{1-\phi^2} < \frac1{16}$ and $\gamma \geq 2\cdot \frac{\omega \phi^2 + \phi}{1 - \phi}$. Let $\Ac$ be a (non-private) $\omega$-approximation algorithm for $k$-means, let $\cP \in (B(\pt{0},\Lambda))^n$ and let $C^* = \set{\pc_1^*,\ldots,\pc_k^*} \in (\bbR^d)^k$ be the \emph{optimal} $k$-means for $\cP$. Assume that $\cP$ is $(\phi,\xi)$-separated for $k$-means, where $$\xi = \xi\paren{s, \beta/t} = \tilde{O}\paren{\Lambda^2 k d \log(n t/\beta) \cdot \frac{n}{s} + \Lambda \sqrt{k d \log(n t/\beta) \cdot \omega \OPT_{k}(\cP) \cdot \frac{n}{s}}}$$
	is the function from \cref{prop:cost-of-sample-is-good}.
	Then when executing $\AlgGenerateCenters^{\Ac}(\cP,k,s,t)$, the event $E_{C^*}^{\gamma}$ (\cref{def:event-ECgamma}) occurs with probability at least $1-\beta$.
\end{lemma}
\begin{proof}
	For $j \in [t]$, let $\cS_j$ and $\tilde{C}_j$ be the values in steps \ref{step:sample} and \ref{step:non-priv-centers}  of $\AlgGenerateCenters$ (respectively). Note that by 
	\cref{prop:cost-of-sample-is-good} and the union bound, with probability at least $1-\beta$ it holds that
	\begin{align}\label{eq:small-cost}
		\forall j \in [t]: \quad \COST_{\cP}(\tilde{C}_j) \leq \omega\cdot \OPT_k(\cP) + \xi \leq \omega \phi^2 \OPT_{k-1}(\cP),
	\end{align}
	where the last inequality holds by the assumption that $\cP$ is $(\phi,\xi)$-separated for $k$-means and that $\omega \geq 1$.
	In the following, assume that (\ref{eq:small-cost}) occurs. Since $\cP$ is (in particular) $\phi$-separated, and since the conditions of \cref{thm:Ostrovsky} hold with $\nu = \omega \phi^2$, we obtain from \cref{thm:Ostrovsky} that for  
	every $i \in [k]$ and $j \in [t]$, there exists $\tpc_i^j \in \tilde{C}_j$ such that $\norm{\pc_i^* - \tpc_i^j} \leq \gamma D_i$, meaning that event $E_{C^*}^{\gamma}$ occurs, as required.
\end{proof}

As a corollary of \cref{claim:kMeans-utility,lem:bounding-stability-under-sep}, we obtain our main application of algorithm $\AlgPrivatekMeans$.

\begin{theorem}\label{cor:util-result}
	Let $n,s,t,k,d \in \bbN$, $\eps,\delta,\beta \in (0,1]$, $\Lambda > 0$, let  $\phi, \cP, \Ac, \omega, \gamma$ as in \cref{lem:bounding-stability-under-sep}, and let $\zeta, \Ac',\Bc$ as in \cref{claim:kMeans-utility}.
	Then when executing $\AlgPrivatekMeans^{\Ac,\Ac',\Bc}(\cP,s,t,k,\eps,\delta,\beta,\gamma)$, with probability $1-2\beta$, the resulting centers $\hat{C} = \set{\hpc_1,\ldots,\hpc_k}$ satisfy
	\begin{align*}
		\COST_{\cP}(\hat{C}) \leq (1 + 64\gamma) \OPT_{k}(\cP) +  \zeta k (\zeta + 2\Lambda)
	\end{align*}
\end{theorem}
\begin{proof}
	The proof almost immediately holds by \cref{claim:kMeans-utility,lem:bounding-stability-under-sep} when applying them to the optimal $k$-means of $\cP$, which we denote by $C^* = \set{\pc_1^*,\ldots,\pc_k^*}$.  The only missing requirement is to show that $D^* \eqdef \min_{i \neq j} \norm{\pc_i^* - \pc_j^*} \geq 1/n$, as required by \cref{claim:kMeans-utility}. For proving this, note that on the one hand it holds that $\OPT_{k-1}(\cP) \leq D^* n + \OPT_{k}(\cP)$, and on the other hand, since we assume that $\cP$ is $(\phi,\xi)$-separated for $\phi \leq 1$ and $\xi \geq 1$ then it holds that $\OPT_k(\cP) + 1 \leq \OPT_{k-1}(\cP)$. From the two inequalities we conclude that $D^* \geq 1/n$ and the corollary follows.
\end{proof}

We note that by the utility guarantee (\cref{claim:utility-kAverg}) of our $k$-tuple clustering $\AlgPrivatekAverages$ (\cref{alg:FindAverages}), by choosing $\Bc = \AlgPrivatekAverages(\cdot, \eps/6, \delta/(4 e^{\eps}), \beta/2, r_{\min}=\gamma/n)$ we obtain  that $\Bc$ is an $\left(\frac{\eps}{6},\frac{\delta}{4 e^{\eps}}\right)$-differentially private $(t, \alpha=1, \: r_{\min}=\gamma/n, \: \beta/2, \: \Delta = 1/\gamma, \: \Lambda)$-averages-estimator for $k$-tuple clustering, where (ignoring $\polylog(n,d,k)$ factors)
\begin{align*}
	t =\tilde{\Omega}\paren{\frac{d k \cdot  \log^{2.5}(1/\delta)\paren{\sqrt{\log(1/\delta)} +  \log\paren{\Lambda/r_{\min}}}}{\eps^2}}.
\end{align*}
Hence, by taking $n = 2st$ we conclude that $\AlgPrivatekMeans^{\Ac, \Ac', \Bc}(\cdot, k, s, t, \eps,\delta,\beta,\gamma)$ is an $(\eps,\delta)$-differentially private algorithm (\cref{claim:kMeans-privacy}) with the utility guarantee stated in \cref{cor:util-result}. In particular, when taking $n = 2st$ we obtain that our theorem holds for an additive error $\xi$ in the separation (see \cref{lem:bounding-stability-under-sep} for the definition of $\xi$), where (ignoring logarithmic factors)
\begin{align*}
	\xi 
	= \tilde{\Omega}\paren{\Lambda^2 k d t + \Lambda \sqrt{k d t \omega \cdot \OPT_{k}(\cP)}}
\end{align*}


%
%
\section{Mixture of Gaussians}\label{sec:mixture-of-gaus}

In this section we present our second application of $k$-tuple clustering, which is an $(\eps,\delta)$-differentially private algorithm $\AlgPrivatekGaussians$ for learning a mixture of well separated and bounded $k$ Gaussians. We first start with relevant preliminaries for this section.

\subsection{Preliminaries}

The total variation distance between two distributions $P$ and $Q$ over a universe $\cU$ is defined by $\dTV(P,Q) = \sup_{\cS \subseteq \cU} \size{P(\cS) - Q(\cS)}$. Given a matrix $A = (a_{i,j})_{i,j \in [d]} \in \bbR^{d \times d}$, we let $\norm{A} = \sup_{\norm{\px} = 1} \norm{A \px}$ be its $\ell_2$ norm.



\subsubsection{Gaussians}
Let $\cN(0,1)$ be the standard Gaussian distribution over $\bbR$ with probability density function $p(z) = \frac1{\sqrt{2\pi}} e^{-\frac{z^2}2}$.
In $\bbR^d$, let $\cN(\pt{0},\bbI_{d \times d})$ be the standard multivariate Gaussian distribution. That is, if $\pZ \sim \cN(\pt{0},\bbI_{d \times d})$ then $\pZ = (Z_1,\ldots,Z_d)$ where $Z_1,\ldots,Z_d$ are i.i.d. according to $N(0,1)$. Other Gaussian distributions over $\bbR^d$ arise by applying (invertible) linear maps on $\cN(\pt{0},\bbI_{d \times d})$. That is, the distribution $\pX \sim \cN(\pmu, \Sigma = AA^T)$ for $\mu \in \bbR^d$ and (invertible) $A \in \bbR^{d \times d}$ is defined by $\pX = A \pZ + \pmu$, where $\pZ \sim N(\pt{0},\bbI_{d \times d})$, and it holds that $\ex{\pX} = \mu$ and $\Cov(\pX) = \paren{\Cov(X_i,X_j)}_{i,j}$ (the covariance matrix) equals to $\Sigma$. The contours of equal density are ellipsoids around $\mu$: $\set{\px \in \bbR^d \colon (\px - \mu)^T \Sigma^{-1}(\px - \mu) = r^2}$. We let $\cG(d)$ be the family of all $d$-dimensional Gaussian --- that is, the set of all distribution $\cN(\mu,\Sigma)$ where $\mu \in \bbR^d$ and $\Sigma$ is a $d \times d$ positive semidefinite (PSD) matrix.

\begin{definition}[Bounded Gaussian]\label{def:bounded-gaus}
	For $R,\sigma_{\max},\sigma_{\min} > 0$, a Gaussian $\pG = \cN(\mu,\Sigma) \in \cG(d)$ is $(R,\sigma_{\max},\sigma_{\min})$-bounded if $\norm{\mu} \leq R$ and  $\sigma_{\min}^2 \leq \norm{\Sigma} \leq \sigma_{\max}^2$.
\end{definition}

We next define the properties of a general algorithm that learns the parameters of a (single) bounded Gaussian.

\begin{definition}[Learner for Bounded Gaussians]\label{def:LearnAlgSingleGaus}
	Let $\Ac$ be an algorithm that gets as input a database $\cP \in (\bbR^d)^*$ and outputs $(\hat{\mu},\hat{\Sigma})$.
	We say that $\Ac$ is an $(\upsilon,\eta,\beta)$-learner for $(R,\sigma_{\max},\sigma_{\min})$-bounded Gaussians, if for any 
	such bounded Gaussian $\cN(\mu,\Sigma)$, algorithm $\Ac$ given $\upsilon$ i.i.d.\ samples from it as input, outputs w.p. $1-\beta$ a pair $(\hat{\mu},\hat{\Sigma})$ with $\dTV(\cN(\mu,\Sigma), \cN(\hat{\mu},\hat{\Sigma})) \leq \eta$.
\end{definition}

\remove{
\begin{definition}[Private Algorithm for Learning a Bounded Gaussian]\label{def:privLearnAlgSingleGaus}
	Let $\Ac$ be an algorithm that gets as input a database $\cP \in (\bbR^d)^*$ and parameters $d, \eps,\delta,\eta,\beta,R,\sigma_{\max},\sigma_{\min}$, and outputs $(\hat{\mu},\hat{\Sigma})$. Let $s = s(d, \eps,\delta,\eta,\beta,R,\sigma_{\max},\sigma_{\min})$ be a function.
	We say that $\Ac$ is a \textbf{private algorithm for learning a bounded Gaussian with sample complexity $\upsilon = \upsilon(d, \eps,\delta,\eta,\beta,R,\sigma_{\max},\sigma_{\min})$} if given the above parameters, $\Ac$ is an $(\eps,\delta)$-differentially private algorithm that satisfy the following utility guarantee: If $\cN(\mu,\Sigma)$ is a $(R,\sigma_{\max},\sigma_{\min})$-bounded Gaussian, and $\cP$ consists of at least $\upsilon$ i.i.d.\ samples from $\cN(\mu,\Sigma)$, then with probability at least $1-\beta$ it holds that $\dTV(\cN(\mu,\Sigma), \cN(\hat{\mu},\hat{\Sigma})) \leq \eta$.
\end{definition}
}

In our construction, we would like to use a \emph{differentially private} algorithm that learns the parameters of single (bounded) Gaussians.
The best known examples for such algorithms  are the constructions of \cite{KLSU19} and \cite{BDKU20} in the \emph{zero Concentrated DP} model (zCDP \cite{BS16}). For instance, the algorithm of \cite{KLSU19} is $\frac{\eps^2}2$-zCDP $(\upsilon,\eta,\beta)$-learner for $(R,\sigma_{\max},\sigma_{\min})$-bounded Gaussians, for 
$\upsilon = \tilde{O}\paren{\paren{\frac{d^2}{\eta^2} + \frac{d^2}{\eps \eta} + \frac{d^{3/2} \sqrt{\log \paren{\frac{\sigma_{\max}}{\sigma_{\min}}}} + \sqrt{d \log R}}{\eps}}\cdot \log(1/\beta)}$.
We first remark that $\eps$-DP implies $\frac{\eps^2}{2}$-zCDP, and the latter implies $(\eps\sqrt{\log(1/\delta)},\delta)$ for every $\delta > 0$. We also remark that
without privacy, the required sample complexity is $\Theta\paren{\frac{d^2 \log(1/\beta)}{\eta^2}}$, which means that privacy comes almost for free unless $\frac1{\eps}, \frac{\sigma_{\max}}{\sigma_{\min}}$ or $R$ are quite large.

\subsubsection{Gaussian Mixtures}
The class of Gaussian $k$-mixtures in $\bbR^d$ is 
\begin{align*}
\cG(d,k) \eqdef \set{\sum_{i=1}^k w_i \pG_i \colon \pG_1,\ldots,\pG_k \in \cG(d), w_1,\ldots,w_k > 0, \sum_{i=1}^k w_i = 1}
\end{align*}
A Gaussian mixture can be specified by a set of $k$ triplets: $\set{(\mu_1,\Sigma_1,w_1), \ldots, (\mu_k,\Sigma_k,w_k)}$, where each triplet represents the mean, covariance matrix, and mixing weight of one of its components.

\begin{definition}[Bounded Mixture of Gaussians]
	For $R,\sigma_{\max},\sigma_{\min},w_{\min} > 0$, a Gaussian mixture $\cD = \set{(\mu_1,\Sigma_1,w_1), \ldots, (\mu_k,\Sigma_k,w_k)} \in \cG(d,k)$ is $(R,\sigma_{\max},\sigma_{\min},w_{\min})$-bounded if for all $i \in [k]$, the Gaussian $\cN(\mu_i,\Sigma_i)$ is $(R,\sigma_{\max},\sigma_{\min})$-bounded and $w_i \geq w_{\min}$.
\end{definition}

\begin{definition}[Separated Mixture of Gaussians]
	Let $\cD = \set{(\mu_1,\Sigma_1,w_1), \ldots, (\mu_k,\Sigma_k,w_k)}$ be a mixture of $k$ Gaussians over $\bbR^d$, for $i \in [k]$ let $\sigma_i^2 = \norm{\Sigma_i}$, and let $h > 0$.
	We say that $\cD$ is $h$-separated if
	\begin{align*}
	\forall 1 \leq i < j \leq k: \text{ }\norm{\mu_i-\mu_j} \geq h \cdot \max\set{\sigma_i,\sigma_j}.
	\end{align*}
\end{definition}

We next define a labeling algorithm for a mixture $\cD$.

\begin{definition}[Labeling Algorithm for a Mixture of Gaussians]\label{def:LabelingAlg}
	Let $s,k \in \bbN$, $\beta  \in (0,1)$ and let
	$\cD = \set{(\mu_1,\Sigma_1,w_1), \ldots, (\mu_k,\Sigma_k,w_k)}$  be a mixture of $k$ Gaussians. We say that an Algorithm $\Ac$ is an \textbf{$(s,\beta)$-labeling algorithm for the mixture $\cD$} if with probability $1-\beta$, when sampling a database $\cP$ of $s$ i.i.d.\ samples from $\cD$,  algorithm $\Ac$ on inputs $\cP,k$, outputs a labeling function $L \colon \cP \rightarrow [k]$ such that for all $\px, \px' \in \cP$: $\quad L(\px) = L(\px')$ $\iff$ $\px$ and $\px'$ were drawn from the same Gaussian.
\end{definition}

There are various examples of non-private algorithms that learns the parameters of mixtures of Gaussian under different separations assumptions, and most of them can be easily converted into a labeling algorithm.
For instance, \cite{DS00,AK01} showed how to learn mixtures with separation that is only proportional to $d^{1/4}$. Moreover, there is a wide line of works that show how to handle mixtures with separation that is independent of $d$: Separation that is proportional to $\sqrt{k}$ \cite{AM05}, $k^{1/4}$ \cite{VW04}, $k^{\eps}$ \cite{HSJ18,KPSJS18,diakonikolas18}, or even $\sqrt{\log k}$ \cite{RV17}. In \cref{sec:GausAlg} we show that our algorithm can transform each such non-private algorithm into a private one, as long as we are given $n$ points from a mixture that is at least $\tilde{\Omega}(\log n)-$separated.

\subsubsection{Concentration Bounds}

\begin{fact}[One-dimensional Gaussian]\label{fact:one-Gaus-concet}
	Let $\pX \sim \cN(0,\sigma^2)$. Then for any $\beta > 0$ it holds that
	\begin{align*}
	\pr{\pX \geq \sigma \sqrt{2 \log(1/\beta)}} \leq \beta.
	\end{align*}
\end{fact}

\begin{fact}[follows by the Hanson-Wright inequality \cite{HV71}]\label{fact:gaus-concent-new}
	If $\pX \sim \cN(\mu,\Sigma)$ then with probability at least $1-\beta$ it holds that
	\begin{align*}
	\norm{\pX - \mu} \leq \paren{\sqrt{d} + \sqrt{2 \log(1/\beta)}} \cdot \sqrt{\norm{\Sigma}}.
	\end{align*}
\end{fact}

The following fact is an immediate corollary of \cref{fact:gaus-concent-new}.

\begin{fact}\label{fact:gaus-avg}
	Let $X_1,\ldots,X_m$ be i.i.d. random variables distributed according to a $d$-dimensional Gaussian $\cN(\mu,\Sigma)$, and let $\sigma^2 = \norm{\Sigma}$.  Then with $1-\beta$ it holds that
	\begin{align*}
	\norm{\Avg(X_1,\ldots,X_m) - \mu} \leq \frac{\sqrt{d} + \sqrt{2 \log (1/\beta)}}{\sqrt{m}} \cdot \sigma,
	\end{align*}
\end{fact}
\begin{proof}
	Follows by \cref{fact:gaus-concent-new} since  $\Avg(X_1,\ldots,X_m)$ is distributed according to $\cN(\mu,\frac1{m}\cdot \Sigma)$.
\end{proof}


%

%

\subsection{Algorithm $\AlgPrivatekGaussians$}\label{sec:GausAlg}
In this section we describe our algorithm $\AlgPrivatekGaussians$ (\cref{alg:kGaus}) that privately learns a mixture of separated and bounded $k$ Gaussians $\cD = \set{(\mu_1,\Sigma_1,w_1), \ldots, (\mu_k,\Sigma_k,w_k)}$.

\begin{algorithm}[$\AlgCollectEmpiricalMeans$]\label{alg:genBalancedSamples}
	\item Input: A database  $\cP' = \set{\px_1,\ldots,\px_{n}}$ and parameters $k, s,t \in \bbN$, where $n \geq st$.
	
	\item Oracle: a (non-private) labeling algorithm $\Ac$ for a mixture of Gaussians.
	
	\item Operation:~
	\begin{enumerate}
		
		\item For each $j \in [t]$:
		
		\begin{enumerate}
			\item Let $\cS_j = \set{\px_{(j-1) s + 1},\ldots,\px_{j s}}$.\label{step:sample-gaus}
			
			\item Execute $\Ac$ on inputs $\tilde{\cP} = \cS_j, \tilde{k} =k$, and let $L_j \colon \cS_j \rightarrow [k]$ be the resulting labeling function.\label{step:labeling}
			
			\item For each $i \in [k]:$
			\begin{itemize}
				\item Compute $\bar{\mu}_{j,i} = \Avg\paren{\set{\px \in \cS_j \colon L_j(\px) = i}}$.\label{step:compute-emp-mean}
			\end{itemize}
			
			\item Set $M_j = \set{\bar{\mu}_{j,1}, \ldots,\bar{\mu}_{j,k}} \in (\bbR^d)^k$.\label{step:balanced-samples}
		\end{enumerate}
		
		\item Output $\cT = \set{M_1, \ldots, M_t} \in ((\bbR^d)^k)^t$.\label{step:new-multisets-gaus}
	\end{enumerate}
\end{algorithm}

%
%
%
%
%
%
%
%

\begin{algorithm}[$\AlgPrivatekGaussians$]\label{alg:kGaus}
	
	\item Input: A database  $\cP = \set{\px_1,\ldots,\px_{2n}} \in \paren{\bbR^d}^{2n}$, parameters $k,s,t \in \bbN$ s.t. $n \geq s t$,, and privacy parameter $\eps,\delta > 0$.
	
	\item Oracles: A (non-private) labeling algorithm $\Ac$, an $(\eps/4,\delta/2)$-DP learner $\Ac'$ for (single) Gaussians, and an $(\eps,\delta)$-DP $k$-tuple clustering algorithm $\Bc$.
	
	\item Operation:~
	\begin{enumerate}
		
		\item Let $\cP' = \set{\px_1,\ldots,\px_{n}}$ and $\cP'' =  \set{\px_{n+1},\ldots,\px_{2n}}$.\label{step:split-database}
		
		
		
		
		\item Compute $\cT =  \AlgCollectEmpiricalMeans^{\Ac}(\cP',k,s,t)$.\label{step:cM}
		
		
		
		\item Compute $\set{\hat{\pa}_1,\ldots,\hat{\pa}_k} = \Bc(\cT)$.\label{step:kAveragesOnM}
		
		
		
		\item For each $i \in [k]:$\label{step:Gaussian:forloop}
		\begin{enumerate}
			\item Let $\cP_i''$ be the points in $\cP''$ that $\hpa_i$ is the closest point to them among $\set{\hat{\pa}_1,\ldots,\hat{\pa}_k}$. \label{step:compute-clusters-kmeans-gaus}
			
			\item Compute $(\hat{\mu}_i,\hat{\Sigma}_i) = \Ac'(\cP_i'')$.\label{step:priv-single-Gauss-est}
			
			
			\item Let $\hat{n}_i \la \size{\cP_i''} + \Lap(4/\eps)$.\label{step:Lap}
			
		\end{enumerate}
		
		\item For each $i \in [k]: \quad $ Set $\hat{w}_i = \frac{\hat{n}_i}{\sum_j \hat{n}_j}$.\label{step:wi}
		
		\item Output $\hat{\cD} = \set{(\hat{\mu}_1,\hat{\Sigma}_1,\hat{w}_1),\ldots,(\hat{\mu}_k,\hat{\Sigma}_k,\hat{w}_k)}$.
	\end{enumerate}
\end{algorithm}

\subsubsection{Properties of $\AlgPrivatekGaussians$}
The following theorem summarizes the privacy guarantee of $\AlgPrivatekGaussians$. 

As a running example, fix the following target parameters: An accuracy parameter $\eta > 0$, a confidence parameter $\beta > 0$, privacy parameters $\eps,\delta \in (0,1)$ and bounding parameters $R,\sigma_{\max},\sigma_{\min} > 0$.
Also, think of the (non-private) labeling algorithm $\Ac$ as the one of \cite{AM05} that needs $s = \tilde{O}\paren{\frac{dk}{w_{\min}}}$ samples, think of the private  (single) Gaussian learner $\Ac'$ as the $(\eps/2,\delta/4)$-DP variant of the algorithm of \cite{KLSU19} that needs $\upsilon = \tilde{O}\paren{\frac{d^2}{\eta^2} + \paren{\frac{d^2}{\eps \eta} + \frac{d^{3/2} \sqrt{\log \paren{\frac{\sigma_{\max}}{\sigma_{\min}}}} + \sqrt{d \log R}}{\eps}} \cdot \sqrt{\log(1/\delta)}}$ samples,
and think of algorithm $\Bc$ as our averages-estimator $\AlgPrivatekAverages$ (\cref{alg:FindAverages}) that needs $t =\tilde{O}\paren{\frac{d k \cdot  \log^{2.5}(1/\delta)\paren{\sqrt{\log(1/\delta)} +  \log\paren{\Lambda/r_{\min}}}}{\eps^2}}$ tuples. The values of $r_{\min}$ and $\Lambda$ that we should use are determined by our utility guarantee (\cref{thm:kGauss-utility}). This means that for the oracle $\Bc$ we actually should wrap $\AlgPrivatekAverages$ such that if there is a tuple that contains a point which is not in $B(\pt{0},\Lambda)$, then $\Bc$ replaces it with some arbitrary fixed tuple (e.g., the all-zero tuple $\pt{0}^k$).


We first state the privacy guarantee of $\AlgPrivatekGaussians$.

\begin{theorem}[Privacy of $\AlgPrivatekGaussians$]\label{thm:kGauss-privacy}
	Let $\Ac$ be an (arbitrary, non-private) labeling algorithm,
	let $\Ac'$ be an $(\eps/4,\delta/2)$-DP algorithm, and let $\Bc$ be an $(\eps,\delta)$-differentially private algorithm for databases $\cT \in ((\bbR^d)^k)^t$ (i.e., of size $t$).
	Then for every $d,k,R,\sigma_{\max},\sigma_{\min},w_{\min} > 0$ and  $\eta,\beta,\eps,\delta, \gamma \in (0,1)$, algorithm $\AlgPrivatekGaussians^{\Ac,\Ac',\Bc}(\cdot, k, s,t,\eta,\beta,\eps, \delta,R, \sigma_{\max}, \sigma_{\min})$ is $(\eps,\delta)$-differentially private for databases $\cP \in \paren{\bbR^d}^*$.
\end{theorem}

\begin{proof}
	Assume for simplicity (and without loss of generality) that the input algorithm $\Ac$ is deterministic, let $\cP,\tilde{\cP} \in \paren{\bbR^d}^{2n}$ be two neigboring databases, and consider two executions $\AlgPrivatekGaussians(\cP)$ and $\AlgPrivatekGaussians(\tilde{\cP})$ (both with the same input parameters and oracles). Let $\cP'$, $\cP''$, $\cT$ be the multisets from the execution $\AlgPrivatekGaussians(\cP)$, and  let $\tilde{\cP}'$, $\tilde{\cP}''$, $\tilde{\cT}$ be the corresponding multisets in the execution $\AlgPrivatekGaussians(\tilde{\cP})$. 
	
	
	If $\cP'\neq \tilde{\cP}'$ (i.e., neighboring), then assume w.l.o.g. that the two executions share the same randomness in Steps \ref{step:cM} (i.e., in the execution of $\AlgCollectEmpiricalMeans$), but use independent randomness in the execution of $\Bc$ in Step \ref{step:kAveragesOnM} and in the next steps of the algorithm. Therefore, $\cT$ and $\cT'$ differ by at most one $k$-tuple.
	Hence, by the privacy guarantee of $\Bc$ along with group privacy (\cref{fact:group-priv}) we obtain that the resulting outcome $\set{\hpa_1,\ldots,\hpa_k}$ in Step~\ref{step:kAveragesOnM} of both executions is $(\eps,\delta)$-indistinguishable. Since $\cP' \neq \tilde{\cP}'$ implies that $\cP'' = \tilde{\cP}''$, we conclude by post-processing (\cref{fact:post-processing}) that the final outcome $\hat{\cD}$ is also $(\eps,\delta)$-indistinguishable. 
	
	In the rest of the analysis we focus on the case that $\cP' = \tilde{\cP}'$ and $\cP'' \neq \tilde{\cP}''$ (i.e., neighboring). In this case, 
	we assume that both executions share the same randomness up to (and including) Step~\ref{step:kAveragesOnM}, and use independent randomness from \ref{step:Gaussian:forloop} till the end. 
	Therefore, the value of $\set{\hpa_1,\ldots,\hpa_k}$ in Step~\ref{step:kAveragesOnM} is identical in both executions. Let $\cP''_1,\ldots,\cP''_k$ be the multisets from Step~\ref{step:compute-clusters-kmeans-gaus} in the execution $\AlgPrivatekGaussians(\cP)$, and let $\tcP''_1,\ldots,\tcP''_k$ be these multisets in the execution $\AlgPrivatekGaussians(\tcP)$. 
	Since $\cP''$ and $\tcP''$ are neighboring, there exists at most two values $i,j \in [k]$ such that $\cP''_i \neq \tcP_i''$ and $\cP''_j \neq \tcP_j''$, and in both cases the multisets are neighboring (in the other indices the multisets are equal). By the properties of the private algorithm $\Ac'$ and basic composition (\cref{thm:composition1}), the vector $((\hat{\mu}_1,\hat{\Sigma}_1),\ldots,(\hat{\mu}_k,\hat{\Sigma}_k))$ computed in Step~\ref{step:priv-single-Gauss-est} of both executions is $(2\cdot \frac{\eps}{4},2 \cdot \frac{\delta}{2})$-indistinguishable. Moreover, by the properties of the Laplace Mechanism along with basic composition, the vector $(\hat{n}_1,\ldots,\hat{n}_k)$ is $(2\cdot \frac{\eps}{4}, 0)$-indistinguishable. By applying again basic composition we deduce that both vectors together are $(\eps,\delta)$-indistinguishable, and therefore we conclude by post-processing (\cref{fact:post-processing}) that the resulting $\hat{\cD}$ in both execution is $(\eps,\delta)$-indistinguishable.
\end{proof}

The following theorem summarizes the utility guarantee of $\AlgPrivatekGaussians$. 

\def\thmKGaussUtility{
	Let $n,d,k,s,t,\upsilon \in \bbN$, $R,\sigma_{\max},\sigma_{\min},w_{\min}, \gamma > 0$, $\eta,\beta,\eps,\delta \in (0,1)$, let $\Delta = 8 + 12/\gamma$, and let $\cD =  \set{(\mu_1,\Sigma_1,w_1), \ldots, (\mu_k,\Sigma_k,w_k)}$ be an $(R,\sigma_{\max},\sigma_{\min},w_{\min})$-bounded $(1+\gamma)h$-separated
	mixture of $k$ Gaussians in $\bbR^d$, for $h \geq 2\sqrt{2\log\paren{8n/\beta}}$.
	In addition, let $\Ac$ be a (non-private) $\paren{s,\frac{\beta}{8t}}$-labeling algorithm for $\cD$ (\cref{def:LabelingAlg}), let $\Ac'$ be an $(\upsilon,\frac{\eta}2,\frac{\beta}{16k})$-learner for $(R,\sigma_{\max},\sigma_{\min})$-bounded Gaussians (\cref{def:LearnAlgSingleGaus}), and let $\Bc$ be an $(t,\text{ }\alpha=1,\text{ }r_{\min}=\frac{(1+\gamma) h}{\Delta}\cdot \sigma_{\min},\text{ } \beta/8,\text{ }\Delta,\text{ }\Lambda=R + \frac{(1+\gamma) h}{\Delta}\cdot \sigma_{\max})$-averages-estimator for $k$-tuple clustering (\cref{def:averages-estimator}).
	Assume that
	\begin{align*}
			s \geq \frac{4}{w_{\min}}\cdot \max\biggl\{\log\paren{8 k t/\beta}, \quad \frac{\Delta^2\paren{d + 2 \log\paren{16 k t/\beta}}}{(1+\gamma^2) h^2} \biggr\},
	\end{align*}
	and that
	\begin{align*}
	n \geq \max \biggl\{s\cdot t, \quad \frac{2 \upsilon + \log(16k/\beta)}{w_{\min}}, \quad \frac{4k^2}{\eps \eta} \cdot \log\paren{8k/\beta} \biggr\}
	\end{align*}
	Then with probability $1-\beta$, when sampling a database $\cP$ of $2n$ i.i.d.\ samples from $\cD$, the execution $\AlgPrivatekGaussians^{\Ac,\Ac',\Bc}(\cP,k,s,t,\eps)$ outputs $\hat{\cD}$ such that $\dTV(\cD,\hat{\cD}) \leq \eta$.
}

\begin{theorem}[Utility of $\AlgPrivatekGaussians$]\label{thm:kGauss-utility}
	\thmKGaussUtility
\end{theorem}

The proof of the theorem appears at \cref{missing-proof:thm:kGauss-utility}. Very roughly, 
the first term in the bound on $n$ is because $\AlgCollectEmpiricalMeans$ splits the $n$ samples into $t$ pieces, each contains $s$ samples.
The second and third terms in the bound on $n$ are the number of samples that are needed for guaranteeing that with probability $1-\frac{\beta}4$, for each $i\in [k]$, the resulting $(\hat{\mu}_i,\hat{\Sigma}_i)$ in Step~\ref{step:priv-single-Gauss-est} satisfy $\dTV(\cN(\hat{\mu}_i,\hat{\Sigma}_i), \cN(\mu_i,\Sigma_i)) \leq \frac{\eta}2$ and the resulting $\hat{w}_i$ in Step~\ref{step:wi} satisfy $\size{\hat{w}_i - w_i} \leq \frac{\eta}{k}$, which yields that $\dTV(\hat{\cD},\cD) \leq \eta$ (see \cref{fact:dTV-of-mixtures}). We remark that regardless of the non-private algorithm $\Ac$ that we are using and its assumption on $\cD$, we only require that $\cD$ is more than $2\sqrt{2\log\paren{\frac{8n}{\beta}}}$-separated, which follows by the projection argument in \cref{prop:separation}.

\remove{
\begin{remark}
	\Enote{Add running time analysis}
\end{remark}
}

\subsection{Remarks}\label{sec:Gaussians:remarks}
It is tempting to think that our approach, which relies on the algorithm $\Bc = \AlgPrivatekAverages$ for aggregating the non-private findings by a reduction to $k$-tuple clustering, requires that the distance between the means should be proportional to $\sqrt{d}$, because this is the distance of the samples from their means. However, recall that $\AlgPrivatekGaussians$ do not set the $k$-tuple to be some arbitrary $k$ samples from different Gaussians. Rather, it sets it to the \emph{averages} of the samples in each set (See Step~\ref{step:compute-emp-mean} in \cref{alg:genBalancedSamples}), which decreases the distance from the actual means. In particular, when there are $O(d)$ samples in each such set, the dependency in $d$ is eliminated and the reduction to the $k$-tuple clustering follows (even when the distance between the means is much smaller than $\sqrt{d}$, as we consider).

Furthermore, note that our algorithm  $\AlgPrivatekGaussians$ in Step~\ref{step:compute-clusters-kmeans-gaus} relies on the fact that the output $\set{\hat{\pa}_1,\ldots,\hat{\pa}_k}$ of $\AlgPrivatekAverages$ separates correctly fresh samples from the mixture. 
This might seem strange since even if $\set{\hat{\pa}_1,\ldots,\hat{\pa}_k}$ is very close to the actual means $\set{\mu_1,\ldots,\mu_k}$, the distance of each sample from its mean is proportional to $\sqrt{d}$, while the assumed separation between the means is independent of $d$. This yields that when $d$ is large, then the samples are much far from their means compared to the distance between the means. Namely, 
if $\px$ is sampled from the $i$'th Gaussian and $\norm{\mu_i-\mu_j}$ is independent of $d$ (for large $d$), then $\norm{\px - \mu_i} \gg \norm{\mu_i - \mu_j}$. Yet, in our analysis we use a projection argument (see \cref{prop:separation}) which yields that w.h.p. it holds that $\norm{\px - \mu_i} < \norm{\px- \mu_j}$, even though $\norm{\px - \mu_i} \gg \norm{\mu_i - \mu_j}$.

\subsection{Comparison to the Main Algorithm of \cite{KSSU19}}\label{sec:Gauss:comparison}

The main private algorithm of \cite{KSSU19} mimics the approach of the (non-private) algorithm of \cite{AM05}, which is to use PCA to project the data into a low-dimensional space, and then clustering the data points in that low-dimensional space.
This projection enable both algorithms to learn mixtures that have the following separation
\begin{align}\label{eq:AM05-sep}
	\forall i,j\colon \quad \norm{\mu_i-\mu_j} \geq C \paren{\sqrt{k \log(nk/\beta)} + \frac1{\sqrt{w_i}} + \frac1{\sqrt{w_j}}} \cdot \max\set{\sigma_i,\sigma_j},
\end{align}
for some constant $C > 0$ (albeit that the constant of \cite{KSSU19} is much larger, say $C=100$ instead of $C=4$ as in \cite{AM05}).
But while \cite{AM05} use a simple Kruskal-based clustering method, 
\cite{KSSU19} developed alternative (and much more complicated) clustering methods that are more amenable to privacy. Finally, after the clustering phase, \cite{KSSU19} use a variant of the private algorithm of \cite{KLSU19} to learn the parameters of each Gaussian. Overall, the algorithm of \cite{KSSU19} learns an $(R,\sigma_{\max},\sigma_{\min},w_{\min})$-bounded mixture of Gaussian that is separated as in \cref{eq:AM05-sep}, with sample complexity 
\begin{align*}
	n \geq \paren{\frac{d^2}{\eta^2 w_{\min}} + \frac{d^2}{\eps \eta w_{\min}} + \frac{\poly(k) d^{3/2}}{w_{\min} \eps}} \cdot \polylog\paren{\frac{d k R \sigma_{\max}}{\eta \beta \eps \delta \sigma_{\min}}}
\end{align*}

In the following, we compare between \cite{KSSU19}'s algorithm and ours (Algorithm $\AlgPrivatekGaussians$) in two different aspects: separation assumption and sample complexity.

\subsubsection{Separation Assumption}
The utility guarantee of $\AlgPrivatekGaussians$ (\cref{thm:kGauss-utility}) only requires a separation of slightly more than $h =2\sqrt{2 \log(8n/\beta)}$. Therefore, our algorithm can transform any non-private algorithm (in a modular way) that learns mixtures with separation $X$ into a private algorithm that learns with separation $\max\set{X, h}$. In particular, we can use \cite{AM05} as our non-private labeling algorithm $\Ac$ to learn mixtures with separation as in \cref{eq:AM05-sep} (with the small constant $C=4$), and we can also use any other non-private algorithm (like \cite{VW04,HSJ18,KPSJS18,diakonikolas18,RV17}) and inherent their separation assumption.
In contrast, the approach of the main algorithm of \cite{KSSU19} may only be extended to methods that use statistical properties of the data (like PCA), and not to other algorithmic machineries such as the sum-of-squares that are used for reducing the separation assumption.

\subsubsection{Sample Complexity}

The main algorithm of \cite{KSSU19} learns an $(R,\sigma_{\max},\sigma_{\min},w_{\min})$-bounded mixture of Gaussians that is separated as in \cref{eq:AM05-sep}, with sample complexity (roughly) $\tilde{O}\paren{\frac{\upsilon}{w_{\min}} + \frac{k^9 d^{3/2}}{w_{\min} \eps}}$ (ignoring logarithmic factors), where $\upsilon = \upsilon(d, \eps,\delta,\eta,\beta,R,\sigma_{\max},\sigma_{\min}) = \tilde{O}\paren{\frac{d^2}{\eta^2} + \frac{d^2}{\eps \eta}}$ is the sample complexity of  \cite{KLSU19} for learning the parameters of a single Gaussian (ignoring logarithmic factors in $R,\sigma_{\max}/\sigma_{\min},1/\delta,1/\beta$).

By \cref{thm:kGauss-utility}, 
the sample complexity of our algorithm is $\tilde{O}\paren{s \cdot t + \frac{t \cdot d}{w_{\min}} +  \frac{\upsilon}{w_{\min}} +  \frac{4k^2}{\eps \eta}}$ (ignoring logarithmic factors), where $s$ is the sample complexity needed by the non-private algorithm $\Ac$ for labeling correctly the samples with confidence $\leq \frac{\beta}{8 t}$ (e.g., if we use the algorithm of \cite{AM05}, then $s = \tilde{O}\paren{\frac{dk}{w_{\min}}}$, and for simplifying the comparison, we assume that this is indeed the algorithm that we use). Since $t = \tilde{O}\paren{\frac{d k}{\eps^2}}$, we obtain a sample complexity of (roughly) $\tilde{O}\paren{\frac{k^2 d^2}{\eps^2 w_{\min}}  +  \frac{\upsilon}{w_{\min}} +  \frac{4k^2}{\eps \eta}}$, which might me larger than the one of \cite{KSSU19} if $d$ or $1/\eps$ are very large (compared to $k$). Yet, we can easily improve the dependency in both $d$ and $\eps$.

Using sub-sampling, we can execute Step~\ref{step:cM} of $\AlgPrivatekGaussians$ on an $\eps n$-size random subset of $\cP'$ (for the small desired $\eps$), but now we only need a constant $\eps$ for these steps. This immediately reduces the $1/\eps^2$ in our sample complexity into $1/\eps$.

In addition, as mentioned in \cref{sec:reducing-by-JL}, using the average algorithm of \cite{NSV16}  in $\AlgPrivatekAverages$ (instead of the average algorithm from \cref{prop:approx-aver-Rd}), we can reduce a factor of $\sqrt{d}$.

For summary, using sub-sampling and the algorithm of \cite{NSV16}, we obtain an improved sample complexity of $\tilde{O}\paren{\frac{k^2 d^{3/2}}{\eps w_{\min}}  +  \frac{\upsilon}{w_{\min}} +  \frac{4k^2}{\eps \eta}}$, which strictly improves the sample complexity of  \cite{KSSU19}.

\remove{
\subsubsection{Running-Time and Practicality}

\Enote{TBD. add running time analysis to all our algorithms, and check whether \cite{KSSU19} provide concrete running time analysis. Also, mention that our algorithm uses in a black-box way the the labeling algorithm and private learning algorithm of a single Gaussian. And also, that our algorithm can be easily parallelized}

}
\remove{
\Enote{Consider mentioning also that our algorithm is much more simpler, probabily more efficient and practical, and also can be easily parallelized.}

\Enote{
	Consider adding the following discussion to the intro:
	
	Can we use JL transform for learning a mixture of Gaussian? Suppose we have two spherical Gaussians $\cG_1 = \cN(\mu_1,\sigma^2 \bbI_{d \times d})$ and $\cG_2 =\cN(\mu_2,\sigma^2 \bbI_{d \times d})$ that we would like to separate, where $\norm{\mu_1 - \mu_2}$ is large enough (but independent of $d$). In the JL transform, we choose $k = \log n$ vectors  $\py_1,\ldots,\py_k \sim \cN(0,\bbI_{d \times d})$ and we transform each $\px \in \bbR^d$ into $(\ip{\px,\py_1},\ldots,\ip{\px,\py_k})$. Note that each $\ip{\px,\py_i}$ is distributed according to the $1$-dimensional Gaussian $\cN(0,d(\sigma^2+1))$, which yields that if $\pX \sim \cN(\mu,\sigma^2 \bbI_{d \times d})$ then $f(\pX) \sim \cN(\mu,\sigma^2 (d+1) \bbI_{k \times k})$. So the problem is that the variance of the new Gaussian depends on $d$, while the new separation between $\mu_i$ and $\mu_j$ is independent of $d$ (only increases by a factor of $\sqrt{k}$)!
}
}
\section{Empirical Results}

We implemented in Python our two main algorithms for $k$-tuple clustering: $\AlgPrivatekAverages$ and $\AlgPrivateNoisykCenters$. We compared the two algorithms in terms of the sample complexity that is needed to privately separate the samples from a given mixture of Gaussians. Namely, how many $k$-tuples we need to sample
such that, when executing $\AlgPrivatekAverages$ or $\AlgPrivateNoisykCenters$, the resulting $k$-tuple $Y = \set{\py_1,\ldots,\py_k}$ satisfies the following requirement: For every $i \in [k]$, there exists a point in $Y$ (call it $\py_i$), such that for every sample $\px$ that was drawn from the $i$'th Gaussian, it holds that $i = \argmin_{j \in [k]} \norm{\px - \py_j}$. 
We perform three tests, where in each test we considered a uniform mixture of $k$ standard spherical Gaussians
around the means 
$\set{R\cdot \pt{e}_i, -R\cdot \pt{e}_i}_{i=1}^{k/2}$, where $\pt{e}_i$ is the $i$'th standard basis vector.
In all the tests, we generated each $k$-tuple by running algorithm k-means++ \cite{kmeansplusplus} over enough samples.

\begin{figure}[t]
	\centerline{\includegraphics[scale=.5]{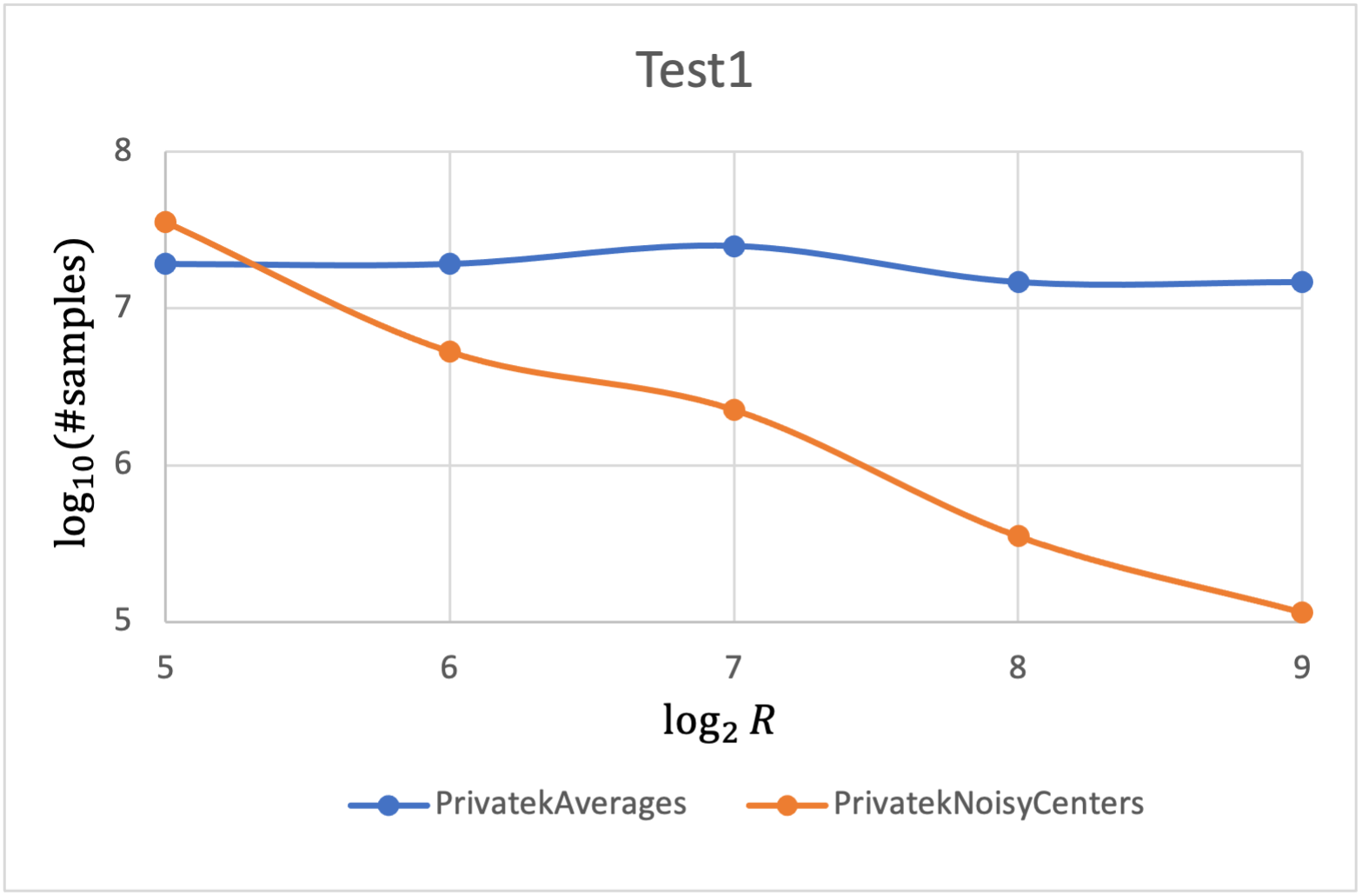}}
	\caption{The case $d=1$ and $k=2$, for varies $R$.}
	\label{Test1}
\end{figure}

\begin{figure}[t]
	\centerline{\includegraphics[scale=.5]{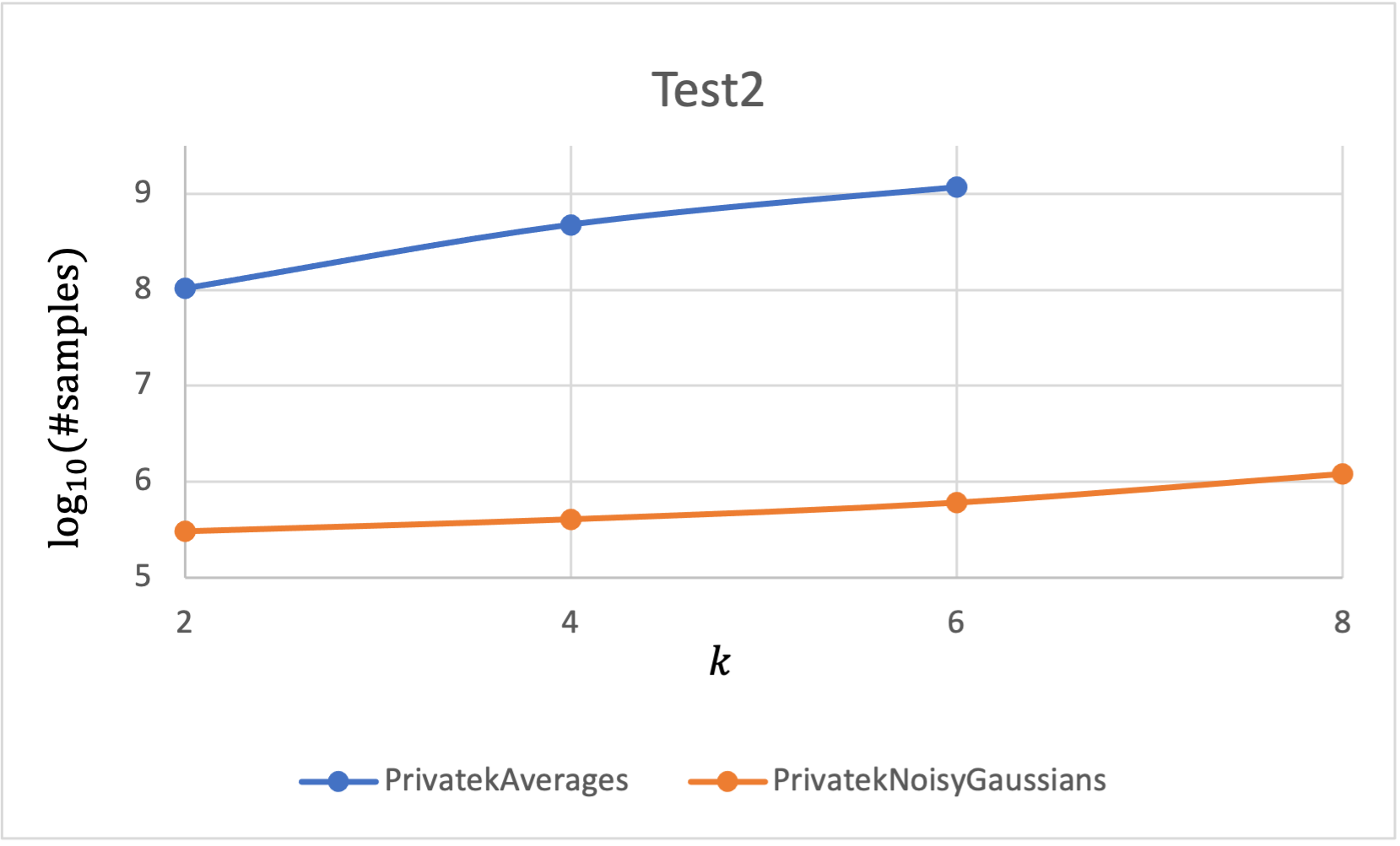}}
	\caption{The case $d=4$ and $R=512 \cdot k$, for varies $k$.}
	\label{Test2}
\end{figure}

\begin{figure}[b]
	\centerline{\includegraphics[scale=.5]{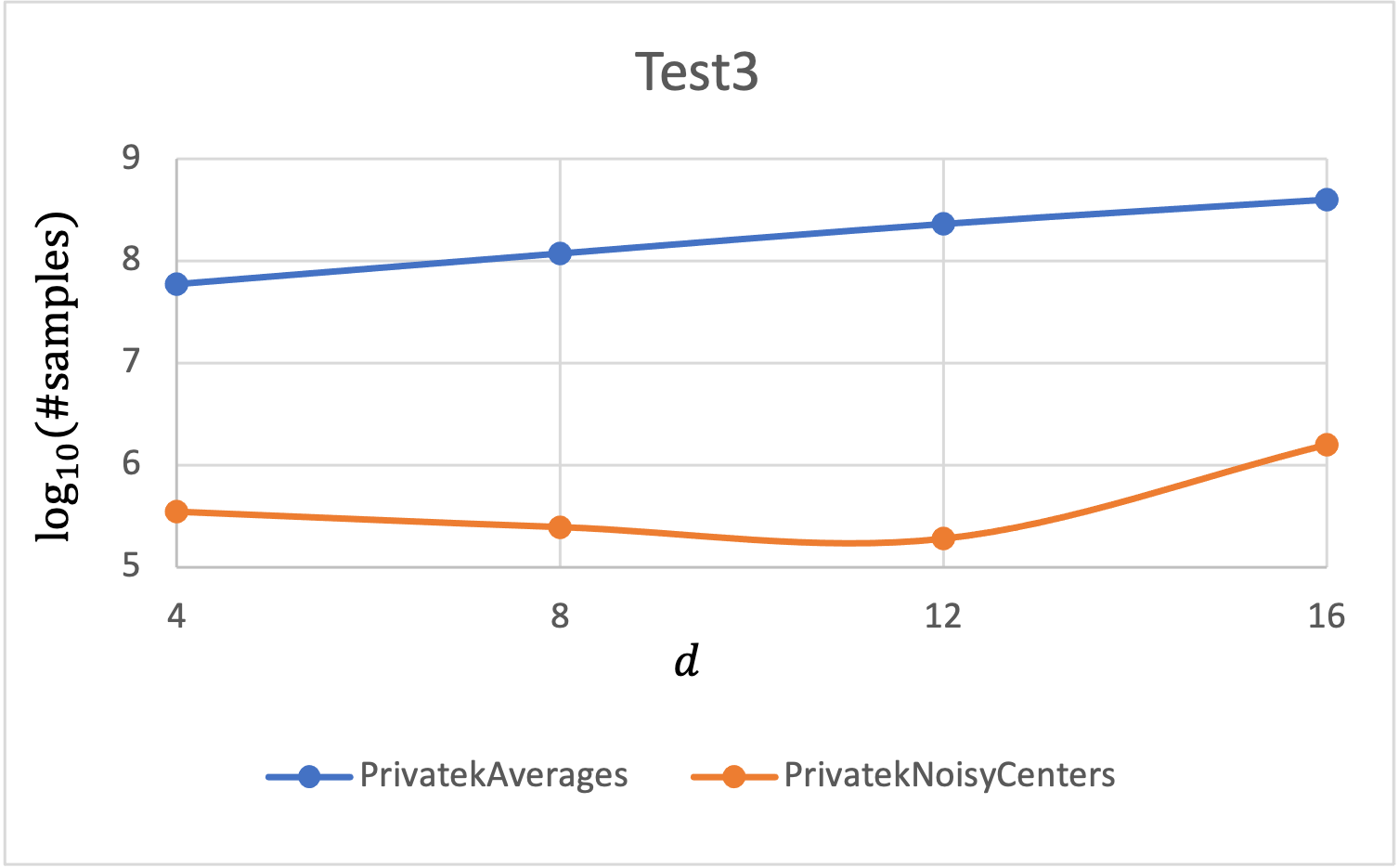}}
	\caption{The case $k=2$, $R = 256 \sqrt{d}$, for varies $d$.}
	\label{Test3}
\end{figure}

In Test1 (\cref{Test1}) we examined the sample complexity in the case $d=1$, $k=2$, for $R \in \set{2^5,2^6,\ldots,2^{9}}$. In Test2 (\cref{Test2}) we examined the case $d=4$, $R = 512 \cdot k$, for $k \in \set{2,4,6,8}$. In Test3 (\cref{Test3}) we examined the case $k=2$, $R=256\sqrt{d}$, for $d \in \set{4,8,12,16}$. In all the experiments we used privacy parameters $\eps = 1$ and $\delta = e^{-28}$, and used $\beta = 0.05$. In all the tests of $\AlgPrivateNoisykCenters$, we chose $\Delta = \frac{10}{\eps}\cdot k \log(k/\delta) \sqrt{\log(k/\beta)}$, the number of $k$-tuples that we generated was exactly $3781$ (the minimal value that is required for privacy), but the number of samples per $k$-tuple varied from test to test. In the tests of $\AlgPrivatekAverages$, we chose $\Lambda =  2^{10} \cdot k \sqrt{d}$ and $r_{\min} = 0.1$, we generated each $k$-tuple using $\approx 15 \cdot k$ samples, but the number of $k$-tuples varied from test to test.\footnote{By using $\tilde{\Omega}(kd)$ samples for creating each $k$-tuple, in Test3 (\cref{Test3}) we could avoid the dependency of $R$ in $\sqrt{d}$ (see \cref{sec:Gaussians:remarks} for more details). However, since we only used $O(k)$ samples for each $k$-tuple when testing $\AlgPrivatekAverages$, then we could not avoid this dependency.}
All the experiments were tested in a MacBook Pro Laptop with 4-core Intel i7 CPU with 2.8GHz, and with 16GB RAM.

The graphs show the main bottleneck of Algorithm $\AlgPrivatekAverages$ in practice. It requires only $O_{\eps,\delta}(k d)$ tuples (or $O_{\eps,\delta}(k \sqrt{d})$ for large values of $d$) in order to succeed, but the hidden constant is $\approx 500,000$ for our choice of $\eps$ and $\delta$, and this does not improve even when the assumed separation $R$ is very large. The cause of this large constant is the group privacy of size $O(k \ell)$ that we do in Step~\ref{step:computing-noisy-bound-aver}, where recall that $\ell = O\paren{\frac{\log^2(1/\delta)}{\eps \log n}}$ (\cref{def:ell}). While in theory this $\ell$ is relatively small, with our choice of parameters we get $\ell \approx 1000$. This means that we need to execute the private average algorithm with $\hat{\eps} \approx \frac{\eps}{4000 k}$. Internally, this $\hat{\eps}$ is shared between other private algorithms, and in particular, with an Interior Point algorithm that is one of the internal components of the average algorithm from \cref{prop:approx-aver-Rd}. This algorithm is implemented using the exponential mechanism \cite{MT07}, which simply outputs a random noise when the number of points is too small.

We remark that prior work on differentially-private clustering, including in "easy" settings, is primarily theoretical. In particular, we are not aware of implemented methods that we could use as a baseline.\footnote{\Enote{Added:}We remark that in different settings, such as node, edge or weight-differential privacy, there exist some available implementations (e.g., \cite{PinotMYGA18}).}
As a sanity check, we did consider the following naive baseline:
For every sample point, add a Gaussian noise to make it private. Now, the resulting noisy samples are just samples from a new Gaussian mixture. Then, run an off-the-shelf  non-private method to learn the parameters of this mixture. 
We tested this naive method on the simple case $d=1$ and $k=2$, where we generated samples from a mixture of standard Gaussians that are separated by $R = 512$. By the Gaussian mechanism, the noise magnitude that we need to add to each point for guaranteeing $(\eps,\delta)$-differential privacy, is $\sigma \approx \frac{\Lambda}{\eps}\sqrt{\log(1/\delta)} \gg 1$ for some $\Lambda > R$, meaning that the resulting mixture consists of very close Gaussians. We applied GaussianMixture from the package sklearn.mixture to learn this mixture, but it failed even when we used $100 M$ samples, as this method is not intended for learning such close Gaussians. We remark that there are other non-private methods that are designed to learn any mixture of Gaussians (even very weakly separated ones) using enough samples (e.g., \cite{suresh2014near}). 
The  sample complexity and running time of these methods, however, are much worse than ours even asymptotically (e.g., the running time of \cite{suresh2014near} is exponential in $k$), and moreover, we are not aware of any implementation we could use.\footnote{Asymptotically, \cite{suresh2014near} requires at least $\tilde{\Omega}(d k^9)$ samples, and runs in time $\tilde{\Omega}(n^2d + d^2 (k^7 \log d)^{k^2})$. For the setting of learning a mixture of $k$ well-separated Gaussians, the approach of first adding noise to each point and then applying a non-private method such as  \cite{suresh2014near}, results with much worse parameters than our result, which only requires $\tilde{O}(dk)$ samples and runs in time $\tilde{O}(dk^2n)$.}

\section{Conclusion}
We developed an approach to bridge the \remove{gaping }gap between the theory and practice of differentially private clustering methods. For future, we hope to further optimize the "constants" in the $k$-tuple clustering algorithms, making the approach practical for instances with lower separation.  Tangentially, the inherent limitations of private versus non-private clustering suggest exploring different rigorous notions of privacy in the context of clustering.

\section*{Acknowledgements}

Edith Cohen is supported by Israel Science Foundation grant no.\ 1595-19.

Haim Kaplan is supported by Israel Science Foundation grant no.\ 1595-19,  and the Blavatnik Family Foundation.	

Yishay Mansour has received funding from the European Research Council (ERC) under the European Union’sHorizon 2020 research and innovation program (grant agreement No. 882396), by the Israel Science Foundation (grant number 993/17) and the Yandex Initiative for Machine Learning at Tel Aviv University.

Uri Stemmer is partially supported by the Israel Science Foundation (grant 1871/19) and by the Cyber Security Research Center at Ben-Gurion University of the Negev.

\printbibliography
\appendix

\section{Additional Preliminaries}

\subsection{Additional Facts About Differential Privacy}

\subsubsection{The Exponential Mechanism}
We next describe the Exponential Mechanism of \citet{MT07}. Let $\cX$ be a domain and $\cH$ a set of solutions. Given a database $\cS \in \cX^*$, the Exponential Mechanism privately chooses a “good” solution $h$ out of the possible set of solutions $\cH$. This “goodness” is quantified using a quality function
that matches solutions to scores.

\begin{definition}(Quality function)
	A quality function is a function $q\colon \cX^* \times \cH \mapsto \bbR$ that maps a database $\cS \in \cX^*$ and a solution $h \in \cH$ to a real number, identified as the score of the solution $h$ w.r.t the database $\cS$.
\end{definition}

Given a quality function $q$ and a database $\cS$, the goal is to chooses a solution $h$ approximately maximizing $q(\cS,h)$. The Exponential Mechanism chooses a solution probabilistically, where the probability mass that is assigned to each solution $h$ increases exponentially with its quality $q(\cS,h)$:

\begin{definition}(The Exponential Mechanism)\label{def:exp-mech}
	Given input parameter $\eps$, finite solution set $\cH$, database $\cS \in \cX^m$, and a sensitivity $1$ quality function $q$, choose randomly $h \in \cH$ with probability proportional to $\exp(\eps\cdot q(\cS,h)/2)$.
\end{definition}

\begin{proposition}(Properties of the Exponential Mechanism)\label{prop:exp-mech}
	(i) The Exponential Mechanism is $\eps$-differentially private. (ii) Let $\hat{e} \eqdef \max_{f \in \cH}\set{q(\cS,f)}$ and $\Delta > 0$. The Exponential Mechanism outputs a solution $h$ such that $q(\cS,h) \leq \hat{e} - \Delta$ with probability at most $\size{\cH}\cdot \exp\paren{-\eps\Delta/2}$.
\end{proposition}

\subsubsection{Private Interior Point  and Bounding Segment in $\bbR$}\label{sec:interior-point}

\begin{proposition}[Finding an Interior Point in $\bbR$]\label{prop:interior-point}
	Let $\eps \in (0,1)$, $\Lambda > 0$ and $\grid \in [0,\Lambda]$.  There exists an efficient $\eps$-differentially private algorithm that takes an $n$-size database $\cS$ of numbers in the segment $[-\Lambda,\Lambda]$ and outputs a number $z \in [-\Lambda,\Lambda]$ that with probability $1 - 2(\Lambda/\grid + 1) \cdot  \exp\paren{-\eps n/4}$ it holds that $z \in [\min(\cS) - \grid, \max(\cS) + \grid]$. The algorithm runs in time $\tilde{O}\paren{n}$ (ignoring $\log\paren{\frac{n \Delta}{g}}$ factors).
\end{proposition}
\begin{proof}
	Define the grid $G = \set{-\Lambda, -\Lambda+\grid, \ldots, -\Lambda + \ceil{\frac{2\Lambda}{\grid}}\cdot \grid}$, and for every $x \in G$ let $\lleft(x) = - \Lambda + \floor{\frac{x + \Lambda}{\grid}} \cdot \grid$ (i.e., the closest grid point to $x$ from the left side) and $\rright(x) = - \Lambda + \ceil{\frac{x + \Lambda}{\grid}} \cdot \grid$ (i.e., the closest grid point to $x$ from the right side). Now, apply the exponential mechanism (\ref{def:exp-mech}) with the quality function 
	\begin{align*}
		\forall y \in G\colon \quad q(\cS,y) = \min\set{\size{\set{x \in \cS \colon \lleft(x) \leq y}}, \size{\set{x \in \cS \colon \rright(x) \geq y}}}
	\end{align*}
	For the utility analysis, let $m$ be the median of $\cS$, and note that $q(\cS,\lleft(m)),  q(\cS,\rright(m)) \geq n/2$.
	Therefore, by \cref{prop:exp-mech}, with probability $\geq 1 - \size{G}\cdot \exp\paren{-\eps n/4} \geq 1 - 2(\Lambda/\grid + 1) \cdot \exp\paren{-\eps n/4}$, the mechanism outputs a point $z$ with $q(S,z) > 0$, which yields in particular that $z \in [\min(\cS) - \grid, \max(\cS) + \grid]$.
	
	For the running time analysis, we implement the sampling as follows: For $x \in \cS$ we let $A_x = \set{\lleft(x)-\grid,\lleft(x),\rright(x), \rright(x) + g}$, and let $A = \cup_{x \in \cS} A_x$. Note that for every consecutive grid points $y, y'=y+g \in G$ with $q(\cS, y) \neq q(\cS, y')$, it holds that $y,y' \in A$: If $q(\cS, y) > q(\cS, y')$, there must exist $x \in \cS$ such that $x \in (y-\grid,y]$, yielding that $y \in [x,x+\grid) \implies y = \rright(x), y' = \rright(x) + \grid$. Otherwise (i.e., $q(\cS, y) < q(\cS, y')$), there must exist $x \in \cS$ such that $x \in [y',y'+\grid)$, yielding that $y' \in (x-\grid,x] \implies y' = \lleft(x), y = \lleft(x) - \grid$.  

	Then, we sort $A$ in time $\tilde{O}(n)$, and let $a_1 \leq \ldots \leq a_m$ be the sorted elements in $A$ (recall that $m = \size{A} \leq 4n$). For each $i \in [m+1]$, we compute $w(\cS,a_i) = q(\cS,a_i) \cdot \size{G \cap (a_{i-1},a_i]}$ (i.e., $w(\cS,a_i)$ is the the original quality of $a_i$ times the number of grid points in $(a_{i-1},a_i]$, where $a_0 = -\Lambda-\grid$ and $a_{m+1} = \Lambda+\grid$), and choose a value $a_i$ with probability $\propto w(\cS,a_i)$. Note that the computation of each $w(\cS,a_i)$ can be done in time $\tilde{O}(1)$ using simple binary searches over the (sorted) multisets $\cS_{\lleft} = \cup_{x \in \cS} \set{\lleft(x)}$ and $\cS_{\rright} = \cup_{x \in \cS} \set{\rright(x)}$ (a ``multiset'' union, that includes duplications). 
	
	Finally, given the chosen $a_i$ from the mechanism, it is left to sample a uniform point in $G \cap (a_{i-1},a_i]$ (since we know, by the property of $A$, that all the point there have the same value of $q(\cS,\cdot)$). This can be easily implemented in time $O(\log \size{G}) = \tilde{O}(1)$.
\end{proof}

\begin{proposition}[Finding a Bounding Segment of Points in $\bbR$]\label{prop:bounding-seg}
	Let $\beta, \eps \in (0,1)$, $\Lambda > 0$ and $\grid \in [0,\Lambda]$.  There exists an efficient $\eps$-differentially private algorithm that takes an $n$-size database $\cS$ of numbers in the segment $[-\Lambda,\Lambda]$ and outputs a segment $[x,y]$ such that with probability at least $1-\beta$ the following holds:
	\begin{itemize}
		\item $\size{\cS \cap [x,y]} \geq n - \frac{8}{\eps} \log \paren{\frac{4\Lambda}{\grid \beta}} - 2$ (i.e., the segment contain most of the points in $\cS$), and
		\item $y - x \leq \max(\cS)-\min(\cS)+ 4\grid$.
	\end{itemize}
	The algorithm runs in time $\tilde{O}\paren{n}$ (ignoring $\log\paren{\frac{n \Delta}{\eps \beta \grid}}$ factors).
\end{proposition}
\begin{proof}
	In the following assume that $n \geq \frac{8}{\eps} \log \paren{\frac{4\Lambda}{\grid \beta}} + 2$ (otherwise the proof trivially holds for any segment $[x,x]$). Let $\cS_0$ be the smallest $\frac{4}{\eps} \log \paren{\frac{4\Lambda}{\grid \beta}} + 1$ points in $\cS$, and let $\cS_1$ be the largest $\frac{4}{\eps} \log \paren{\frac{4\Lambda}{\grid \beta}} + 1$ points in $\cS$. For each $b \in \set{0,1}$ apply \cref{prop:interior-point} (interior point) on $\cS_b$ for finding a number $z_b \in [-\Lambda,\Lambda]$ that belongs to $[\min(\cS_{b}) - \grid, \max(\cS_b) + \grid]$ with probability at least $1 - 2(\Lambda/\grid + 1) \cdot \exp\paren{-\eps \size{\cS_b}/4} \geq 1-\beta/2$.  Therefore, by setting $x = z_0-\grid$ and $y = z_1+\grid$ we get that with probability $1-\beta$ it holds that: (1) $[\max(\cS_0), \min(\cS_1)] \subseteq [x,y]$ and that (2) $[x,y] \subseteq [\min(\cS_0)-2\grid, \max(\cS_1)+2\grid] = [\min(\cS)-2\grid, \max(\cS)+2\grid]$. By (1) we get that all points in $\cS$ expect (at most) $(\size{\cS_0} - 1) + (\size{\cS_1} - 1) \leq \frac{8}{\eps} \log \paren{\frac{4\Lambda}{\grid \beta}}$ are inside $[x,y]$, and by (2) we get that $y-x \geq \max(\cS)-\min(\cS) + 4\grid$, as required.
	
	For the running time analysis, note that by sorting $\cS$ we can determine $\cS_0$ and $\cS_1$ in time $\tilde{O}(n)$, and the cost of executing the algorithm from \cref{prop:interior-point} on each $\cS_b$ is  $\tilde{O}\paren{\frac1{\eps}} = \tilde{O}\paren{n}$.
\end{proof}

\subsubsection{Estimating the Average of Points}\label{sec:approx-aver}

\begin{proposition}[Estimating the Average of Bounded Points in $\bbR$]\label{prop:approx-aver-R}
	Let $\beta, \eps,\delta \in (0,1)$, $\Lambda > 0$ and $r_{\min} \in [0,\Lambda]$. There exists an efficient $(\eps,\delta)$-differentially private algorithm that takes an $n$-size database $\cS$ of numbers in the segment $[-\Lambda,\Lambda]$ and satisfy the following utility guarantee: If $n \geq \frac{16}{\eps} \log \paren{\frac{4\Lambda}{r_{\min} \beta}} +4$, then with probability $1-\beta$, the algorithm outputs a number $\hat{a} \in \bbR$ such that
	\begin{align*}
		\size{\hat{a} - \Avg(\cS)} \leq O\paren{\frac{\max\set{r,r_{\min}}}{\eps n} \paren{\sqrt{\log(1/\delta) \log(1/\beta)} + \log \paren{\frac{\Lambda}{r_{\min} \beta}}}},
	\end{align*}
	where $r = \max(\cS) - \min(\cS)$.
	The algorithm runs in time $\tilde{O}(n)$ (ignoring $\log\paren{\frac{n \Delta}{r_{\min} \eps \beta}}$ factors). 
\end{proposition}
\begin{proof}
	The algorithm does the following: (1) Privately find a bounding segment $[x,y]$ using \cref{prop:bounding-seg} with parameters $\beta/2,\eps/2,\grid = r_{\min},\Lambda$,  let $\hat{r} = y-x$ and let $\cS' = \cS \cap [x,y]$ (2) Use the ($1$-dimensional) Gaussian mechanism (\cref{fact:Gaus}) with $\lambda = \frac{\hat{r}}{\size{\cS'}}$ and parameters $\beta/2,\eps/2, \delta$ for computing a noisy average $\hat{a}$ of $\cS'$ (see \cref{obs:Gaus-aver}). By the properties of the Gaussian mechanism (see \cref{remark:Gaus-add-del}) along with basic composition it holds that the above algorithm is $(\eps,\delta)$-differentially private. For the utility analysis, note that with probability $1-\beta$, the segment $[x,y]$ satisfies the conditions of \cref{prop:bounding-seg} and the noise added to the average in the second step is at most $O\paren{\frac{\hat{r}}{\eps \size{\cS'}} \sqrt{\log(1/\delta) \log(1/\beta)}}$. In the rest of the analysis we assume that this event occurs. Now, by definition of $r$, it holds that
	\begin{align*}
		\size{\Avg(S) - \Avg(\cS')} \leq \frac{r \size{\cS \setminus \cS'}}{n} \leq \frac{8 r}{\eps n} \log \paren{\frac{2\Lambda}{r_{\min} \beta}} +\frac{2 r}{n}
	\end{align*}
	Moreover, it holds that
	\begin{align*}
		\size{\hat{a} - \Avg(\cS')} \leq O\paren{\frac{\hat{r}}{\eps \size{\cS'}} \sqrt{\log(1/\delta) \log(1/\beta)}} \leq 
		O\paren{\frac{\max\set{r,r_{\min}}}{\eps n} \sqrt{\log(1/\delta) \log(1/\beta)}},
	\end{align*}
	where the second inequality holds since $\hat{r} \leq r + 4r_{\min}$ and $\size{\cS'} \geq n/2$ by the assumption on $n$. The proof now follow by the above two inequalities.
	
	For the running time analysis, step (1) takes $\tilde{O}(n)$ time (\cref{prop:bounding-seg}). Step (2) that executes the Gaussian Mechanism,  takes $\tilde{O}(n)$ time for computing the average, and $\tilde{O}(1)$ for sampling a number from a single one-dimensional.
\end{proof}

\begin{proposition}[Estimating the Average of Bounded Points in $\bbR^d$ (Restatement of \cref{prop:approx-aver-Rd})]
	\propEstAvgInRd
\end{proposition}
\begin{proof}
	The algorithm does the following: For each $i \in [d]$, let $\cS_i = \set{x_i \colon (x_1,\ldots,x_d) \in \cS}$ and compute an estimation $\hat{a}_i$ of $\Avg(\cS_i)$ (in time $\tilde{O}(n)$) by applying \cref{prop:approx-aver-R} with parameters $r_{\min},\Lambda$, $\tilde{\eps} = \frac{\eps}{2\sqrt{2d \log(2/\delta)}}$, $\tilde{\delta} = \frac{\delta}{d}$, $\tilde{\beta} = \frac{\beta}{d}$. Finally, output $\hpa = (\hat{a}_1,\ldots, \hat{a}_d)$.  It is clear by advanced composition (\cref{thm:composition2}) that the algorithm is $(\eps,\delta)$-differentially private. For the utility guarantee, note that with probability at least $1-\beta$, for every $i \in [d]$ it holds that 
	\begin{align*}
		\size{\hat{a}_i - \Avg(\cS_i)} 
		&\leq O\paren{\frac{r}{\tilde{\eps} n} \paren{\sqrt{\log(1/\tilde{\delta}) \log(1/\tilde{\beta})} + \log \paren{\frac{\Lambda}{r_{\min} \tilde{\beta}}}}}\\
		&= O\paren{\frac{r \sqrt{d \log(1/\delta)}}{\eps n} \paren{\sqrt{\log(d/\delta) \log(d/\beta)} + \log \paren{\frac{\Lambda d}{r_{\min} \beta}}}},
	\end{align*}
	and hence 
	\begin{align*}
		\norm{\hpa - \Avg(\cS)} 
		&= \sqrt{\sum_{i=1}^d (\hat{a}_i - \Avg(\cS_i))^2}\\
		&\leq O\paren{\frac{r d \sqrt{\log(1/\delta)}}{\eps n} \paren{\sqrt{\log(d/\delta) \log(d/\beta)} + \log \paren{\frac{\Lambda d}{r_{\min} \beta}}}}
	\end{align*}
\end{proof}

\begin{remark}\label{remark:bound-aver-add-del}
	The above two algorithms guarantee differential-privacy whenever two neighboring databases have equal size. However, they can be easily extended to a more general case in which the privacy guarantee also holds in cases of addition and deletion of a point, by extending the Gaussian mechanism used in \cref{prop:approx-aver-R} (see \cref{remark:Gaus-add-del}) with essentially the same noise magnitude.
\end{remark}

\section{Missing Proofs}\label{sec:missing-proofs}

\subsection{Proving \cref{prop:cost-of-sample-is-good}}\label{missing-proof:cost-of-sample-is-good}

In this section we prove \cref{prop:cost-of-sample-is-good}, restated below

\begin{proposition}[Restatement of \cref{prop:cost-of-sample-is-good}]
	\propCostOfSampleIsGood
\end{proposition} 

In the following, fix values of $s$ and $\beta$, let $\xi = \xi(s,\beta)$ and $M = M(s,\beta)$.
The following event and claims are with respect to the random process in \cref{prop:cost-of-sample-is-good}.

\begin{claim}[Event $E$ \cite{ShechnerSS20}]\label{claim:E}
	Let $E$ be the event that for every $C \in B(0,\Lambda)^k$, we have that
	\begin{align*}
	\size{\frac{n}s \cdot \COST_{\cS}(C) -  \COST_{\cP}(C)} \leq \sqrt{M \cdot  \COST_{\cP}(C)} \eqdef \Delta(C)
	\end{align*}
	Then it holds that $\pr{E} \geq 1-\beta$.
\end{claim}

We next prove some useful facts that holds when event $E$ occurs.
\begin{claim}\label{claim:cost-tC}
	Conditioned on event $E$, it holds that
	\begin{align*}
	\COST_{\cP}(\tilde{C}) \leq \omega\cdot \OPT_k(\cP) + \Delta(C^{*}_{\cP}) + \Delta(\tilde{C}),
	\end{align*}
	letting $\tilde{C}$ be the set from \cref{prop:cost-of-sample-is-good}, and letting $C^{*}_{\cP}$ be the optimal $k$-means of $\cP$.
\end{claim}
\begin{proof}
	Let $C^{*}_{\cS}$ be the optimal $k$-means of $\cS$.
	By the assumption on the algorithm $\Ac$, the set $\tilde{C}$ satisfies $\COST_{\cS}(\tilde{C}) \leq \omega \cdot \OPT_k(\cS)$. The proof follows by the following calculation
	\begin{align*}
	\COST_{\cP}(\tilde{C}) 
	&\leq \frac{n}{s} \cdot \COST_{\cS}(\tilde{C}) + \Delta(\tilde{C})\\
	&\leq \omega\cdot \frac{n}{s} \cdot \COST_{\cS}(C^{*}_{\cS}) + \Delta(\tilde{C})\\
	&\leq \omega\cdot \frac{n}{s} \cdot \COST_{\cS}(C^{*}_{\cP}) + \Delta(\tilde{C})\\
	&\leq \omega \cdot \frac{n}{s} \cdot \paren{\frac{m}{n}\cdot \COST_{\cP}(C^{*}_{\cP})  + \frac{s}{n} \cdot \Delta(C^{*}_{\cP})} + \Delta(\tilde{C})\\
	&= \omega\cdot \OPT_k(\cP) + \Delta(C^{*}_{\cP}) + \Delta(\tilde{C}),
	\end{align*}
	where the third inequality holds by event $E$, 
\end{proof}

We now prove a corollary of \cref{claim:cost-tC}.

\begin{corollary}\label{cor:cost-Ct:1}
	Conditioned on event $E$, it holds that
	\begin{align*}
		\Delta(\tilde{C}) \leq 2\paren{M +  \sqrt{M \omega \OPT_k(\cP)}}
	\end{align*}
\end{corollary}
\begin{proof}
	Let $x = \Delta(\tilde{C}) = \sqrt{M \cdot \COST_{\cP}(\tilde{C})}$.
	By \cref{claim:cost-tC}, it holds that
	\begin{align*}
		\frac{x^2}{M} - x \leq \omega\cdot \OPT_k(\cP) + \sqrt{M \cdot \OPT_k(\cP)}.
	\end{align*}
	Since $x \geq 0$, we conclude that
	\begin{align}\label{eq:bounding-x}
		x 
		&\leq \frac12\cdot \paren{M + \sqrt{M^2 + 4M \omega \OPT_k(\cP) + 4M^{1.5} \sqrt{\OPT_k(\cP)}}}\nonumber\\
		&\leq M + \sqrt{M \omega \OPT_k(\cP)} + M^{0.75} \cdot \OPT_k(\cP)^{1/4}\\
		&\leq 2\paren{M +  \sqrt{M \omega \OPT_k(\cP)}},\nonumber
	\end{align}
	where the second inequality holds by the fact that $\sqrt{a+b} \leq \sqrt{a} + \sqrt{b}$ for $a,b \geq 0$, and the last inequality holds since the third term in (\ref{eq:bounding-x}) is either smaller than the first term, or smaller than the second one (recall that $M \geq 1$).
\end{proof}

The proof of \cref{prop:cost-of-sample-is-good} now immediately follows by \cref{claim:cost-tC} and \cref{cor:cost-Ct:1}.

\subsection{Proving \cref{prop:close-centers-have-similar-cost}}\label{missing-proof:close-centers-have-similar-cost}

\begin{proposition}[Restatement of \cref{prop:close-centers-have-similar-cost}]
	\propCloseCentersHaveSimilarCost
\end{proposition}

\begin{proof}
	In the following, for $\px \in \cP$ let $i_{\px} = \argmin_{i}\set{\norm{\px - \pc_{i}}}$ (i.e., the index of the closest center to $\px$ in $C$), and let $j_{\px} = \argmin_{j}\set{\norm{\px - \pc_{j}'}}$ (i.e., the index of the closest center to $\px$ in $C'$).
	It holds that
	\begin{align*}
	\sum_{i=1}^k \OPT_1(\cP_i)
	&\leq \sum_{i=1}^k \sum_{\px \in \cP_i} \norm{\px - \pc_i}^2\\
	&= \sum_{\px \in \cP} \norm{\px - \pc_{j_{\px}}}^2\\
	&= \sum_{\px \in \cP} \norm{\px - \pc_{i_{\px}}}^2 + \sum_{\px \in \cP} \paren{\norm{\px - \pc_{j_{\px}}}^2 - \norm{\px - \pc_{i_{\px}}}^2}\\
	&= \COST_{\cP}(C) + \sum_{\px \in \cP} \paren{\norm{\px - \pc_{j_{\px}}}^2 - \norm{\px - \pc_{i_{\px}}}^2}
	\end{align*}

	In the following, fix $\px \in \cP$.
	We now bound
	\begin{align*}
	\norm{\px - \pc_{j_{\px}}}^2-\norm{\px -\pc_{i_{\px}}}^2 
	=  \left(\norm{\px - \pc_{j_{\px}}}-\norm{\px
		-\pc_{i_{\px}}}\right)\left(\norm{\px
		-\pc_{j_{\px}}}+\norm{\px -\pc_{i_{\px}}}\right)
	\end{align*}

	First, since $\norm{\px - \pc_{j_{\px}}'} \leq \norm{\px - \pc_{i_{\px}}'} $ it holds that
	\begin{align}\label{eq:x-cjx}
	\norm{\px - \pc_{j_{\px}}} \leq \norm{\px - \pc_{j_{\px}}'} + \norm{\pc_{j_{\px}}' - \pc_{j_{\px}}} \leq \norm{\px - \pc_{i_{\px}}'} + \gamma \norm{\pc_{i_{\px}} - \pc_{j_{\px}}}
	\end{align}
	Second,
	\begin{align*}
	\norm{\px - \pc_{i_{\px}}} \geq \norm{\px - \pc_{i_{\px}}'} - \norm{\pc_{i_{\px}}' - \pc_{i_{\px}}} \geq \norm{\px - \pc_{i_{\px}}'} - \gamma \norm{\pc_{i_{\px}} - \pc_{j_{\px}}}
	\end{align*}
	Therefore
	\begin{align*}
	\norm{\px - \pc_{j_{\px}}} - \norm{\px - \pc_{i_{\px}}} \leq 2\gamma \norm{\pc_{i_{\px}} - \pc_{j_{\px}}}
	\end{align*}

	Now, $\norm{\px -\pc_{i_{\px}}} \leq \norm{\px -\pc_{j_{\px}}}$ and therefore
	\begin{align*}
	\norm{\px -\pc_{j_{\px}}}+\norm{\px -\pc_{i_{\px}}}
	&\leq 2 \norm{\px -\pc_{j_{\px}}}\\
	&\leq 2\norm{\px - \pc_{i_{\px}}'} + 2\gamma \norm{\pc_{i_{\px}} - \pc_{j_{\px}}}\\
	&\leq 2\paren{\norm{\px - \pc_{i_{\px}}} + \norm{\pc_{i_{\px}}' - \pc_{i_{\px}}}}+ 2\gamma \norm{\pc_{i_{\px}} - \pc_{j_{\px}}}\\
	&\leq 2\norm{\px - \pc_{i_{\px}}} + 4\gamma \norm{\pc_{i_{\px}} - \pc_{j_{\px}}},
	\end{align*}
	where the second inequality holds by \cref{eq:x-cjx}.

	We now like to bound $\norm{\pc_{i_{\px}} - \pc_{j_{\px}}}$ as
	a function of $\norm{\px - \pc_{i_{\px}}}$.
	We first bound  $\norm{\pc_{i_{\px}} - \pc_{j_{\px}}}$ as a
	function of $\norm{\px - \pc_{i_{\px}}'}$.
	\begin{align}\label{eq:px-pci-lowbound-ym-2}
	2\norm{\px - \pc_{i_{\px}}'}
	&\geq \norm{\px - \pc_{i_{\px}}'} + \norm{\px - \pc_{j_{\px}}'}\nonumber\\
	&\geq \norm{\pc_{i_{\px}}' - \pc_{j_{\px}}'}\nonumber\\
	&\geq \norm{\pc_{i_{\px}} - \pc_{j_{\px}}} - \norm{\pc_{i_{\px}} - \pc_{i_{\px}}'} - \norm{\pc_{j_{\px}} - \pc_{j_{\px}}'}\nonumber\\
	&\geq (1-2\gamma) \norm{\pc_{i_{\px}} - \pc_{j_{\px}}},
	\end{align}
	In addition
	\[
	\norm{\px - \pc_{i_{\px}}'}\leq \norm{\px - \pc_{i_{\px}}} +
	\norm{\pc_{i_{\px}} - \pc_{i_{\px}}'} \leq  \norm{\px -
		\pc_{i_{\px}}} + \gamma \norm{\pc_{i_{\px}} - \pc_{j_{\px}}}
	\]
	Therefore,
	\[
	2\norm{\px - \pc_{i_{\px}}} \geq
	(1-4\gamma)
	\norm{\pc_{i_{\px}} - \pc_{j_{\px}}}
	\]
	We have that
	\begin{align*}
	\norm{\px - \pc_{j_{\px}}}^2-\norm{\px -\pc_{i_{\px}}}^2 &=
	\left(\norm{\px - \pc_{j_{\px}}}-\norm{\px
		-\pc_{i_{\px}}}\right)\left(\norm{\px
		-\pc_{j_{\px}}}+\norm{\px -\pc_{i_{\px}}}\right)\\
	&\leq \left(2\gamma \norm{\pc_{i_{\px}} -
		\pc_{j_{\px}}}\right)\left(2 \norm{\px - \pc_{i_{\px}}}
	+4\gamma \norm{\pc_{i_{\px}} - \pc_{j_{\px}}}\right)\\
	&\leq \left(\frac{4\gamma }{1-4\gamma}\norm{\px - \pc_{i_{\px}}}
	\right)\left((2+\frac{8\gamma }{1-4\gamma}) \norm{\px -
		\pc_{i_{\px}}} \right)\\
	&\leq 32\gamma \norm{\px - \pc_{i_{\px}}}^2,
	\end{align*}
	where the least inequality holds since $\gamma \leq 1/8$.
	Now we can get the bound on the summation:
	\begin{align*}
	\sum_{\px \in \cP} \paren{\norm{\px - \pc_{j_{\px}}}^2 - \norm{\px - \pc_{i_{\px}}}^2}
	\leq  \sum_{\px \in \cP} 32\gamma \norm{\px - \pc_{i_{\px}}}^2
	\leq 32\gamma \COST_{\cP}(C)
	\end{align*}
\end{proof}

\subsection{Proving \cref{thm:kGauss-utility}}\label{missing-proof:thm:kGauss-utility}

In this section we prove the utility guarantee of $\AlgPrivatekGaussians$. 
We first by proving the following proposition that states the following: Assume that $\pX \sim \cN(\mu,\Sigma)$ with $\norm{\Sigma} = \sigma^2$,
and let $\py,\pz \in \bbR^d$ such that (1) $\norm{\py - \mu}$ is ``large enough'' (larger than $\Omega\paren{\sigma \sqrt{\log(1/\beta)}}$ ) , and (2) $\norm{\pz - \mu}$ is ``small enough''. Then with probability $1-\beta$ (over $\pX$) it holds that $\norm{\pX - \pz} < \norm{\pX - \py}$. Note that such an argument is trivial when $\norm{\py - \mu}$ is at least $\Omega(\sigma \sqrt{d \log(1/\beta)})$, but using a standard projection argument, we can avoid the dependency in $d$.

\begin{proposition}\label{prop:separation}
	Let $\pX \sim \cN(\mu,\Sigma)$ where $\norm{\Sigma} = \sigma^2$,  let $\py \in \bbR^d$ with $\norm{\py - \mu} \geq 2(1+\gamma)\sqrt{2\log\paren{1/\beta}} \cdot \sigma$ for some $\gamma > 0$, and let $\pz \in \bbR^d$ with $\norm{\pz - \mu} \leq \frac{\gamma}{3(1+\gamma)} \norm{\py - \mu}$. Then with probability $1-\beta$ (over the choice of $\pX$), it holds that $\norm{\pX - \pz} < \norm{\pX-\py}$.
\end{proposition}
\begin{proof}
	Assume w.l.o.g. that $\mu = \pt{0}$. 
	Let $\pW = \pz + \frac{\iprod{\pX-\pz,\py-\pz}}{\norm{\py-\pz}^2} (\py-\pz)$ be the projection of $\pX$ onto the line between $\py$ and $\pz$. In the following we bound the probability that $\frac{\iprod{\pX-\pz,\py-\pz}}{\norm{\py-\pz}^2} < \frac12$, which implies that $\norm{\pW - \pz} < \norm{\pW-\py}$, and therefore, $\norm{\pX - \pz} < \norm{\pX-\py}$. Note that $\ip{\pX,\py-\pz}$ is distributed according to the (one dimensional) Gaussian $\cN(\pt{0}, (\py-\pz)^T \Sigma (\py-\pz))$ and it holds that $(\py-\pz)^T \Sigma (\py -\pz) \leq \sigma^2 \norm{\py-\pz}$. Therefore, by \cref{fact:one-Gaus-concet} we obtain that with probability $1-\beta$ it holds that $\ip{\pX,\py-\pz} < \sigma \norm{\py-\pz} \sqrt{2 \log(1/\beta)}$, and in the following we continue with the analysis assuming that this occurs. The proposition now follows by the following calculation.
	\begin{align*}
	\frac{\iprod{\pX-\pz,\py-\pz}}{\norm{\py-\pz}^2}
	&= \frac{\ip{\pX,\py-\pz} - \ip{\pz,\py-\pz}}{\norm{\py-\pz}^2}\\
	&< \frac{\sigma \norm{\py-\pz} \sqrt{2 \log(1/\beta)} + \norm{\pz} \norm{\py -\pz}}{\norm{\py-\pz}^2}\\
	&\leq \frac{\sigma \sqrt{2 \log(1/\beta)}}{\paren{1 - \frac{\gamma}{3(1+\gamma)}} \norm{y}} + \frac{\frac{\gamma}{3(1+\gamma)}}{1 - \frac{\gamma}{3(1+\gamma)}}\\
	&\leq \frac{1}{2(1+\gamma) \paren{1 - \frac{\gamma}{3(1+\gamma)}}} + \frac{\frac{\gamma}{3(1+\gamma)}}{1 - \frac{\gamma}{3(1+\gamma)}}\\
	&= \frac{1 + \frac{2 \gamma}{3}}{2(1 + \gamma)\frac{3 + 2\gamma}{3(1+\gamma)}}\\
	&= \frac12,
	\end{align*}
	where in the second inequality holds since $\norm{\py-\pz} \geq \norm{\py} - \norm{\pz} \geq \paren{1 - \frac{\gamma}{3(1+\gamma)}} \norm{y}$, and the third inequality holds by the assumption on $\norm{\py}$.
\end{proof}

In addition, we use the following fact.

\begin{fact}\label{fact:dTV-of-mixtures}
	Let $\cD = \sum_{i=1}^k w_i \cD_i$ be a mixture of the $k$ distributions $\cD_1, \ldots, \cD_k$, and let $\cD' = \sum_{i=1}^k w_i' \cD_i'$ be a mixture of the $k$ distributions $\cD_1', \ldots, \cD_k'$. Assume that for every $i \in [k]$ it holds that $\dTV(\cD_i,\cD_i') \leq \frac{\alpha}{2}$ and $\size{w_i - w_i'} \leq \frac{\alpha}{k}$. Then $\dTV(\cD,\cD) \leq \alpha$.
\end{fact}

We now ready to prove \cref{thm:kGauss-utility}, stated for convenient below.

\begin{theorem}[Restatement of \cref{thm:kGauss-utility}]
	\thmKGaussUtility
\end{theorem}
\begin{proof}
		Let $E_1 =  \bigwedge_{j \in [t], i \in [k]} E_1^{t,i}$ where $E_1^{j,i}$ is the event that the $s$-size set $\cS_j$ in Step~\ref{step:sample-gaus} of $\AlgCollectEmpiricalMeans$ contains at least $\frac{w_i s}{2}$ samples from the $i$'th Gaussian. Note that for every $j \in [t]$ and $i \in [k]$, it holds that
	\begin{align*}
	\pr{E_1^{j,i}} 
	&= \pr{\Bin(s,w_i) \geq \frac{s w_i}{2}}\\
	&\geq 1 - \pr{\Bin(s,w_{\min}) < \frac{s w_{\min}}{2}}\\
	&\geq 1 - e^{-\frac{w_{\min} s}{4}}
	\end{align*}
	where the last inequality holds by \cref{fact:binom_concentration}. Therefore, we obtain that $\pr{E_1^{j,i}}  \geq 1 - \frac{\beta}{8 k t}$ whenever $s \geq \frac{4}{w_{\min}} \log\paren{8 k t/\beta}$, which holds by the assumption on $s$. By the union bound, we deduce that 
	
	\begin{align}\label{eq:E1}
	\pr{E_1} \geq 1 - \beta/8
	\end{align}
	
	In the following, assume that event $E_1$ occurs. For $j \in [t]$ and $i \in [k]$ let 
	$\hat{\cS}_j^i$ be all the points in $\cS_j$ that have been drawn from the $i$'th Gaussian $\cN(\mu_{i},\Sigma_{i})$, and let $\hat{\mu}_{j,i} =  \Avg\paren{\hat{\cS}_j^i}$. Let $E_2 =  \bigwedge_{j \in [t], i \in [k]} E_2^{j,i}$, where  $E_2^{j,i}$ is the event that $\norm{\hat{\mu}_{j,i} - \mu_i} \leq \frac{\gamma h}{16}  \cdot \sigma_{i}$. Since $\hat{\mu}_{j,i}$ is the average of at least $\frac{w_{i} s}{2}$ samples from the Gaussian $\cN(\mu_i,\Sigma_i)$, we obtain by \cref{fact:gaus-avg} that with probability $1 - \frac{\beta}{8 k t}$ it holds that
	\begin{align}\label{eq:hmu_i-to-mu_i}
	\norm{\hat{\mu}_{j,i} - \mu_i} 
	\leq \frac{\sqrt{2d} + 2\sqrt{\log \paren{\frac{8 k t}{\beta}}}}{\sqrt{w_i s}} \cdot \sigma_i
	\leq \frac{(1+\gamma) h}{\Delta} \cdot \sigma_i,
	\end{align}
	where the last inequality holds by the assumption on $s$ (by the second term in the maximum). Therefore, event $E_2^{j,i}$ occurs with probability at least $1 - \frac{\beta}{8 k t}$, and we conclude by the union bound that 
	\begin{align}\label{eq:E2}
	\pr{E_2 \mid E_1} \geq 1 - \beta/8
	\end{align}
	
	Let $E_3 = \bigwedge_{j=1}^t E_3^j$, where $E_3^j$ is the event that the resulting labeling function $L_j$ in Step~\ref{step:labeling} of the $j$'th iteration in $\AlgCollectEmpiricalMeans$ satisfies:
	\begin{align*}
	\forall \px,\px' \in \cS_j:\quad  \px,\px' \text{ were drawn from the same Gaussian } \iff L_j(\px) = L_j(\px').
	\end{align*}
	Since $\Ac$ is a $(s,\frac{\beta}{8 t})$-labeling algorithm for $\cD$, it holds that $\pr{E_3^j} \geq 1 - \frac{\beta}{8  t}$ for every $j \in [t]$, and we deduce by the union bound that
	\begin{align}\label{eq:E3}
	\pr{E_3} \geq 1 - \beta/8
	\end{align}
	
	In the rest of the analysis we assume that event $E_1 \land E_2 \land E_3$ occurs. This means that for every $j \in [t]$ there exists a permutation $\pi_j$ over $[k]$  such that for each $i \in [k]$, the set of all points in $\cS_j$ that have been drawn from the $i$'th Gaussian (which we denoted by $\cS_j^i$) equals to $\set{\px \in \cS_j \colon L_j(\px) = \pi_j(i)}$, and assume without loss of generality that  for all $j \in [t]$, $\pi_j$ is the identity (i.e., $\pi_j(i) = i$). Therefore, for all $j \in [t]$ and $i \in [k]$ it holds that $\hat{\mu}_{j,i} = \bar{\mu}_{j,i}$, where $\bar{\mu}_{j,i}$ is the empirical mean from Step~\ref{step:compute-emp-mean}. Namely, we obtained that 
	\begin{align}\label{eq:bar_mu-vs-mu}
	\forall j \in [t], i \in [k]: \quad \norm{\bar{\mu}_{j,i} - \mu_i} \leq \frac{(1+\gamma) h}{\Delta} \cdot \sigma_{i},
	\end{align}
	and in particular, it holds that
	\begin{align}\label{eq:bar_mu-le-Lambda}
	\forall j \in [t], i \in [k]: \quad \norm{\bar{\mu}_{j,i}} \leq \norm{\mu_{j,i}} + \frac{(1+\gamma) h}{\Delta} \cdot \sigma_{i} \leq \Lambda
	\end{align}
	Therefore, we deduce that $\cT$ from Step~\ref{step:kAveragesOnM} of $\AlgPrivatekGaussians$ is contained in $(B(\pt{0},\Lambda)^k)^*$, and is partitioned by the $\Delta$-far balls $\cB = \set{B_i(\mu_i, r_i = \frac{(1+\gamma) h}{\Delta} \cdot \sigma_i )}_{i=1}^k$ (\cref{def:sep-balls}), where $\Partition(\cT)$ is exactly $\set{\cP_1 =\set{ \bar{\mu}_{j,1}}_{j=1}^t,\ldots,\cP_k = \set{\bar{\mu}_{j,k}}_{j=1}^t}$ (note that the balls are indeed $\Delta$-far by the separation assumption that $\norm{\mu_i - \mu_j} \geq (1+\gamma) h \max\set{\sigma_i,\sigma_j}$).
	Therefore, since $\Bc$ is an $(t,\text{ }\alpha=1,\text{ }r_{\min}=\frac{(1+\gamma) h}{\Delta}\cdot \sigma_{\min},\text{ }\beta/8, \text{ }\Delta,\text{ }\Lambda=R + \frac{(1+\gamma) h}{\Delta}\cdot \sigma_{\max})$-averages-estimator, we obtain that the output $\set{\hat{\pa}_1,\ldots,\hat{\pa}_k}$  of $\Bc(\cT)$ in Step~\ref{step:kAveragesOnM} satisfy w.p. $1-\beta/8$ that:
	
	\begin{align}\label{eq:a-to-avg}
		\forall i \in [k]: \quad \norm{\hpa_i - \Avg\paren{\cP_i}} \leq \max\set{r_i,r_{\min}} \leq \frac{(1+\gamma)h}{\Delta}\cdot \sigma_i
	\end{align}
	
	In the following, we denote by $E_4$ the event that \cref{eq:a-to-avg} occurs, where recall that we proved that
	\begin{align}
	\pr{E_4 \mid E_1 \land E_2 \land E_3} \geq 1 - \beta/8
	\end{align}
	
	In the following, we also assume that event $E_4$ occurs.
	Recall that by \cref{eq:bar_mu-vs-mu}, for each $j \in [t]$ and $i \in [k]$ it holds that
	\begin{align}\label{eq:avg-to-mu_i}
	\norm{\Avg\paren{\cP_i} - \mu_i}
	\leq  \frac{(1+\gamma)h}{\Delta} \cdot \sigma_i,
	\end{align}
	and we deduce by \cref{eq:a-to-avg,eq:avg-to-mu_i} that for all $i \in [k]$ it holds that
	\begin{align}\label{eq:hpa_i-to-mu_i}
	\norm{\hpa_i - \mu_i} \leq  \frac{2(1+\gamma)h}{\Delta} \cdot \sigma_i.
	\end{align}
	Therefore, for all $i \neq j$ it holds that
	\begin{align}\label{eq:hpa_i-to-mu_j}
	\norm{\hpa_j - \mu_i} 
	\geq \norm{\mu_i - \mu_j} - \norm{\hpa_j - \mu_j} 
	&\geq \paren{1-2/\Delta} (1+\gamma)h \cdot \max\set{\sigma_i,\sigma_j}\\
	&\geq (1+\gamma/2)h \cdot \max\set{\sigma_i,\sigma_j}\nonumber
	\end{align}
	where the second inequality holds by the separation assumption along with \cref{eq:avg-to-mu_i}, and the last one holds by the choice of $\Delta$.
	Hence, we deduce from \cref{eq:hpa_i-to-mu_i,eq:hpa_i-to-mu_j} that for each $i \neq j$ it holds that 
	\begin{align}\label{eq:hpa_i-to-mu_i-sec}
		\norm{\hpa_i - \mu_i} \leq \frac{2}{\Delta - 2} \cdot \norm{\hpa_j - \mu_i} = \frac{\gamma/2}{3(1+\gamma/2)}\cdot \norm{\hpa_j - \mu_i},
	\end{align}
	where the last inequality holds since recall that $\Delta = 8 + 12/\gamma$.
	Since $h \geq 2 \sqrt{2 \log\paren{1/\beta'}}$ for $\beta' = \frac{\beta}{8n}$, then by \cref{prop:separation} (when setting $\py = \hpa_j$ and $\pz = \hpa_i$) along with \cref{eq:hpa_i-to-mu_j,eq:hpa_i-to-mu_i-sec}, for every $i \neq j$, when sampling a point $\px$ from the $i$'th Gaussian $\cN(\mu_i,\sigma_i)$, then with probability $1 - \frac{\beta}{8n}$ it holds that $\norm{\px - \hpa_i} < \norm{\px - \hpa_j}$. Therefore, let $E_5$ be the event that for all $i \in [k]$ and all $\px \in \cP''$ that have been sampled from the $i$'th Gaussian $\cN(\mu_i,\Sigma_i)$, it holds that $\hpa_i$ is the closest point to each of them among $\set{\hat{\pa}_1,\ldots,\hat{\pa}_k}$. Then by the union bound it holds that
	\begin{align}\label{eq:E_5}
	\pr{E_5 \mid E_1 \land E_2 \land E_3 \land E_4} \geq 1 - \beta/8
	\end{align}
	
	In the following we also assume that event $E_5$ occurs. Let $E_6 =  \bigwedge_{i \in [k]} E_6^{i}$ where $E_6^{i}$ is the event that $\cP''$ contains at least $\frac{w_i n}{2}$ samples from the $i$'th Gaussian (namely, $\size{\cP''_i} \geq \frac{w_i n}{4}$). Similar calculation to bounding $\pr{E_1}$, it holds that 
	\begin{align}\label{eq:E_6}
	\pr{E_6 \mid E_1 \land \ldots \land E_5} \geq 1 - \beta/8
	\end{align}
	provided that $n \geq \frac{4}{w_{\min}} \log\paren{8 k/\beta}$, which holds by the assumption on $n$.
	
	In the following we assume that event $E_6$ occurs, and let $E_7 = \land_{i=1}^k E_7^i$, where $E_7^i$ is the event that the output $(\hat{\mu}_i, \hat{\Sigma}_i)$ of the private algorithm $\Ac'$ in Step~\ref{step:priv-single-Gauss-est} of the $i$'th iteration satisfies $\dTV(\cN(\mu_i,\Sigma_i), \cN(\hat{\mu}_i, \hat{\Sigma}_i)) \leq \eta/2$.  By the assumption on algorithm $\Ac'$, we obtain that 
	$\pr{E_7^i} \geq 1 - \frac{\beta}{16k}$ whenever $\size{\cP''_i} \geq \upsilon$, which holds w.p. $1-\frac{\beta}{16}$ when $n \geq \frac{2\upsilon + \log(16k/\beta)}{w_i}$ (follows by \cref{fact:binom_concentration} for $\Bin(n,w_i)$ and $t = \sqrt{w_i n \log(16k/\beta)}$). Since $n \geq \frac{2\upsilon+ \log(16k/\beta)}{w_{\min}}$ by assumption, we obtain by the union bound that  
	\begin{align}\label{eq:E_7}
	\pr{E_7 \mid E_1 \land \ldots \land E_6} \geq 1 - \beta/8.
	\end{align}
	
	In the following, for $i \in [k]$ let $L_i$ be the value of the Laplace noise in Step~\ref{step:Lap} of the $i$'th iteration, let $E_8^i$ be the event that $\size{L_i} \leq \frac2{\eps} \log\paren{16 k/\beta}$, and let $E_8 = \land_{i=1}^k E_8^i$. By \cref{fact:laplace-concent}, for any fixing of $i \in [k]$ it holds that $\pr{E_8^i} \geq 1 - \frac{\beta}{8 k}$, and therefore, by the union bound it holds that 
	\begin{align}\label{eq:E_8}
	\pr{E_8} \geq 1 - \beta/8.
	\end{align}
	In the following we also assume that $E_8$ occurs. It is left to show that when event $E_1 \land \ldots \land E_8$ occurs, then for every $i \in [k]$ it holds that
	\begin{align}\label{eq:hw_i-w_i}
	\forall i \in [k]: \quad \size{\hat{w}_i - w_i} \leq \eta/k.
	\end{align}
	Indeed, given \cref{eq:hw_i-w_i} and event $E_7$, we deduce by \cref{fact:dTV-of-mixtures} that $\dTV(\cD,\hat{\cD}) \leq \eta$, which holds with probability at least $\pr{E_1 \land \ldots \land E_8} \geq 1-\beta$ (holds by \cref{eq:E1} to \cref{eq:E_8}).
	
	We now prove that \cref{eq:hw_i-w_i} holds when $E_1 \land \ldots \land E_8$ occurs.
	Fix $i \in [k]$, let $L = \sum_{j=1}^k L_j$, and compute
	\begin{align*}
	\size{\hat{w}_i - w_i}
	&= \size{\frac{\hat{n}_i}{\hat{n}} - \frac{n_i}{n}}
	=  \size{\frac{n_i + L_i}{n + L} - \frac{n_i}{n}}\\
	&= \size{\frac{n L_i - n_i L}{n(n+L)}}
	= \size{\frac{(n-n_i)L_i - n_i \sum_{j \neq i} L_j}{n(n+L)}}\\
	&\leq \frac{\frac{2k}{\eps} \log\paren{8 k/\beta}}{n - \frac{2k}{\eps} \log\paren{8 k/\beta}}\\
	&\leq \eta/k,
	\end{align*}
	where the first inequality holds by event $E_8$, and the last one holds whenever $n \geq \frac{4 k^2}{\eps \eta} \cdot \log\paren{8 k/\beta}$, which holds by the assumption on $n$.
	
\end{proof}

\end{document}